\documentclass[fleqn,10pt]{wlscirep}
\usepackage{siunitx}
\usepackage[utf8]{inputenc}
\usepackage[english]{babel}
\usepackage[T1]{fontenc}
\usepackage{amsmath}
\usepackage{amssymb}
\usepackage{graphicx}
\usepackage{multirow}
\usepackage{booktabs}
\usepackage[colorinlistoftodos]{todonotes}
\usepackage{float}
\usepackage[utf8]{inputenc}
\usepackage[hang]{subfigure}
\usepackage{rotating}
\usepackage{caption}
\usepackage{colortbl}
\usepackage{tablefootnote}
\usepackage{silence}
\usepackage{ulem}
\usepackage{hyperref}
\usepackage{fancyhdr}
\usepackage{comment}
\usepackage{setspace}
\title{\textbf{In Pursuit of Interpretable, Fair and Accurate Machine Learning for Criminal Recidivism Prediction}}
\author[1,+]{Caroline Wang}
\author[2,+]{Bin Han}
\author[3]{Bhrij Patel}
\author[3,4,**]{Cynthia Rudin}
\affil[1]{Department of Computer Science, The University of Texas at Austin, Austin, TX, 78712, USA}
\affil[2]{Department of Information Science, The University of Washington, Seattle, WA, 98195, USA}
\affil[3]{Department of Computer Science, Duke University, Durham, NC, 27708, USA}
\affil[4]{Department of Statistical Science, Duke University, Durham, NC, 27708, USA}
\affil[+]{These authors contributed equally to this work}
\affil[*]{Correspondence and requests for materials should be addressed to B.Han (bh193@uw.edu).}
\affil[**]{Proofs should be sent to C. Rudin, addressed to LSRC D342, Research Drive, Durham NC, 27708 USA. }

\begin{document}
\maketitle

\doublespacing

\section*{Declarations}
% \subsection*{Funding}
\textbf{Funding \;\;}  This study was partially supported by Arnold Ventures, the Department of Computer Science at Duke University, the Department of Electrical and Computer Engineering at Duke University, and the Lord Foundation of North Carolina. This report represents the findings of the authors and does not represent the views of any of the funding agencies.
\\
% \subsection*{Conflicts of interest/Competing interests}
\textbf{Conflicts of interest/Competing interests \;\;} No additional institutional conflicts.
\\
% \subsection*{}
\textbf{Data Availability Statement \;\;}
The Broward County, FL dataset generated and analyzed during the current study is available from the corresponding author on request. The Kentucky dataset is not publicly available but can be accessed through a special data request to the Kentucky Department of Shared Services, Research and Statistics. 
\\
% \subsection*{Code}
\textbf{Code \;\;}
Our code is here: \url{https://github.com/BeanHam/interpretable-machine-learning}
\\
% \subsection*{Acknowledgements}
\textbf{Acknowledgements \;\;}
We thank the Broward County Sheriff's office and the Kentucky Department of Shared Services, Research and Statistics for their assistance and provision of data. We would also like to thank Daniel Sturtevant from the Kentucky Department of Shared Services, Research and Statistics for providing significant insight into the Kentucky data set, and Berk Ustun for his advice on training RiskSLIM. Finally, we thank Brandon Garrett from Duke, Stuart Buck and Kristin Bechtel from Arnold Ventures, and Kathy Schiflett, Christy May, and Tara Blair from Kentucky Pretrial Services for their thoughtful comments on the article. 

%\newpage 
\section*{Abstract}
\textbf{Objectives \;\;} We study interpretable recidivism prediction using machine learning (ML) models and analyze performance in terms of prediction ability, sparsity, and fairness. Unlike previous works, this study trains interpretable models that output probabilities rather than binary predictions, and uses quantitative fairness definitions to assess the models. This study also examines whether models can generalize across geographic locations.
\\ 
\textbf{Methods \;\;} We generated black-box and interpretable ML models on two different criminal recidivism datasets from Florida and Kentucky. We compared predictive performance and fairness of these models against two methods that are currently used in the justice system to predict pretrial recidivism: the Arnold PSA and COMPAS. We evaluated predictive performance of all models on predicting six different types of crime over two time spans.
\\ 
\textbf{Results \;\;} Several interpretable ML models can predict recidivism as well as black-box ML models and are more accurate than COMPAS or the Arnold PSA. These models are potentially useful in practice. Similar to the Arnold PSA, some of these interpretable models can be written down as a simple table. Others can be displayed using a set of visualizations. Our geographic analysis indicates that ML models should be trained separately for separate locations and updated over time. We also present a fairness analysis for the interpretable models. 
\\
\textbf{Conclusions \;\;} Interpretable machine learning models can perform just as well as non-interpretable methods and currently-used risk assessment scales, in terms of both prediction accuracy and fairness. Machine learning models might be more accurate when trained separately for distinct locations and kept up-to-date. 
\\
% Please include a maximum of four to five keywords (JOQC)
\textbf{Keywords \;\;} criminal recidivism, interpretability, fairness, COMPAS, machine learning.

%%%%%%%%%%%%%%%%%%%%%%%%%%%%%%%%%
\section*{Abstract of Main Results for Criminal Justice Practitioners}

Our goal is to study the predictive performance, interpretability, and fairness of machine learning models for pretrial recidivism prediction. Machine learning methods are known for their ability to automatically generate high-performance models that sometimes even surpass human performance from data alone. However, many of the most common machine learning approaches produce ``black-box'' models---models that perform well, but are too complicated for humans to understand. ``Interpretable'' machine learning techniques seek to produce the best of both worlds: models that perform as well as black-box approaches, but also are understandable to humans. In this study, we generate multiple black-box and interpretable machine learning models. We compare the predictive performance and fairness of the machine learning models we generate, against two models that are currently used in the justice system to predict pretrial recidivism---namely, the Risk of General Recidivism and Risk of Violent Recidivism scores from the COMPAS suite, and the New Criminal Activity and New Violent Criminal Activity scores from the  Arnold Public Safety Assessment.  

We first evaluate the predictive performance of all models, based on their ability to predict recidivism for six different types of crime: \texttt{general, violent, drug, property, felony, and misdemeanor}. Recidivism is defined as a new charge for which an individual is \textit{convicted} within a specified time frame, which we specify as six months or two years. We consider each type of recidivism over the two time periods to control for time, rather than to consider predictions over an arbitrarily long or short pretrial period. Next, we examine whether a model constructed using data from one region suffers in predictive performance when applied to predict recidivism in another region. Finally, we consider the latest fairness definitions created by the machine learning community. Using these definitions, we examine the behavior of the interpretable models, COMPAS, and the Arnold Public Safety Assessment, on race and gender subgroups.

Our findings and contributions can be summarized as follows: 
\begin{itemize}
    \item We contribute a set of interpretable machine learning models that can predict recidivism as well as black-box machine learning methods and better than COMPAS or the Arnold Public Safety Assessment for the location they were designed for. These models are potentially useful in practice. Similar to the Arnold Public Safety Assessment, some of these interpretable models can be written down as a simple table that fits on one page of paper. Others can be displayed using a set of visualizations.
    
    \item We find that recidivism prediction models that are constructed using data from one location do not tend to perform as well when they are used to predict recidivism in another location, leading us to conclude that models should be constructed on data from the location where they are meant to be used, and updated periodically over time.
    
    \item We reviewed the recent literature on algorithmic fairness, but most of the fairness criteria don't pertain to risk scores, they pertain only to yes/no classification decisions. Since we are interested in criminal justice risk scores in this work, the vast majority of the algorithmic fairness criteria are not relevant. We chose to focus on the evaluation criteria that were relevant, namely calibration and balanced group AUC (BG-AUC). We present an analysis of these fairness measures for two of the interpretable models (RiskSLIM and Explainable Boosting Machine) and the Arnold Public Safety Assessment (New Criminal Activity score) on the two-year general recidivism outcome in Kentucky. We found that the fairness criteria were approximately met for both interpretable models for blacks/whites and males/females---that is, the models were fair according to these criteria. The Arnold Public Safety Assessment's New Criminal Activity score failed to satisfy calibration for higher values of the score. The results on fairness were not as consistent for the ``Other'' race category. 
    It is difficult to interpret the fairness result for the ``Other'' race category, due to low-resolution race data.
\end{itemize}

%%%%%%%%%%%%%%%%%%%%%%%%%%%%%%%
\section{Introduction} \label{Introduction}

Predicting criminal recidivism using statistics has been the subject of almost a hundred years of research in criminal justice, psychology, and law. Today, actuarial risk assessments are in wide use across many countries, helping judges make life-changing decisions in pretrial release, sentencing, and probation. Risk assessments can help reduce costs, racial disparity, and incarceration rates---and these benefits have already been realized in some jurisdictions \citep{berk2017impact}. However, some of the most widely used algorithms are secret, black-box models created by corporations. As a result, individuals affected by these algorithms cannot know how these decisions were made, or whether they were made in error. These problems resulted in various lawsuits over the last decade, and came to the fore in 2016, when investigative journalists from the nonprofit organization ProPublica claimed that the COMPAS black-box recidivism prediction model --- standing for \textbf{C}orrectional \textbf{O}ffender \textbf{M}anagement \textbf{P}rofiling for \textbf{A}lternative \textbf{S}anctions ---  was rife with racial bias \citep{LarsonMaKiAn16, freemanloomis}. 

Though ProPublica's findings were not validated \citep{Rudin19AgeofUnfairness, flores16, northpointeresponse}, the COMPAS scandal demonstrated the issues with for-profit, secret algorithms making decisions in the justice system---namely, possible violations of defendants' due process rights, difficulty in ensuring that the scores were calculated based on correct inputs, and the lack of independent fairness or performance guarantees. It highlighted the ways that systemic bias in data can be propagated into the future, and was symptomatic of growing public distrust in the algorithms that impact our daily lives \citep{nytoped, oneil2016weapons, nyt-computers-crim-justice}. 

To prevent errors, prevent due process violations, allow independent validation of models, and gain public trust, we must create interpretable and fair models. Fortunately, techniques for interpretable machine learning and theories of fairness have advanced considerably over the last few years. Multiple works have demonstrated that publicly available interpretable machine learning algorithms can perform as well as black-box machine learning algorithms \citep{ZengUsRu2017, angelino2018, loucaruana2013}. Moreover, high-dimensional data sets on criminal recidivism have become increasingly available. However, most machine learning papers are focused on algorithm construction and do not consider factors such as data quality or ease of computating model predictions, which are paramount for creating models that would be useful in practice. To our knowledge, there is only one prior work \citep{fairbydesign} that jointly considers interpretability, fairness, and predictive performance; however, it does not do so in a comprehensive way and focuses primarily on the design of a new algorithm. 

Beyond the problem of model optimization, various  methodological questions remain with existing risk assessment systems. First, existing systems---such as COMPAS (Correctional Offender Management Profiling System for Alternative Sanctions)  and LSI-R (Level of Service Inventory Revised)---are often used across states, or even countries, with only minor normalization \citep{LSI2017brochure, compas}. However, populations in different states can significantly differ because the data generation process is not the same, so applying the same model across states may not lead to the best possible performance. Second, empirical evidence indicates that the underlying probability distribution of recidivism has changed over time in multiple locations \citep{changingdist}. For instance, a significant shift in the age distribution---a key predictor in many recidivism prediction models---has been observed in New York \citep{totalarrestrate}.
Thus, rather than using a static model with uneven performance across districts, a better solution might be to algorithmically generate models, so that they can be trained for specific locations and retrained if recidivism distributions shift over time.

Using modern tools of both interpretable and black-box machine learning, we revisit the recidivism prediction problem. We define recidivism as a new charge that an individual is convicted for within a certain time frame: six months or two years. We find that (1) black-box models do not perform significantly better than interpretable models for any of the twelve recidivism problems we consider. (2) Interpretable models generally perform better than existing actuarial risk assessments. (3) Models do not generalize well across regions. (4) Only a small subset of the many proposed fairness definitions can be applied to regression problems and they vary across different models. We also note that existing techniques to enforce fairness generally require non-interpretable transformations, and therefore do not work well with interpretable models. 

This paper is structured as follows. Section \ref{Contribution} describes our contributions. Section \ref{Background} discusses the evolution of risk assessment in America, the current debate over risk assessments, and briefly reviews the machine learning literature on risk assessment. Section \ref{Data} describes the study's data sources. Section \ref{Methodology} discusses aspects of our methodology, including the prediction problems, problem setup, and the existing risk assessments we compare against. Section \ref{Baselines}  presents the performance of baseline, non-interpretable machine learning methods, while Section \ref{Interpretable} presents the performance of interpretable machine learning methods. Section \ref{Generalization} examines the generalization of recidivism prediction models across states. In Section \ref{Fairness}, we describe the selection of fairness metrics and assess the fairness of the interpretable models. In Section \ref{Discussion}, we discuss broader impacts and future lines of inquiry.

%%%%%%%%%%%%%%%%%%%%%%%%%%%%%%%

\section{Contribution} \label{Contribution}

Our main contribution is a set of interpretable, risk-calibrated linear models that perform approximately as well as--- sometimes better than---existing actuarial risk assessments, and predict specific crime types. Other important aspects of our contribution are as follows:

\begin{itemize}
    \item We consider multiple types of recidivism (\texttt{general}, \texttt{violent}, \texttt{drug}, \texttt{property}, \texttt{felony}, and \texttt{misdemeanor}) at two time scales (six-month, two-year) for a total of 12 prediction problems. 
    
    \item Our analysis was conducted on two criminal history data sets (one from Broward County, Florida, and the other from the state of Kentucky), which allowed us to understand variability in model performance across locations. We found that models do not generalize well between locations, and conclude that models should be trained on data from the location where they are meant to be used.
    
    % \item We discuss how our models satisfy fairness and interpretability criteria. To our knowledge, there are no prior machine learning studies on pretrial risk scoring that rigorously evaluate the combination of predictive performance, interpretability, and fairness.
    %beyond the fields of fair and interpretable machine learning, there are few papers that discuss both fairness and interpretability with the same attention as predictive performance. 
    
    \item The risk models trained as part of this study are interpretable, and could potentially be useful in practice after a careful, location-specific evaluation of their accuracy and fairness. 
    
    \item We provide an understanding of how to evaluate both interpretability and fairness in an important real application. The same type of analysis could be ported to financial lending decisions, hiring decisions, or any other type of high-stakes decisions that require an assessment of both interpretability and fairness.
\end{itemize}{}

Similar to \citet{ZengUsRu2017}, we use machine learning techniques optimized for interpretability, and address multiple prediction problems. This work is an improvement in the following ways. We use interpretable machine learning techniques to create risk scores representing probabilities of recidivism rather than making binary predictions---techniques which were not available at the time of publication for \citet{ZengUsRu2017}. We compare with COMPAS and the Arnold Public Safety Assessment (PSA), two models currently used in the justice system, whereas \citet{ZengUsRu2017} compared only with other machine learning methods. We use data obtained at the pretrial stage rather than at prison-release. Since many jurisdictions utilize prediction instruments to determine pretrial release, this better aligns with the use cases of risk scores. Our data come from two locations, and include more detailed information than in \citet{ZengUsRu2017}, and are more recent than 1994. Finally, models are assessed for multiple definitions of fairness in addition to performance. 

%%%%%%%%%%%%%%%%%%%%%%%%%%%%%%%
\section{Background} \label{Background}

Algorithmic risk assessment dates back to the early 1900s \citep{history}, and is used today at various stages of the criminal justice system, such as at pretrial, parole, probation, or even sentencing. In this work, we focus on forecasting recidivism at the pretrial stage. Though some states have implemented their own tools (Virginia, Pennsylvania, Kentucky), many utilize systems produced by companies, non-profits and other organizations \citep{kehl2017}. These externally-produced risk assessments and some of the jurisdictions that utilize them include COMPAS (Florida, Michigan, Wisconsin, Wyoming, New Mexico), the Public Safety Assessment  (New Jersey, Arizona, Kentucky,\footnote{Kentucky created and implemented their own tool in 2006 but transitioned to the Arnold PSA in 2013.} Phoenix, Chicago, Houston), LSI-R (Delaware, Colorado, Hawaii), and the Ohio Risk Assessment System \citep{psaabout, ohiorisk, epic}. The United States is not alone in using actuarial risk assessments. Canada uses the Static-2002 to assess risk of violent and sexual recidivism \citep{hanson2003notes}; the Netherlands uses the Quickscan to assess static and dynamic risks of recidivism  \citep{tollenaar2013method}; the U.K. uses the Offender Group Reconviction Scale to predict reoffense while on probation \citep{howard2009ogrs}.

%%%%%%%%%%%%%%%%%%%%%%%%%%%%%
\subsection{The Debate over Risk Assessments} \label{Debate}

Since the inception of actuarial risk assessments, there has been debate over whether they should be used in the criminal justice system at all. Proponents claim that statistical models reduce overall violence levels and ensure the most efficient use of treatment and rehabilitative resources by helping judges identify the individuals that are truly dangerous. A large body of evidence appears to support this claim. Various studies have shown that statistical models are more accurate than human experts \citep{dawes1989clinical, grove1996comparative}. Others have shown that a small percentage of  individuals commit the majority of crimes \citep{wolfgang1987delinquency, sherman2007power, milgram2014ted}, indicating that correctly identifying dangerous individuals could lead to substantial decreases in violence levels.
Proponents also claim that risk assessments are instrumental to reducing racial/economic disparity, allocating social services, and reducing mass incarceration \citep{james2018congressional}. In particular, some jurisdictions have adopted risk assessments at the pretrial stage to replace cash bail, since cash bail is widely viewed as biased against poor defendants \citep{zweig2010bail, desmarais2019rebuttal}. 

In practice, reducing overall violence levels, mass incarceration, and racial/economic disparity through actuarial risk assessment is complex \citep{ludwig21fragile}. Critics have argued that as recidivism prediction models always rely on racially-biased features such as arrest records, actuarial risk assessment will only exacerbate racial and socioeconomic disparity, and should therefore be abolished \citep{civilrightsstatement, pji2020}. In a well-known incident, ProPublica claimed that COMPAS was biased against African-Americans because there was a disparity in false positive rates and false negative rates between African-Americans and Caucasians \citep{AngwinLaMaKi16}. Follow-up research showed that this bias was likely a property of the data generation process rather than the COMPAS model, and that even a model that only relied on age showed a similar disparity in false positives and false negatives \citep{Rudin19AgeofUnfairness}. Actuarial risk assessment might be vulnerable to feature bias, but it is important to remember that other parts of the court system (such as bail and sentencing guidelines for judges) are not immune to feature bias either---they also use criminal history and arrest records. 
Similarly, in one of the first large-scale empirical studies, \citet{mstevenson} showed that in Kentucky, the use of the Arnold PSA seemingly increased disparity between whites and blacks at pretrial release. Because the risk scores were applied differently by judges in different counties, it seemed that white people benefited more than black people in terms of pretrial release numbers---but within the same county, white and black defendants saw similar \textit{increases} in release. Thus, rather than eliminating the use of risk scores, using them uniformly across counties may have made risk assessments more fair across the state, and could have reduced overall incarceration. 

Others have argued that a fundamental flaw with risk assessments is that their simple labels obscure the true uncertainty behind their predictions \citep{nytoped}. This may be true for currently used risk assessments, but merely underscores the necessity for researchers to develop models that \textit{do} quantify uncertainty. While actuarial risk assessments are not perfect, we must remember that in the absence of risk assessments, judges can only rely on their intuition---and human intuition has been shown to be less reliable than statistical models \citep{dawes1989clinical, grove1996comparative, desmarais2019rebuttal, skeem}.  

Another problem is that some of the most widely used risk prediction algorithms are for-profit and secret. For instance, while COMPAS's guidelines are published and validation studies have been performed, the full forms of the models are not available and some of the validation studies do not conform to standards of open science because they do not publish the validation data \citep{pattern}, thus yielding concerns over due process rights. In the 2017 Wisconsin Supreme Court case, Loomis v. Wisconsin, Loomis challenged  the use of the proprietary risk prediction software, COMPAS, on the grounds that this violated his due process and equal protection rights  \citep{freemanloomis}. Yet today, there are plenty of equally accurate, transparent risk prediction tools that publish their guidelines and full models. See Table \ref{table:risk_assessments_comparison} in the Appendix for examples. In this article, we compare against the Arnold PSA, an interpretable and publicly available tool which is used in multiple jurisdictions.

There is also a general fear that the use of risk assessments could lead to situations similar to those depicted in the movie, ``Minority Report.'' In Minority Report, individuals were punished \textit{before} they committed a crime based on oracles' visions of the future. However, one of the major principles common to American criminal justice texts \citep{sentencingbook, criminalsourcebook} is that individuals should be punished based on the crimes they committed in the past. This illustrates why risk assessments have played only a minor role in sentencing. In reality, risk prediction tools are most heavily used in bail, parole, and social services decisions.

Risk scores will not solve everything, but abolishing risk assessment without a useful alternative plan will not solve the problems above either. Reducing feature bias requires generations of community investment; jurisdictions must train judges on how to use risk scores; and communities must provide treatment resources for those deemed high risk. Risk assessments and other evidence-driven practices can be an important part of this solution. In the most recent revision of the Model Penal Code, the American Law Institute has supported giving people shorter prison terms or sending them to the community through the use of risk assessment tools \citep{starr2015riskassessment, modelpenalcode2017}. By providing simple and interpretable risk scores, we hope to mitigate the possibility that risk assessments are miscomputed, and enable judges and defendants to fully understand their scores.

%%%%%%%%%%%%%%%%%%%%%%%%%%
\subsection{Black-box and Interpretable Machine Learning for Predicting Criminal Recidivism} 

% Current machine learning work predicting recidivism using black-box models
There is an abundance of past research on using machine learning methods to predict criminal recidivism. However, many of these studies utilize black-box, non-interpretable models, and only optimize for predictive performance. For instance, \citet{neuilly2011predicting} used random forests to predict homicide offender recidivism. Other black-box models applied to this problem include stochastic gradient boosting  \citep{friedman2002stochastic},  neural networks \citep{paloscayneural}, and ensemble methods \citep{singh21riskassess}.  

% discuss interpretable work
In comparison, there is relatively little work using interpretable machine learning techniques to forecast recidivism, and there is not a consensus on how interpretability should be defined in this domain.  \citet{berk2005developing}  used classical decision trees to build a simple screener for forecasting domestic violence for the Los Angeles Sheriff’s Department.  \citet{goelfrisk} created a simple scoring system by rounding logistic regression coefficients, which helped address stop-and-frisk for the New York Police Department. \citet{ZengUsRu2017} was the first work using modern machine learning methods that globally optimized over the space of sparse linear integer models to predict criminal recidivism. Despite the range of interpretable models that have been applied to the criminal recidivism problem, a common thread among these works is that simple, interpretable models can do just as well as black-box models, and better than humans. For instance, \citet{angelino2018} found that COMPAS shows no benefit in accuracy over very simple machine learning models involving age and criminal history. \citet{skeem} showed that algorithms outperformed humans on predicting criminal recidivism in three data sets, and demonstrated that the performance gap was especially large when abundant risk factors were considered for risk prediction.

The approaches outlined above achieved interpretability through training models with interpretable forms. Another major approach is \textit{post-hoc explainability}, in which a simpler model provides insights into a black-box model. However, post-hoc explanations are notoriously unreliable, or are not thorough enough to fully explain the black-box model \citep{Rudin18}. Additionally, there seems to be no clear benefit of black-box models over inherently interpretable models in terms of prediction accuracy on the criminal recidivism problem \citep{ZengUsRu2017,tollenaar2013method}. Thus, for a high-stakes problem such as predicting criminal recidivism, we choose not to utilize these methods. 

In fact, there have been cases in criminal justice where post-hoc explanations led to incorrect conclusions and pervasive misconceptions about what information some of the most common recidivism models use. The 2016 COMPAS scandal---where ProPublica reporters accused the proprietary COMPAS risk scores of an explicit dependence on race \citep{AngwinLaMaKi16}---was caused by a flawed,  post-hoc explanation of a black-box model. In particular, ProPublica reasoned that if a post-hoc explanation of COMPAS depended linearly on race, then COMPAS depended on race, even after controlling for age and criminal history. However, as  \citet{Rudin19AgeofUnfairness} demonstrated, just because an explanation model depends on a variable \textit{does not} mean that the black box model depends on that variable. Thus, ProPublica's reasoning was incorrect. In particular, this analysis found that COMPAS does not seem to depend linearly on some of its input variables (age), and does not seem to depend on race after conditioning on age and criminal history variables. Criminologists have also criticized the ProPublica work for other reasons \citep{flores16}. Despite the flaws in the ProPublica article, it is widely viewed as being a landmark paper on fairness in machine learning.  

A notable advantage of interpretable modelling for criminal justice is that some interpretable models allow a decision-maker to incorporate factors not in the database in a way that black-box models cannot. For instance, scoring systems---linear models with integer coefficients---place all of the model inputs onto the same scale: every input receives a number of points. The points of each factor in the model provide clarity on how important each input is relative to the others. 

%%%%%%%%%%%%%%%%%%%%%%%%%%%%%%%%%%%%%%%%%%%%%%%%%%%%%%%%%%%%%%%%%%%%%%%%%%%%%%%%%%%%%%%%%%%%%%%%
\subsection{Fair Machine Learning} \label{Fair_ML}

Fairness is a crucial property of risk scores. As such, the recidivism prediction problem is a key motivator for many of these works. However, recidivism prediction is rarely the primary focus of fairness papers. Many of these papers seek to make theoretical contributions by proposing definitions of fairness and creating algorithms to achieve these definitions, using recidivism prediction as a case study \citep{hardt2016eqodds, agarwal2018fairbinaryclass}. Others have proven fairness impossibility theorems, showing when different fairness constraints cannot be achieved simultaneously. For instance, the two fairness definitions at the heart of the debate over COMPAS' fairness (calibration and balance for positive/negative class) cannot be achieved simultaneously in nontrivial cases. However, by placing relaxations on the conditions, the fairness definitions can be \textit{approximately} satisfied simultaneously. \citep{kleinberginherent, pleiss2017calib}. These theorems show that many fairness  definitions directly conflict, so there cannot be a single universal definition of fairness \citep{kleinberginherent, binnsfair, verma2018fairnessreview}. Moreover, there is often a trade-off between performance and fairness \citep{berk2017, berk2019, corbettdavies2017faircost}. The emerging consensus is that any decision about the ``best'' definition of fairness must rely heavily on model characteristics and domain-specific expertise. 

The question of what should count as fair in criminal recidivism prediction can be answered by discussion among ethicists, judges, legislators, and stakeholders in the criminal justice system. Existing American anti-discrimination law provides a general legal framework for addressing this question. Under Title VII of the Civil Rights Act of 1964, there are two theories of liability: disparate impact and disparate treatment \citep{barocasdispimpact2016}. In this article, we use the definitions of fairness from the field of fair machine learning, as they apply directly to machine learning models and are more specific than the general legal guidelines of disparate impact and treatment. Moreover, some of the definitions of fairness proposed by the field of fair machine learning community are inspired by these guidelines.  See  \citet{corbett-daviesmeasure} for a detailed discussion of the relationship between algorithmic definitions of fairness and economic/legal definitions of discrimination. 

%%%%%%%%%%%%%%%%%%%%%%%%
\section{Data} \label{Data}

In this study, we used criminal history data sets from Broward County, Florida, and the state of Kentucky, allowing us to analyze how models perform across regions. The Broward County data set consists of publicly available criminal history and court data from Broward County, Florida. This data set consists of the full criminal history, probational history, and demographic data for the 11,757 individuals who received COMPAS scores at the pretrial stage from 2013-2014  \citep[as released by ProPublica][]{AngwinLaMaKi16}. The probational history was computed from public criminal records released by the Broward Clerk’s Office. Though the full data set includes 11,757 individuals, this analysis includes only the 1,954 for which we could also compute the PSA. We processed the Broward data using the same methods as  \citet{Rudin19AgeofUnfairness}. From the processed data, we computed various features such as number of prior arrests, prior charges, prior felonies, prior misdemeanors, etc.  

The Kentucky pretrial and criminal court data was provided by the Department of Shared Services, Research and Statistics in Kentucky. The data came from two systems: the Pretrial Services Information Management System (PRIM) and CourtNet. PRIM data contain records regarding defendants, interviews, PRIM cases, bonds, etc., that were connected with the pretrial service interviews conducted between July 1, 2009 and June 30, 2018. CourtNet data provide further information about cases, charges, sentences, dispositions, etc. In total, the Kentucky data set consists of over 25 million tuples. When constructing features from the Kentucky data set, we computed features that were as similar as possible to the Broward features (e.g., prior arrests, prior charges with different types of crimes, age at current charge) in order to compare models between the two regions. There are several features from Broward data which could not be computed from the Kentucky data, such as ``age at first offense'' and ``prior juvenile charges.'' A limitation of the Kentucky data set is that the policies governing risk assessments changed over the period when the data was gathered, possibly impacting the consistency of the data collection. 

A difference in the data processing between the two data sets is that when constructing prediction features and predictive labels, we considered non-convicted charges in the Broward data, but considered convicted charges in the Kentucky data. The reason for this choice is sample size. The processed Broward data contains only 1,954 records, and limiting the scope to convicted charges would yield only 1,297 records. The use of convicted versus non-convicted charges between the two regions might explain some discrepancies in the results in Section \ref{Generalization}, where we discuss the generalization of recidivism prediction models between states. Note that many models currently implemented within the justice system rely on non-convicted charges such as counts of prior arrests, but for the applications such as bail and parole, the use of non-convicted charges could be problematic---it holds individuals accountable for crimes that they may not have committed. 

Please refer to the Section \ref{broward-processing} and \ref{kentucky-processing} in the Appendix for more details on data processing and a full list of features.

%%%%%%%%%%%%%%%%%%%%%%%
\section{Methodology} \label{Methodology}

% comparing against COMPAS and Broward
Throughout our analysis, we compare with two tools that are currently used to predict recidivism in the U.S. justice system: COMPAS (Correctional Offender Management Profiling for Alternative Sanctions) and the Arnold PSA (Public Safety Assessment, created by Arnold Ventures, which was previously named the Laura and John Arnold Foundation). Although we would have liked to compare against more assessment tools, many of them use data that are not publicly available, or are owned by for-profit companies that do not release their models. For a detailed discussion of the other risk assessments we considered and the features we were missing, please consult Section \ref{other-tools} in the Appendix. 

More specifically, we compared our models against the Arnold PSA's New Criminal Activity (NCA) and New Violent Criminal Activity (NVCA) scores on the \texttt{general} and \texttt{violent} recidivism problems, respectively. Note that the time-frames and labels for prediction are important here, and our choices distinguish this study from past works on recidivism prediction. Let us explain the time-frames next.

It is important that we chose \textit{fixed} time-frames for prediction, in our case, two years or six months past the current charge dates. In reality, the scores are used to assess risks during the pretrial period. However, there is a huge amount of variation in pretrial periods, which can span a few days to a few years: the average pretrial time span in Kentucky is 109 days, and could last upwards of 3-4 years. Since the pretrial period depends on the jurisdiction, we chose to fix time spans so that the models do not depend on the policy used for determining how long the pretrial period would be. That way, the risk calculations we produce depend mainly on the inherent characteristics of the individual, rather than the length of the pretrial period, which is potentially a characteristic of the jurisdiction. Also, this way, individuals with the same propensity to commit a new crime within six months (or two years) are given identical risk scores, even if they have different expected time periods until their respective trials. The six-month time span represents an approximate length of pretrial period. The two-year time span provides more balanced labels, since two years provides more time to commit crimes than six months. Additionally, our evaluation metric is AUC, which is a rank statistic, and considers relative risk rather than absolute risk; that is, an individual who actually commits crimes within two years of their current charge date should be ranked higher than an individual who does not. The relative risk within the two-year time-frame is related to the relative risk for other shorter or longer time-frames, allowing these models to potentially generalize to varying pretrial time-frames.

Another important aspect of our prediction problems is the \textit{definition of recidivism} we chose. We predict the occurrence of a \textit{convicted} charge within six months/two years for Kentucky. In other words, we would like to predict whether someone will be arrested, within six month or two years from their current charge, for another crime that they were later convicted for. This definition potentially alleviates a due process concern: if we instead include non-convicted charges, our models might be more likely to predict who will be arrested, which is tied to policing practices. For Broward, where we did not have conviction information for later charges, we predicted \textit{any charge} within six months/two years, which is the typical approach to recidivism prediction.

In Broward, we directly computed Arnold PSA scores, as the Arnold PSA is publicly available. The features used by the Arnold PSA are provided in Tables \ref{table:arnold_psa_nca} \& \ref{table:arnold_psa_nvca} in the Appendix. For Kentucky, we used the unscaled Arnold PSA scores that came with the data set, because those are what are reported to the judges in Kentucky. We compared against COMPAS' Risk of General Recidivism and Risk of Violent Recidivism risk scores on the two-year \texttt{general} and two-year \texttt{violent} prediction problems, respectively. Note that both models are designed to predict recidivism within two years. The COMPAS suite is proprietary, but COMPAS General and Violent scores were provided with the Broward County data set. We do not compare against COMPAS on the Kentucky data set. The COMPAS General and COMPAS Violent scores appear to have been developed for a parole population \citep{compas}, but have been applied for pretrial decisions in Broward. In this study, we consider the COMPAS scores for the outcomes they were actually applied for (pretrial decisions), rather than the outcomes they were developed for (parole decisions). 

In Sections \ref{Baselines} and \ref{Interpretable}, we compare the performance of black-box and interpretable algorithms on the Broward and Kentucky data sets. We caution readers against comparing an algorithm's performance in Broward with its performance in Kentucky. An algorithm's differences in performance between the data sets could be attributed to the many differences between the two regions. For instance, the Broward data set is at the county level while the Kentucky data set is at the state level. As the Kentucky data is at the state level, it embeds diverse information about 120 counties (e.g., demographics, legislation, culture, local policing practices). Thus, in Sections \ref{Baselines} and \ref{Interpretable}, the comparisons between baseline models and interpretable models are conducted \textit{within} each data set. In Section \ref{Generalization}, we discus in detail the regional differences between  Broward County and Kentucky, and present a set of experiments that illustrate model performance gaps resulting from these regional differences. 

%%%%%%% prediction labels
\subsection{Prediction Labels} \label{Labels}

In addition to two-year \texttt{general} recidivism and two-year \texttt{violent} recidivism---the two types of criminal recidivism considered by COMPAS and the PSA---we computed recidivism prediction labels specific to various crime types, such as \texttt{property}, \texttt{drug} related recidivism and recidivism with \texttt{felony} or \texttt{misdemeanor} level charges. For clarity, we apply the \(\texttt{typewrite}\) font to indicate prediction tasks. Note that an individual could have multiple positive labels, indicating that the newly committed crime involves multiple charge types. We defined recidivism as a recorded charge within a certain time frame. Out of all the possible recidivism prediction tasks we considered, we selected the six most balanced: \texttt{general}, \texttt{violent}, \texttt{drug}, \texttt{property}, \texttt{felony}, and  \texttt{misdemeanor}. To investigate the effect of temporal scale on predictive performance, we generated these six tasks using the time windows \texttt{two-years} and \texttt{six-months} after the current charge date (or release date, if the individual went to prison for their current charge), for a total of twelve tasks. The summary of prediction tasks and the base rate of recidivism for each task is provided in Table \ref{table:prediction_labels}.

\begin{table*}[htp]
\begin{center}
\tabcolsep=0.2cm
\scriptsize
\renewcommand{\arraystretch}{1.5}
\begin{tabular}{|c|c|c|c|c|} \hline
    & \multicolumn{1}{c}{\textbf{Kentucky}} & & \multicolumn{1}{c}{\textbf{Broward}} &  \\ \hline
    \textbf{Labels} & \textbf{Two Year} $P(y_i = 1)$ & \textbf{Six Month} $P(y_i = 1)$ & \textbf{Two Year} $P(y_i = 1)$ & \textbf{Six Month} $P(y_i = 1)$ \\ \hline
    \texttt{General} & 20.4\% & 5.7 \% & 45.5\% & 21.8 \% \\ \hline
    \texttt{Violent} & 3.4\% & 0.7\% & 21.0\% & 8.4\%\\\hline
    \texttt{Drug} & 8.7\% & 2.0\% & 9.3\% & 4.0\% \\\hline
    \texttt{Property} & 3.9\% & 0.9\% & 9.0\% & 5.0 \% \\\hline
    \texttt{Felony} & 9.6\% & 2.4\% & 17.6\% & 8.9 \% \\\hline
    \texttt{Misdemeanor} & 15.6\% & 3.9\% & 27.2\% & 12.5 \% \\ \hline
    \multicolumn{5}{l}{\footnotesize Note: \(y_i = 1\) if the defendant had the corresponding type (\texttt{general, violent, drug etc.)}} \\
    \multicolumn{5}{l}{\footnotesize of charge within two years (resp. six months) from current charge/release date}
\end{tabular}
\end{center}
\caption{Label distributions for Broward and Kentucky.}
\label{table:prediction_labels}
\vspace{-4mm}
\end{table*}

\subsection{Problem Setup} \label{Setup}

Due to the binary nature of recidivism tasks, we approached these prediction problems as binary classification problems, but do not binarize the final predicted probabilities/scores of the machine learning models for the following reasons. First, existing risk scores are usually nonbinary. For instance, the Arnold PSA's unscaled New Criminal Activity (NCA) score takes integer values from 0 through 13, while the COMPAS Risk of Recidivism and Risk of Violent Recidivism scores take on integer values from 1 through 10 \citep{compas, psaabout}. Second, we want to create more nuanced risk scores both by predicting highly specific types of recidivism, in addition to coarser categories like general recidivism, and by presenting nonbinary scores which reflect a range of risk values. Since the predictions are nonbinary, we use Area Under the Curve (AUC) as our evaluation metric. This decision also impacts the fairness metrics we assess, which we discuss in Section \ref{Fairness}. We applied nested cross validation process to train the models. Please refer to Section \ref{nestedcv} in the  Appendix for the details.

%%%%%%%%%%%%%%%%
\section{Baseline Machine Learning Methods}  \label{Baselines}

To provide a basis of comparison for the interpretable models presented in Section \ref{Interpretable}, we evaluated the performance of six common, non-interpretable machine learning methods in this section. Baseline models and descriptions are provided below. The tuned hyperparameters and packages used for each problem are provided in Section \ref{hyperparameters} in the Appendix. 
\begin{itemize}
    \item \textbf{$\ell_2$ Penalized Logistic Regression}: To prevent overfitting, there is an $\ell_2$ penalty term on the sum of squared coefficients in the loss function for logistic regression. Although this method produces linear models, we consider $\ell_2$-penalized logistic regression  to be non-interpretable because if the number of input features is large, there could be a large number of nonzero terms in the model. 
    
    \item \textbf{$\ell_1$ Penalized Logistic Regression}:  To prevent overfitting, there is an $\ell_1$ penalty term on the sum of absolute values of coefficients in the loss function for logistic regression. This algorithm creates sparser models than \(\ell_2\) penalized logistic regression. Notice that the sparsity of the model depends on the magnitude of the penalty and must be balanced with consideration of prediction performance. In our experiments, $\ell_1$ models with Broward data were sparse yet maintained good predictive performances. However, the best $\ell_1$ models with Kentucky data still had too many features, which made it difficult to interpret the results. Therefore,  we classified $\ell_1$-penalized logistic regression as a non-interpretable algorithm.
    
    \item \textbf{SVM with a Linear Kernel \citep{vapnik1964svm}:} 
    An algorithm that outputs a hyperplane that separates two classes by maximizing the sum of margins between the hyperplane and all points. Incorrectly classified points are penalized. Although SVM with linear kernel yields a linear model, the concerns with $\ell_1$ and $\ell_2$ penalized logistic regressions apply here as well: the number of nonzero terms could be large, making it difficult to interpret the model. 
     
    \item \textbf{Random Forest \citep{breiman1984classification}}: An ensemble method that combines the predictions of multiple decision trees, each of which is trained on a bootstrap sample of the data. The implementation we use combines individual trees by averaging the probabilistic prediction of each tree. Random Forest is usually considered a black-box classifier because it is difficult to understand the individual contribution of each feature, which can be found in many trees, and the joint relationship between features.
    
    \item \textbf{Boosted Decision Trees \citep{freund1997decision}}: An ensemble method where a sequence of weak classifiers (decision trees) are fit to weighted versions of the data. Similar to random forest, boosted decision trees produce black-box models because it is difficult to understand the joint relationships of the features. We use the XGBoost implementation \citep{xgboost}.
\end{itemize}

\subsection{Broward Baseline Results} \label{B_baselines}

The performances of baseline algorithms on the Broward data are visualized in Figure \ref{fig:comb_baseline}; details are presented in Table \ref{table:broward_baseline_pred_res} in Section \ref{app-tables} from Appendix. We noticed that in the two-year prediction problems, no algorithm consistently performs better than the others. Simple linear models can even outperform black-box models in some prediction problems. For instance, in the two-year prediction problems, $\ell_2$-penalized logistic regression  and  LinearSVM tie in performance for the \texttt{general} recidivism prediction. XGBoost performs the best in \texttt{violent} and \texttt{property} prediction problems. $\ell_1$-penalized logistic regression has the best performance in \texttt{drug} and \texttt{felony} prediction tasks, while $\ell_2$-penalized logistic regression has the best performance in \texttt{misdemeanor} recidivism prediction. The largest performance gap is 5.1\%, from \texttt{property} recidivism prediction. In the six-month prediction problems, we see the same phenomenon that no single model dominates the others in performance. Overall, the performance gaps across baseline models for the \texttt{general}, \texttt{felony}, and \texttt{misdemeanor} prediction tasks are small, while other prediction problems have larger gaps.

\begin{comment}
\begin{figure}[htbp]
    \centering
    \caption{Visualizations of Broward baseline results from Table \ref{table:broward_baseline_pred_res}. Within each prediction problem, all algorithms performed similarly. No single algorithm consistently outperformed others.}
    \includegraphics[width=0.8\linewidth]{figures/fl_baseline.png}
    \label{fig:broward_baseline_pred_res}
\end{figure}
\end{comment}

\subsection{Kentucky Baseline Results} \label{K_baselines}

\begin{figure}[htbp]
    \centering
    \caption{Visualizations of Broward and Kentucky baseline results.}
    \includegraphics[width=\linewidth]{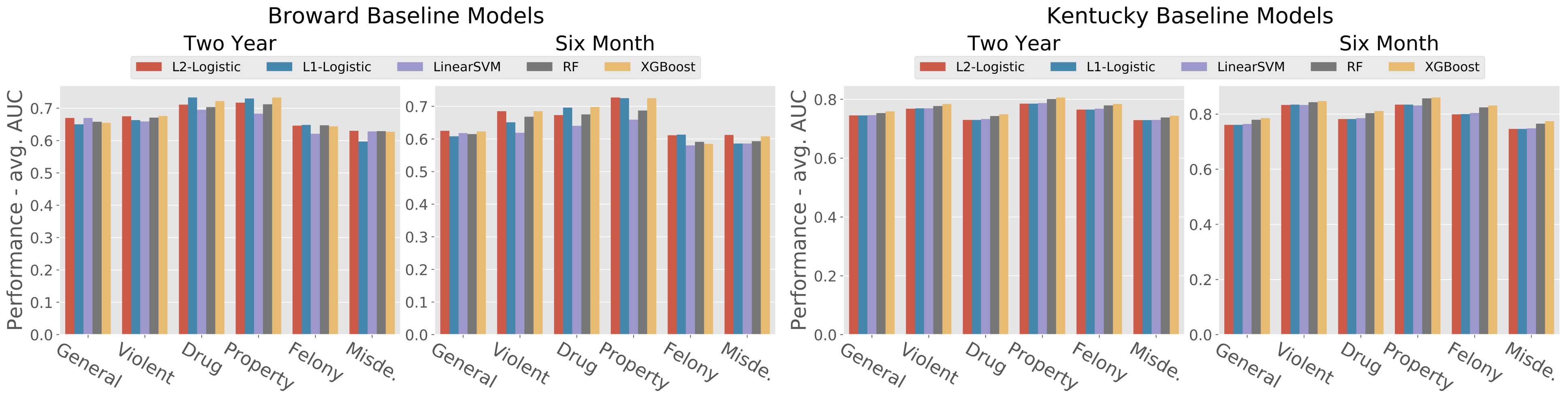}
    \label{fig:comb_baseline}
\end{figure}

The performances of baseline algorithms on the Kentucky data are visualized in Figure \ref{fig:comb_baseline}; details are presented in Table \ref{table:kentucky_baseline_pred_res} in Section \ref{app-tables}. We noticed that complex and nonlinear baselines perform slightly better than linear models, potentially due to the larger size of the Kentucky data set  (1,956 records in Broward versus about 250K records in Kentucky). In particular, Random Forest and XGBoost uniformly perform slightly better than all the other algorithms on all prediction tasks, over both time periods we examined. XGBoost performs the best on all tasks. However, performance gaps, across all prediction problems and in both time frames, are very small. Thus, we conclude that all the baseline algorithms perform similarly over the Kentucky data set. On Kentucky, all algorithms perform slightly better on the six-month recidivism period than on the two-year period.

\begin{comment}
\begin{figure}[htbp]
    \centering
    \caption{Visualizations of Kentucky baseline results from Table  \ref{table:kentucky_baseline_pred_res}. Random Forest and XGBoost consistently perform better than other models, but the results are similar across all models.}  
    \includegraphics[width=0.8\linewidth]{figures/ky_baseline.png}
    \label{fig:KY_baselines_vis}
\end{figure}
\end{comment}

\textbf{Summary of Baseline Models' Results:} We found that all baseline machine learning algorithms performed similarly across recidivism problems for the Kentucky data set. We also found that models performed better on the six-month prediction problems than on the two-year problems on Kentucky data, but not on Broward data. These findings will be discussed throughout the following subsections.

%%%%%%%%%%%%%%%%%%%%%%%%
\section{Interpretable Machine Learning Methods} \label{Interpretable}

For recidivism prediction, we considered several different types of interpretable machine learning methods with different levels of interpretability, ranging from scoring systems  to decision trees, to additive models. Since the Burgess model in 1928 \citep{burgess1928factors}, recidivism risk assessments have traditionally been scoring systems, which are sparse linear models with positive integer coefficients. A scoring system can be visualized as a simple scoring table or set of figures. There have only recently been algorithms designed to optimally learn scoring systems directly from data, without manual feature selection or rounding. Scoring systems have several advantages: they allow an understanding of how variables act \textit{jointly} to form the prediction; they are understandable by non-experts; risks can be computed without a calculator; and they are consistent with the form of model that criminologists have built over the last century, where ``points'' are given to the individual, and the total points are transformed into a risk of recidivism. External information, such as risk factors that are not in any database, can be more easily incorporated into the risk score: it is much easier to determine how many points to assign to a new factor if the points are integer-valued for the known risk factors. For instance, we could choose to subtract three points for drug treatment, to counteract four points of past drug-related arrests. 

While scoring systems appear to be the accepted standard for interpretability in the domain of criminal justice, imposing the constraints of linearity, sparsity, and integrality of coefficients could potentially be strong enough to reduce accuracy. Thus, we also consider modern algorithms that satisfy a subset of the conditions of interpretability: sparsity in features, ability to visualize/explain any variable interactions, linearity, integer coefficients.
Specifically, we tested four interpretable machine learning algorithms: Classification and Regression Trees (CART), Explainable Boosting Machine (EBM), Additive Stumps, and RiskSLIM (Risk-Calibrated Supersparse Linear Integer Models). Algorithm specifics are articulated below and the tested hyperparameters are provided in the Appendix. We also tested two existing risk assessments---the Arnold PSA and COMPAS---and compared their performances to both baseline and interpretable machine learning models. 
\begin{itemize}
    \item \textbf{Classification and Regression Trees (CART) \citep{breiman1984classification}}: A method to create decision trees by continuously splitting input features on certain values until a stopping criterion is satisfied. CART constructs binary trees using the feature and threshold that yields the largest information gain at each node. We constrain the maximum depth of the tree to ensure that it does not use too many features. CART models are nonlinear. They cannot be written as scoring systems, but can be written as logical models.   

    \item \textbf{Explainable Boosting Machine (EBM) \citep{loucaruana2013}}: An algorithm that uses boosting to train Generalized Additive Models with a few interaction terms ($\mathrm{GA^2Ms}$). The contribution by each feature and feature interaction pair can be visualized. The models are interpretable and modular, thus editable by experts. The models are generally not sparse, and cannot be written as scoring systems. 
    
    \item \textbf{RiskSLIM \citep{UstunRu2017KDD,UstunRuRiskSlimJMLR19}}: An algorithm that generates sparse linear models with integer coefficients that have risk-calibrated probabilities. The models generated by RiskSLIM have form similar to that of models used in criminal justice over the last century. 
   
    \item \textbf{Additive Stumps}: An interpretable variation on $\ell_1$-penalized logistic regression: for each  feature, we generate multiple binary stumps (defined in Section \ref{Stumps}), and apply $\ell_1$-penalized logistic regression to these stumps. Ideally, the features will have monotonically increasing (or decreasing) contributions to the estimated probability of recidivism. Models constructed using this method generally use fewer features than those constructed with vanilla $\ell_1$-penalized logistic regression. These models are flexible and nonlinear. These models also cannot be written as scoring systems because they are not sparse in the number of nonlinearities. 
    
    \item \textbf{Arnold PSA \citep{psaabout}}: A widely used, publicly available, interpretable risk assessment system that consists of three scores: New Criminal Activity (NCA), New Violent Criminal Activity (NVCA), and Failure to Appear (FTA). We compare against the NCA for the \texttt{general} recidivism problem, and against the NVCA for the \texttt{violent} recidivism problem, on both two-year and six-month time scales. The NCA has 7 factors, while the NVCA has 5 factors.
    
    \item \textbf{COMPAS \citep{compas}}: A widely used risk assessment system that consists of several scores, including the three that we study: Risk of General Recidivism (COMPAS General), Risk of Violent Recidivism (COMPAS Violent), and Risk of Failure to Appear. We compare against the COMPAS General score for the two-year \texttt{general} recidivism problem, and compare against the COMPAS Violent score for the two-year \texttt{violent} problem.
\end{itemize}

\subsection{Preprocessing Features into Binary Stumps} \label{Stumps}

We performed a data preprocessing technique for two of the interpretable machine learning algorithms: RiskSLIM and Additive Stumps. This technique consists of transforming all original features into binary stumps using Equation \ref{eq:stumps}. Preprocessing the features into stumps allows us to include nonlinear interactions between the features (e.g. age, criminal history) and labels. It also allows us to visualize each Additive Stumps model as a set of monotonically increasing (or decreasing) curves. Formally, stumps are binary indicators, which are created by splitting features at pre-specified thresholds. For a feature $X^{(j)}$, and a set of threshold values $K \in \mathbb{R}$, we generate \textbf{decreasing stumps} $S_{k}^{(j)}$ for all $k \in K$ as follows:
\begin{equation} \label{eq:stumps}
S_{k}^{(j)} = \begin{cases}
                      1, &\text{for } X^{(j)} \leq k\\
                      0, &\text{else} 
                    \end{cases}
\end{equation}
We can generate \textbf{increasing stumps} analogously by substituting $\geq$ for $\leq$ in the definition above. The rationale behind the naming convention is as follows. Linear models constructed from increasing (respectively, decreasing) stumps have the nice property that if one sums the contribution from all stumps corresponding to a fixed original feature, i.e.,  $f(X^{(j)}) = \sum\limits_{k \in K} c_k S_k^{(j)}$ for the feature $X^{(j)}$, and the coefficients $c_k$ are mostly non-negative \footnote{For decreasing (respectively increasing) stumps, if the coefficient for the largest (respectively smallest) stump is negative, the function $f$ will still be monotonic because the negative value will be subtracted from all values of the remaining stumps}, the resulting function $f(X^{(j)})$ is monotonically increasing (respectively decreasing), which is desirable for interpretability. 

More concretely, the ``age\_at\_current\_charge'' feature ranges from 18 to 70 in our data. For all age-related features, we construct decreasing stumps for $k = \{18, 19, ..., ,60\}$. We chose decreasing stumps for age features because based on past studies \citep{Rudin19AgeofUnfairness, StevensonSl18} and criminological theory \citep{changingdist, bindler2018agedistlondon, bushway2007inextricable}, the probability of recidivism decreases with age. On the other hand, intuitively, the probability of recidivism should increase as criminal history increases. Thus, we construct increasing stumps for the remaining features, which relate to criminal history. 

To select a collection of stumps for the RiskSLIM and Additive Stumps model, we selected threshold values for all features by examining each feature visualization from EBM and choosing the threshold values that correspond to sharp drops in the predicted scores. 

%%%%%%%%%%%%%%%%%%%%%%%%%%%%%%%
\subsection{Broward Prediction Results for Interpretable Models} \label{B_interpretable}

Figure \ref{fig:comb_interpretable} show the results of interpretable models on the Broward data set; details are presented in Table \ref{table:broward_interpretable_pred_res} in Section \ref{app-tables} from Appendix. For all prediction problems in both two-year and six-month prediction periods that we examined, we observed that CART consistently performed worse than other algorithms. Additive Stumps and EBM performed similarly on all the prediction tasks and outperformed other models, including the Arnold PSA and COMPAS, on most of the prediction tasks. The performances of the best interpretable models are very similar to that of the best baseline models---this is true for each of the prediction problems we considered. The AUC gaps between the best interpretable models and best baseline models for all two-year prediction tasks range from 0.3\% to 1.7\% in absolute value, and range from 0.2\% to 2.6\% for six-month prediction tasks. Prediction gaps from all other problems are smaller than 1\%.

\begin{comment}
\begin{figure}[htbp]
  \centering
  \caption{Visualizations of Broward interpretable results from Table  \ref{table:broward_interpretable_pred_res}.}
  \includegraphics[width=.8\linewidth]{figures/fl_interpretable.png}
  \label{fig:FL_interpretable_vis}
\end{figure}
\end{comment}

%%%%%%%%%%%%%%%%%%%%%%%%%%%%%%%%%
\subsection{Kentucky Prediction Results for Interpretable Models}

\begin{figure}[htbp]
    \centering
    \caption{Visualizations of Broward and Kentucky interpretable results.}
    \includegraphics[width=\linewidth]{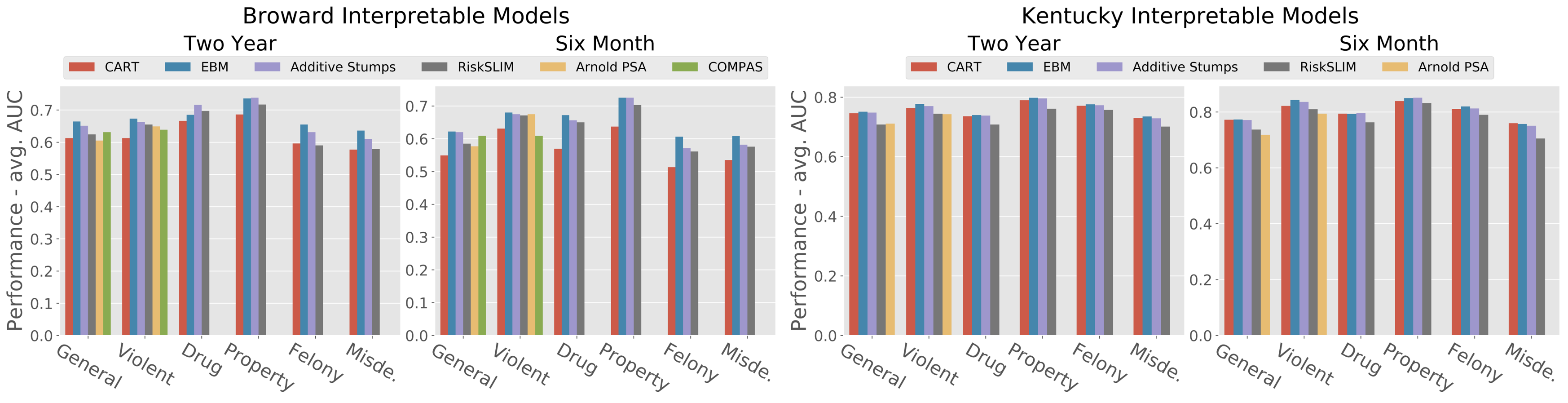}
    \label{fig:comb_interpretable}
\end{figure}

The Kentucky prediction results are visualized in Figure \ref{fig:comb_interpretable}; details are provided in Table \ref{table:kentucky_interpretable_pred_res} in Section \ref{app-tables}. For all prediction problems in both time frames, CART, EBM, and Additive Stumps all had similar performances. RiskSLIM had relatively lower results compared to other interpretable models. All interpretable models performed better than the Arnold PSA, with the exception that the Arnold PSA performed slightly better (0.3\%) than RiskSLIM on two-year \texttt{general} recidivism. Once more, we observed that the best interpretable models can perform approximately as well as the best black-box models (XGBoost). For the two-year prediction tasks, the differences in performance between the best interpretable and the best black-box models ranged from 0.7\% to 0.9\% in absolute value; for six-month problems, the difference ranged from 0.4\% to 1.5\%.

\begin{comment}
\begin{figure}[htbp]
  \centering
  \caption{Visualizations of Kentucky interpretable results from Table  \ref{table:kentucky_interpretable_pred_res}}
  \includegraphics[width=.8\linewidth]{figures/ky_interpretable.png}
  \label{fig:KY_interpretable_vis}
\end{figure}
\end{comment}

\textbf{Summary of Interpretable Models' Results:} We found that the best interpretable models performed approximately as well as the best black-box models, on both data sets and both time periods we considered, which is consistent with previous studies on other data sets \citep{ZengUsRu2017}. The best interpretable models allow judges and defendants a better understanding of the predictions that the model outputs.
\subsection{Tables and Visualizations of Interpretable Models} \label{Tables}

Each of the interpretable machine learning methods produces models that can be visualized, either as a decision tree (CART), scoring table (RiskSLIM), or as a set of visualizations (EBM, Additive Stumps). In this section, we present these tables and visualizations for EBM, Additive Stumps and RiskSLIM, to give a clearer understanding of each model's  interpretability. Here we used the two-year \texttt{general} recidivism prediction problem on Kentucky data as an example.

\subsubsection{EBM Models} \label{EBM}
\begin{figure}[H]
    \centering
    \subfigure{\includegraphics[width=.45\linewidth]{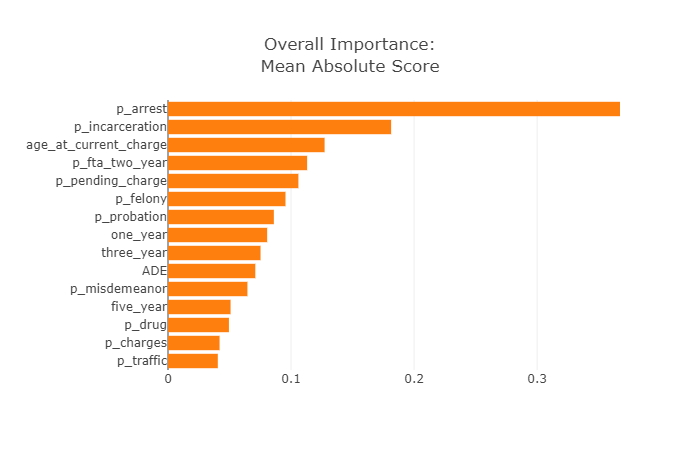}}
    \subfigure{\includegraphics[width=.45\linewidth]{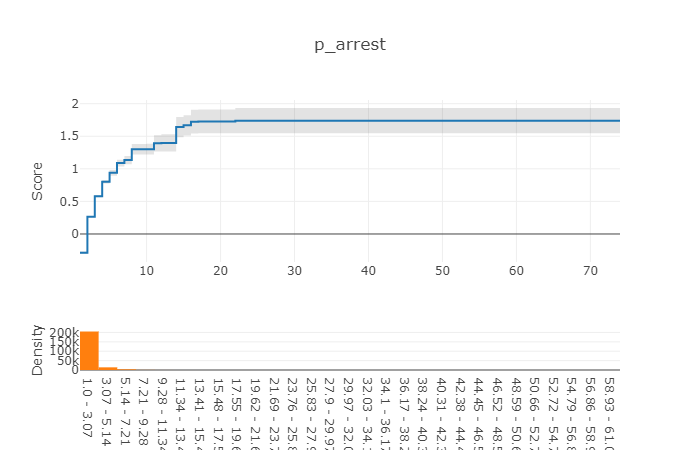}}
    \subfigure{\includegraphics[width=.45\linewidth]{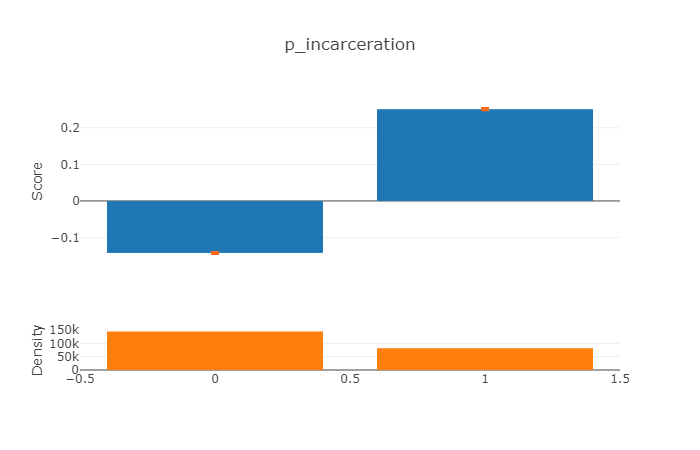}}
    \subfigure{\includegraphics[width=.45\linewidth]{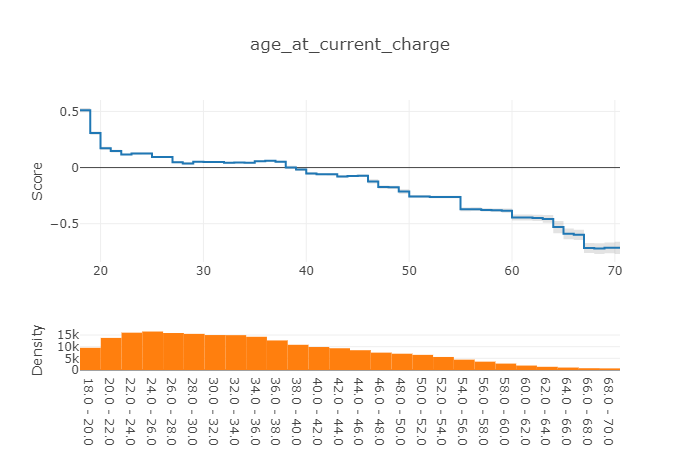}}
    \caption{Visualizations from EBM on two-year \texttt{general} recidivism. Top left: overall importance of each feature, ranked from the most important variable to least important. Remaining three: visualization for the contribution of the feature to the overall score (top) and histograms of feature values to show the distribution (bottom). Features contributions are visualized as bar charts if the feature takes binary value. The shaded grey area represents the confidence region. We see that as values get larger, there is more uncertainty in the predictions, which may be because we have fewer data points for such large feature values.}
\label{fig:ebm_models}
\end{figure}

The EBM package provides visualizations for each feature in the data set along with a bar chart of feature importance, both of which are displayed in an interactive dashboard. The dashboard allows users such as judges to see the scores corresponding to each bar or line by hovering the mouse over it. EBM models are not sparse in the number of features, so there could be visualizations for all features. Here, we show screenshots of the bar chart and visualizations for the three most important features. EBM visualizations are similar to those from Additive Stumps, in that each feature's contribution to the score can be displayed separately. However, EBM scores do not tend to be monotonically increasing or decreasing in each feature. The visualization is provided as Figure \ref{fig:ebm_models}.

\subsubsection{Additive Stumps} \label{Additive}

Additive Stumps models are constructed by thresholding the original features, such as age or criminal history, into binary stumps, followed by running $\ell_1$-penalized logistic regression on the stumps. Choosing an appropriate regularization value for $\ell_1$-penalized logistic regression can give us a model that is sparse in the number of original features---despite the fact that the regularization is directly on the stumps, not on the original features. For the Kentucky two-year \texttt{general} recidivism problem, the final model contains 28 stumps plus an intercept. These stumps are rooted under only 14 original features. Visualizations of the contributions for these 14 features are presented in Figure \ref{figure:stumpsfeatures}. Table \ref{table:ky_stumps_general} in the Appendix, containing a scoring table that includes all 28 stumps plus an intercept.

\begin{figure}[H]
    \centering
    \subfigure{\includegraphics[width=4cm]{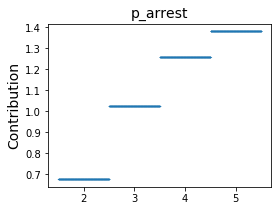}}
    \subfigure{\includegraphics[width=4cm]{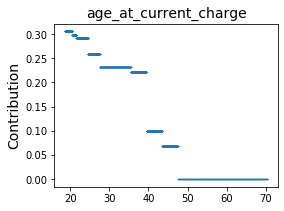}}
    \subfigure{\includegraphics[width=4.1cm]{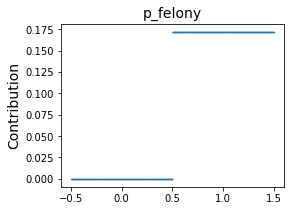}}
    \subfigure{\includegraphics[width=4.1cm]{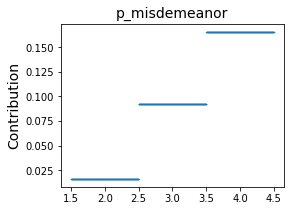}}
    \subfigure{\includegraphics[width=4.2cm]{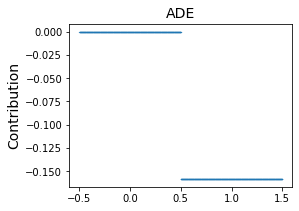}}
    %\vskip0.01ex
    \subfigure{\includegraphics[width=4.1cm]{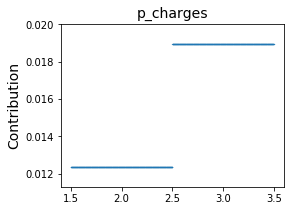}}
    \subfigure{\includegraphics[width=4cm]{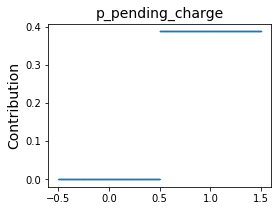}}
    \subfigure{\includegraphics[width=4cm]{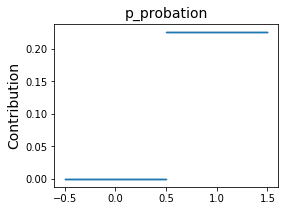}}
    \subfigure{\includegraphics[width=4.15cm]{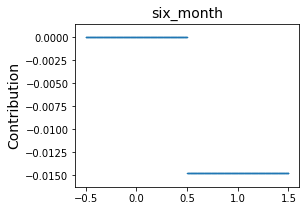}}
    \subfigure{\includegraphics[width=4.15cm]{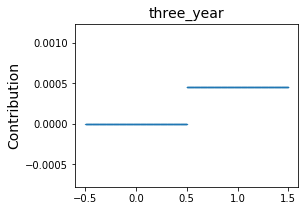}}
    %\vskip0.01ex
    \subfigure{\includegraphics[width=4.1cm]{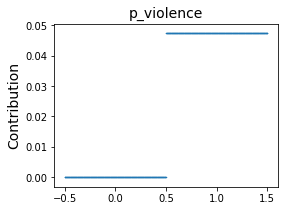}}
    \subfigure{\includegraphics[width=4.1cm]{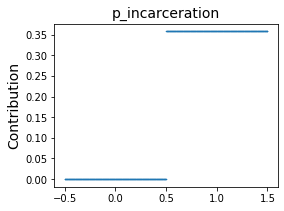}}
    \subfigure{\includegraphics[width=4.1cm]{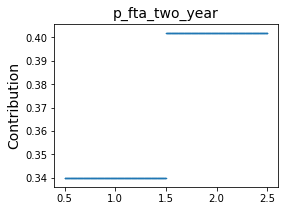}}
    \subfigure{\includegraphics[width=4.1cm]{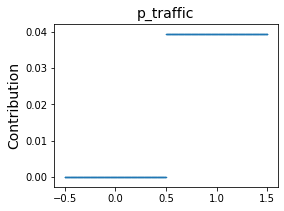}}
    \vspace*{3em}
    \caption{Visualizations of the total contribution for each of the original features in the Additive Stumps model on two-year \texttt{general} recidivism. The contribution from each stump feature is the estimated coefficient from $\ell_1$-penalized logistic regression.}
\label{figure:stumpsfeatures}
\end{figure}

\subsubsection{RiskSLIM} \label{RiskSLIM}

RiskSLIM produces scoring tables with coefficients optimized to be integers (``points''), which makes the predictions easier to calculate and interpret for users, such as judges. The total points are translated into probabilities using the logistic function provided at the top of the table. By examining a RiskSLIM model, users can easily identify which features contribute to the final score and by how much. We provide scoring tables in Table \ref{table:riskslim_table_demonstration} for two-year \texttt{general} recidivism prediction on both Broward and Kentucky data sets. More tables are provided in the Appendix.

We noticed that for each prediction problem, almost all five of the cross validation folds for the RiskSLIM algorithm yielded the same model on the larger Kentucky data set. In more detail, for Kentucky two-year \texttt{drug} and \texttt{violent} recidivism prediction problems, all five RiskSLIM models produced during cross validation were identical. For the rest of the prediction labels, four out of five cross validation models were the same. For the six-month recidivism prediction problems, the \texttt{misdemeanor} prediction problem resulted in five identical RiskSLIM models, and the \texttt{violent} recidivism prediction problem had four models that were the same. Kentucky RiskSLIM models are often the same, despite being trained on different --- albeit overlapping--- subsets of data, suggesting that they are robust to the exact subsample used for training. 
\begin{table}[ht!]
\begin{center}
\small
\tabcolsep=0.15cm
\begin{tabular}{|l|r|r|} \hline
\multicolumn{3}{c}{\textbf{Broward}} \\ \hline
\multicolumn{3}{l}{Pr(Y = +1) = 1 / (1 + exp(-(-2 + score)))} \\ \hline
age at current charge $\leq$31 & 1 points & +... \\ \hline
number of prior misdemeanor charges $\geq$4 & 1 points & ... \\ \hline
had charge(s) within last three years = Yes & 1 points & ... \\ \hline
\textbf{ADD POINTS FROM ROWS 1 TO 3}  &  	\textbf{SCORE} & = ....\\  \hline
\end{tabular}
\hspace{2em}
%\vspace*{4em}
\begin{tabular}{|l|r|r|} \hline
\multicolumn{3}{c}{\textbf{Kentucky}} \\ \hline
\multicolumn{3}{l}{Pr(Y = +1) = 1 / (1 + exp(-(-2 + score)))} \\ \hline
number of prior arrest$\geq$ 2 & 1 points & +... \\ \hline
number of prior arrest$\geq$ 3 & 1 points & +... \\ \hline
number of prior arrest$\geq$ 5 & 1 points & +... \\ \hline
\textbf{ADD POINTS FROM ROWS 1 TO 3}  &  	\textbf{SCORE} & = .....\\  \hline
\end{tabular}
\caption{Two-year \texttt{general} recidivism RiskSLIM models for Broward (left) and Kentucky (right). Each feature is given an integer point. The final predicted probability is calculated by inputting the total score to the logistic function provided on the top of the tables.}
\label{table:riskslim_table_demonstration}
\end{center}
\end{table}

%%%%%%%%%%%%%%%%%%
\section{Recidivism prediction models do not generalize well across regions} \label{Generalization}

It is common practice for recidivism prediction systems to be applied across states, or even countries, with only minor tuning on local populations. Implicit in this practice is the assumption that models trained on data from one collection of locations will perform well when used in another collection of locations---i.e., that models \textit{generalize} across locations. For instance, the Arnold PSA, which was developed on 1.5 million cases from approximately 300 U.S. jurisdictions, has been adopted in the states of Arizona, Kentucky, New Jersey, and many large cities including Chicago, Houston, Phoenix, etc. \citep{psaabout}. These systems have remained in place for years without any updates. 

However, based on our experimental results, we conjecture that different locations would benefit from specialized models that conform to the specific aspects of each location. For instance, let us briefly compare the state of Kentucky and Broward County in Florida. The demographics are completely different: Kentucky is not a diverse state (87.8\% white, 7.8\% black, and 4.4\% other groups in 2019 \citep{kentucky}), whereas Broward County is more racially diverse (62.3\%, white; 17.1\% Hispanic or Latino; 12.2\% black or African American; 5.07\% Asian and other groups \citep{broward}. The geographies of the locations are drastically different as well: Kentucky is an interior state located in the Upland South with a humid subtropical climate, whereas Broward County is at the eastern edge of Florida with a tropical climate. Several studies have indicated an association between climate (temperature, humidity, and precipitation) and crime \citep{climate1, climate2, climate3}. There are many other factors that differ between the locations that might affect the generalization of the recidivism prediction models, such as different local prosecution practices, laws and the way they are administered, social service programs, local cultures,  educational systems, and judges' views.

Because models tend to be used broadly across locations, in this section we aim to investigate how well predictive models generalize between the two locations for which we have data. We trained models on Kentucky and tested on Broward, and vice versa. We looked more closely at \textit{age}, and examined how the joint probability distribution of age and recidivism differs between Broward and Kentucky. We focused on age because of its important relationship to recidivism \citep{StevensonSl18, bushway2007inextricable, kleiman2007agedist}.

\textbf{Major Findings:} Our analysis shows that models do not generalize well across regions, and the joint probability distribution of age and recidivism varies across states. Therefore, we suggest that when possible, recidivism prediction models should be more location-specific, and be updated periodically.

\subsection{Training on One Region and Testing on the Other} 

In order to construct models on one region and test them on the other, we only used the shared features from both data sets. Nested cross validation was used to train both the models that were trained in one region and tested in the other, and the models that were trained and tested in the same region. More details about this procedure can be found in Section \ref{nestedcv}. 

\begin{figure}[htbp]
    \centering
    \caption{Visualizations of prediction differences on Broward and Kentucky data. Broward prediction differences are the AUCs of models trained on Kentucky and tested on Broward minus AUCs of models trained and tested on Broward. Kentucky prediction differences are the AUCs of models trained on Broward and tested on Kentucky minus AUCs of models trained and tested on Kentucky. These results are also presented in Tables \ref{table:train_kentucky_test_broward} - \ref{table:train_kentucky_test_kentucky} in the Appendix. }
    \includegraphics[width=\linewidth]{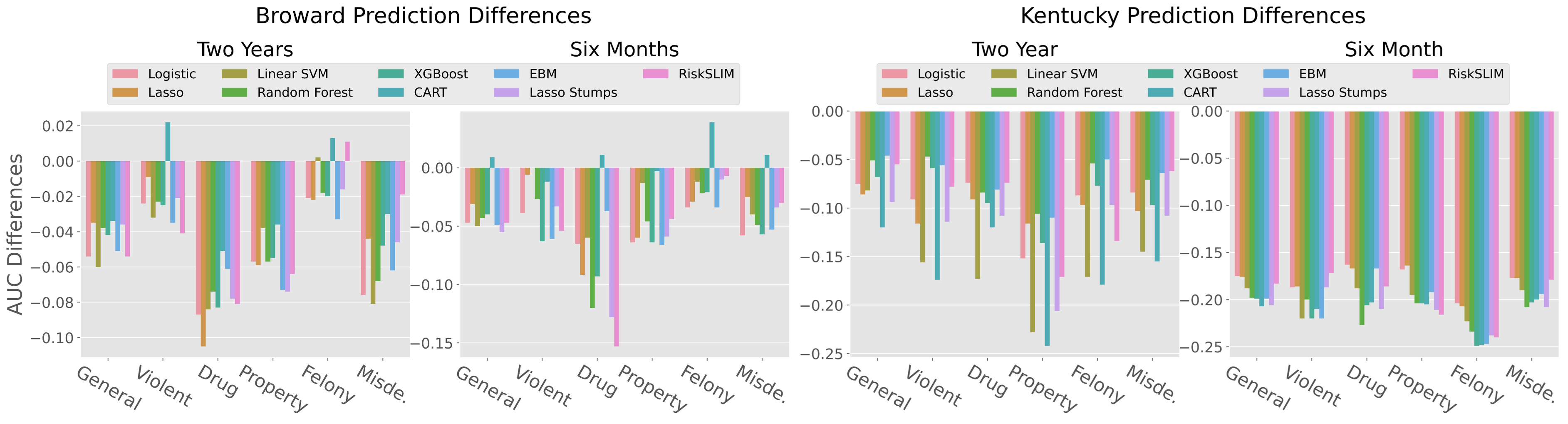}
    \label{fig:comb_geo_diff}
\end{figure}

Figure \ref{fig:comb_geo_diff} shows the difference between the performances of the models trained and tested on different regions and the performances of the models trained and tested on the same region. For Broward prediction results, we observed that there is an overall decrease in model performance when models were trained in Kentucky and tested on Broward. For instance, for the two-year \texttt{general} recidivism problem, the performances drop between 3.5\% to 6.0\% on the baseline models. A similar pattern can be observed for the interpretable models. Conversely, when we trained models on Broward and tested on Kentucky, we observed even larger performance decreases from the models trained and tested on only Kentucky. For the two-year \texttt{general} prediction task, performance gaps from baseline models range between 5.1\% and 8.6\%, while the gaps range from 4.6\% to 12.0\% for interpretable models.

Through this experimentation, we concluded that for at least the twelve prediction problems in our setup, models do not generalize across states. This could be  attributable to differences in the joint probability distribution of features and outcomes between locations. To understand the difference in these distributions more closely, we examine the age feature. 

%%%%%%%%%%%%%%%%%%%%%
\subsection{Age-Recidivism Probability Distributions by Region} \label{Age}

Age has traditionally been a highly predictive factor for recidivism \citep{ StevensonSl18, bushway2007inextricable, kleiman2007agedist}. Therefore, differences in the age distributions between two regions could significantly impact a model's ability to generalize between regions.

\begin{figure}[htbp]
  \centering
  \includegraphics[scale=0.35]{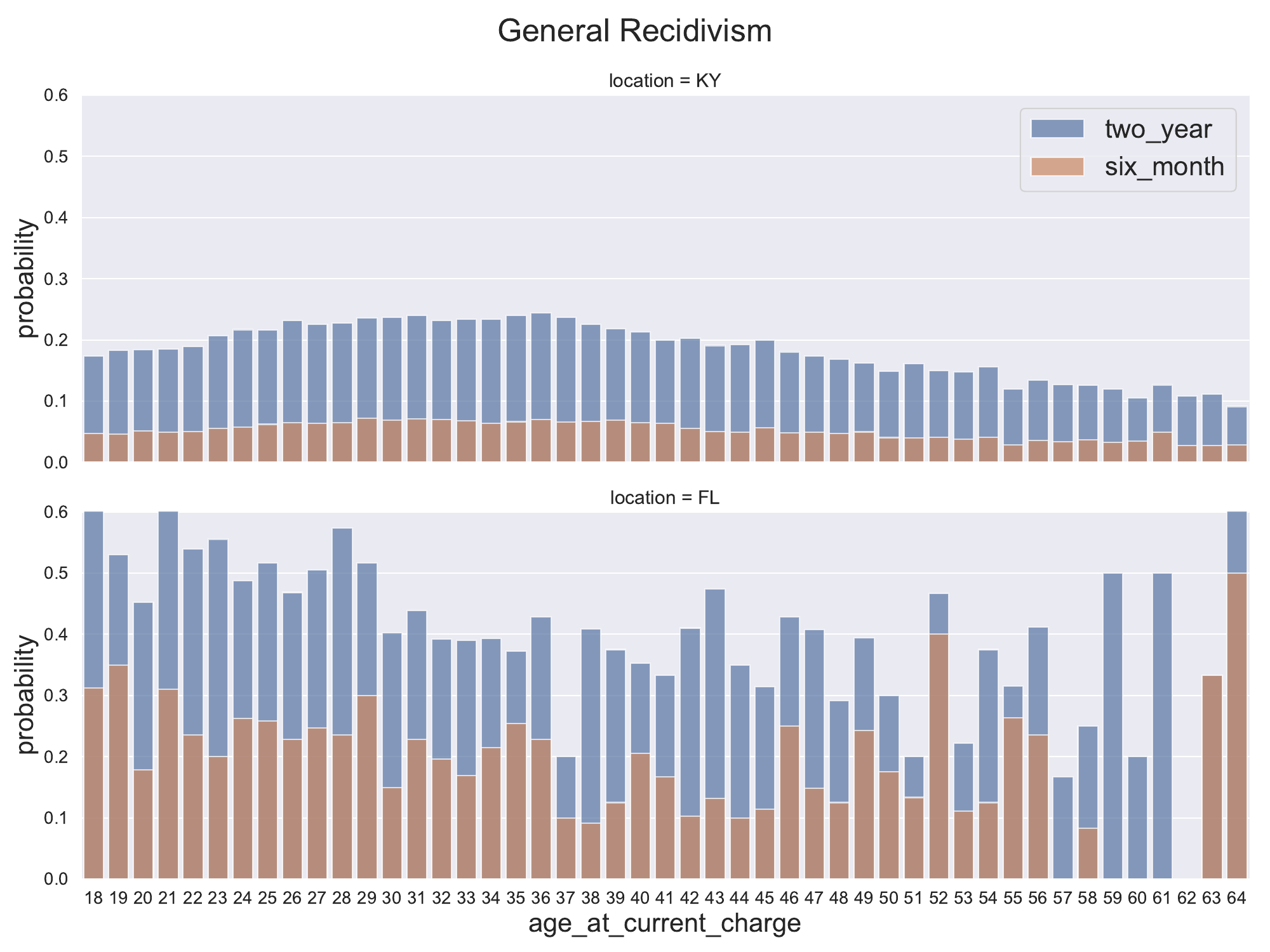}
  \caption{Probability of recidivism vs. age at current charge---\texttt{general} recidivism.}
  \label{fig:age_distribution_general}
\end{figure}

Consider the \texttt{general} recidivism problem as an example. In Kentucky, the probability of general recidivism for both six-month and two-year prediction periods peaks for individuals aged around the early to mid 30s and then decreases as age increases. In Broward County, the age distribution for the corresponding \texttt{general} recidivism problem is substantially different. From Figure \ref{fig:age_distribution_general}, the probabilities seem to peak around ages 18-29, and then decrease after age 29. There are less data for higher ages, causing greater variance in the probabilities. For the \texttt{violent} recidivism problem, please refer to Figure \ref{fig:age_distribution_violent} in the Appendix. Additionally, there is a large gap in the probability magnitudes between the two regions. For instance, the probabilities of \texttt{general} recidivism from the Broward data set can exceed 0.5, while the probabilities of general recidivism from Kentucky data are all less than 0.4. Thus, the populations of individuals from Broward and Kentucky who recidivate are different with respect to age. 

This difference is directly manifested in the interpretable models presented in Section \ref{app-tables}. We found that the selection of features differs between interpretable models trained on Broward and Kentucky data. For instance, referring to the simple RiskSLIM models listed in Section \ref{app-tables} in the Appendix, which show the most important features in each prediction problem, we noticed that with Broward data, almost all prediction problems contain at least one age feature, either ``age at current charge'' or ``age at first offense.'' This suggests that age is important in predicting recidivism across different problems trained on the Broward data. However, none of the RiskSLIM models trained on the Kentucky data set  use age features. Almost all the models use ``prior arrest'' features, reflecting the fact that Kentucky recidivism prediction problems rely more on prior criminal history information than on age.

%%%%%%%%%%%%%%%%%%%%%
\section{Fairness} \label{Fairness}

In this section, we conduct a technical discussion of a small fraction of the various fairness definitions that have emerged recently, and an evaluation of how well the interpretable models satisfy them on the Kentucky data set. We first describe our rationale for selecting fairness definitions (calibration and balanced group AUC). Next, we evaluate how well the Arnold PSA, COMPAS, EBM (the best-performing interpretable models) and RiskSLIM (the most interpretable and most constrained models) satisfy these definitions on the two-year \texttt{general} recidivism and two-year \texttt{violent} recidivism problems in Kentucky. Finally, we discuss how current fairness enforcement procedures interact with interpretability.

\textbf{Major Findings:} Empirically, we found no violations of the fairness definitions (calibration and balanced group AUC) for both interpretable machine learning models we assessed (EBM and RiskSLIM) for the \texttt{two-year} general recidivism problem on the Kentucky data set. We found that the Arnold NCA raw score violated calibration at higher values of the score. Overall, we observed a larger gap in fairness for both fairness measures we examined between the largest and smallest sensitive groups, than between black and white sensitive groups. We also note that existing techniques to enforce fairness generally require non-interpretable transformations, and therefore do not work well with interpretable models.

%%%%%%%%%%%%%%%%%%%%%%
\subsection{Selection of Fairness Metrics: Calibration, Balance for Positive/Negative Class, Balanced Group AUC} \label{Fairness_metrics}

As discussed in Section \ref{Setup}, we do not wish to consider binary risk scores in this study. This decision limits us to a much smaller class of fairness definitions, e.g., statistical parity would not be relevant. Below, we summarize the definitions that apply to regression that we \textbf{do not} consider and the reasons why: 
\begin{itemize}
    \item \textbf{Fairness through unawareness} states that a model should not use any sensitive features \citep{verma2018fairnessreview}. However, if there are proxies for sensitive features present in the data set, the model can still learn an association between a sensitive group and the outcome. Fairness through unawareness could be used if one decides that a proxy feature is permissible to use---for instance, if one decided that age could be used, despite its correlation with race---but we do not presume that this is what is desired for this application. Of course, if fairness through unawareness \textit{is} desired, it is easy to construct models that satisfy this definition.
    
    \item \textbf{Individual fairness} intuitively requires that ``similar'' individuals are treated ``similarly'' by the model---individuals with similar features should be given similar model scores. This type of fairness requires manually and thus subjectively defining a notion of similarity between individuals \citep{dwork2012fairaware}. This type of subjective choice goes beyond the scope of this paper. 
    
    \item \textbf{Balance for Positive/Negative Class} (BPC/BNC) states that it is permissible to give consistently higher (respectively lower) scores to individuals who truly belong to the positive (respectively negative) class. However, BPC/BNC limits the set of attributes where it is permissible to  ``discriminate'' between individuals, to the label $Y$. Suppose the count of prior offenses is an important feature for a recidivism prediction model---higher prior counts lead to higher scores. This is a reasonable model assumption because a higher prior count is correlated with higher recidivism rates. If on average, African-Americans have higher prior counts than Caucasians, the model will not satisfy BPC/BNC. For a model to satisfy BPC/BNC, it must give the same average score to individuals from a certain race and with a certain recidivism label, regardless of distributional differences in prior counts. Those who believe that prior counts and arrests are not racially biased against African-Americans might find this a desirable property of a fairness definition. On the other hand, those who find this undesirable can fix this by conditioning on the prior counts attribute as well. Thus, this fairness definition requires deciding which features are group-biased, a subjective conditioning that also goes beyond the scope of this work.
\end{itemize} 

% say the definitions we consider 
Once we limited ourselves to real-valued outcomes and eliminated the above definitions, only a few definitions remained. In a literature search for nonbinary fairness definitions, we found the fairness definitions of calibration and balanced group AUC (BG-AUC). 

Below, $G$ denotes a (categorical) \textbf{sensitive attribute} such as race, and $g_i$ denotes one of the \textit{sensitive groups} in $G$ (e.g. African-American, Caucasian, and Hispanic, for the sensitive attribute of race). $Y \in \{0, 1\}$ denotes the ground-truth label (recidivism status) and $S$ denotes the predicted score from a model.  

% calibration def

\begin{itemize}
    \item \textbf{Calibration}: We consider two notions of calibration. The first, \textbf{group calibration}, requires that for all predicted scores, the fraction of positive labels is approximately the same across all groups. Mathematically, group calibration over the sensitive attribute $G$ requires:
    \small{$$
    P(Y=1 | S=s, G= g_i) \approx P(Y=1 | S=s, G=g_j), \forall i, j
    $$}
    where $s$ is the given value of a risk score. Note that in the case where scores are binary, group calibration is equivalent to requiring \textit{conditional use accuracy equality}. In practice, it is common to bin the score $S$ if there are many possible values. The second, \textbf{monotonic calibration}, requires that if $s_1 < s_2$, then $P(Y = 1 | S = s_1) < P(Y = 1 | S = s_2)$. \footnote{We note that a real-valued score $S$ between $0$ and $1$ is \textit{well-calibrated} if $P(Y = 1 | S = s) = s$. Well-calibration says that the predicted probability of recidivism should be the same as the true probability of recidivism \citep{verma2018fairnessreview}. Although well-calibration is the definition of calibration that is standard in the statistics community, we consider monotonic-calibration here because any score that is monotonically-calibrated can be transformed to be well-calibrated.}
    
    These types of calibration are of particular concern to designers of current recidivism risk models. Group calibration means that a risk score holds the same ``meaning'' for each race. Monotonic calibration means that if the score increases, the risk also increases. These notions are important because human decision-makers expect risk scores to have these intuitive properties (but not all algorithms produce calibrated models) \citep{chouldechova2016dispimpact}. 

    % BG-AUC Definition
    \item \textbf{Balanced Group AUC (BG-AUC)} requires that the AUC of the risk score is approximately the same for each sensitive group. This definition is our adaptation of \textbf{overall accuracy equality} \citep{berk2017}, which asks that the score's accuracy is the same for each sensitive group. Our risk scores are not binary so we do not assess accuracy in this work, but assessing the AUC for each group is the natural analog.
\end{itemize}
 
% what we looked at 
\textbf{Sensitive attributes}: The two sensitive attributes that are available in the Kentucky data sets are race and gender. In the Kentucky data set, all individuals are partitioned into \texttt{Caucasian, African-American, Indian, Asian, and Other}, but we group the \texttt{Indian} and \texttt{Asian} attributes into \texttt{other} because there are very few individuals with these attributes. See Table \ref{table:FL_KY_sensitive_attrs}  for the distribution of sensitive attributes in Kentucky. The Kentucky data set also partitions individuals into the genders \texttt{female} and \texttt{male}.
To summarize: races in Kentucky = \{\texttt{Caucasian, African-American, other}\}; sexes in Kentucky = \{\texttt{female, male}\}.

%%%%%%%%%%%%%%%%%%%%%%%%%%%
\subsection{Fairness Results} \label{Fairness_results}
We assessed model fairness only on the Kentucky data because the Broward data has a limited sample size, potentially making the fairness results unreliable. We attempted the evaluation on Broward data, but conditioning on race/gender and the true label/score in the Broward data led to subgroups that were too small, and therefore noisy results. We compared the  interpretable models, EBM and RiskSLIM, to the Arnold PSA on Kentucky. EBM has the best performance on most of the prediction problems on the Kentucky data set. RiskSLIM performs relatively worse, but is considerably simpler as there are no more than five features in each model, coefficients are integers, and the model is linear. 

We evaluated the two-year general and two-year violent problems, as they are the primary problems that the Arnold PSA is used for. For the two-year general problem, we evaluated the unscaled Arnold New Criminal Activity (NCA) score; for the two-year violent problem, we assessed the unscaled Arnold New Violent Criminal Activity (NVCA) score. Although Arnold Ventures provides a table to scale the Arnold scores, in Kentucky, judges are presented with the unscaled scores along with a categorization of the scores as  low, medium, and high risk. Results for both two-year general and violent recidivism can be found in the Appendix. 

%%%%%%%%%%%%%%%%%%%%%%%%%%%%%%%
% calibration results for arnold, ebm, and riskslim
\begin{figure}[ht]
\centering
\caption{Calibration results for the Arnold NCA raw, EBM and RiskSLIM for two-year \texttt{general} recidivism on Kentucky.}
\subfigure{
\includegraphics[width=.32\textwidth]{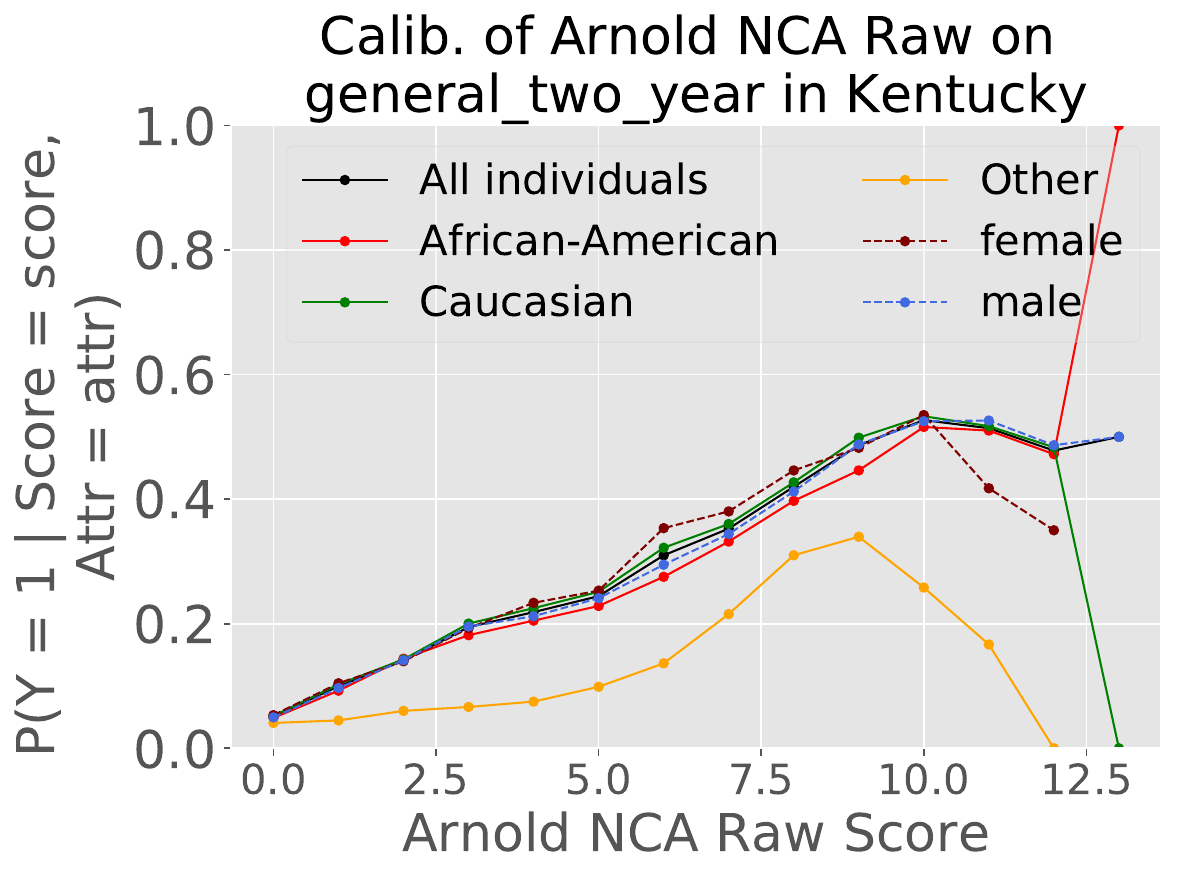}}\;\;\;\;
\subfigure{\includegraphics[width=.32\textwidth]{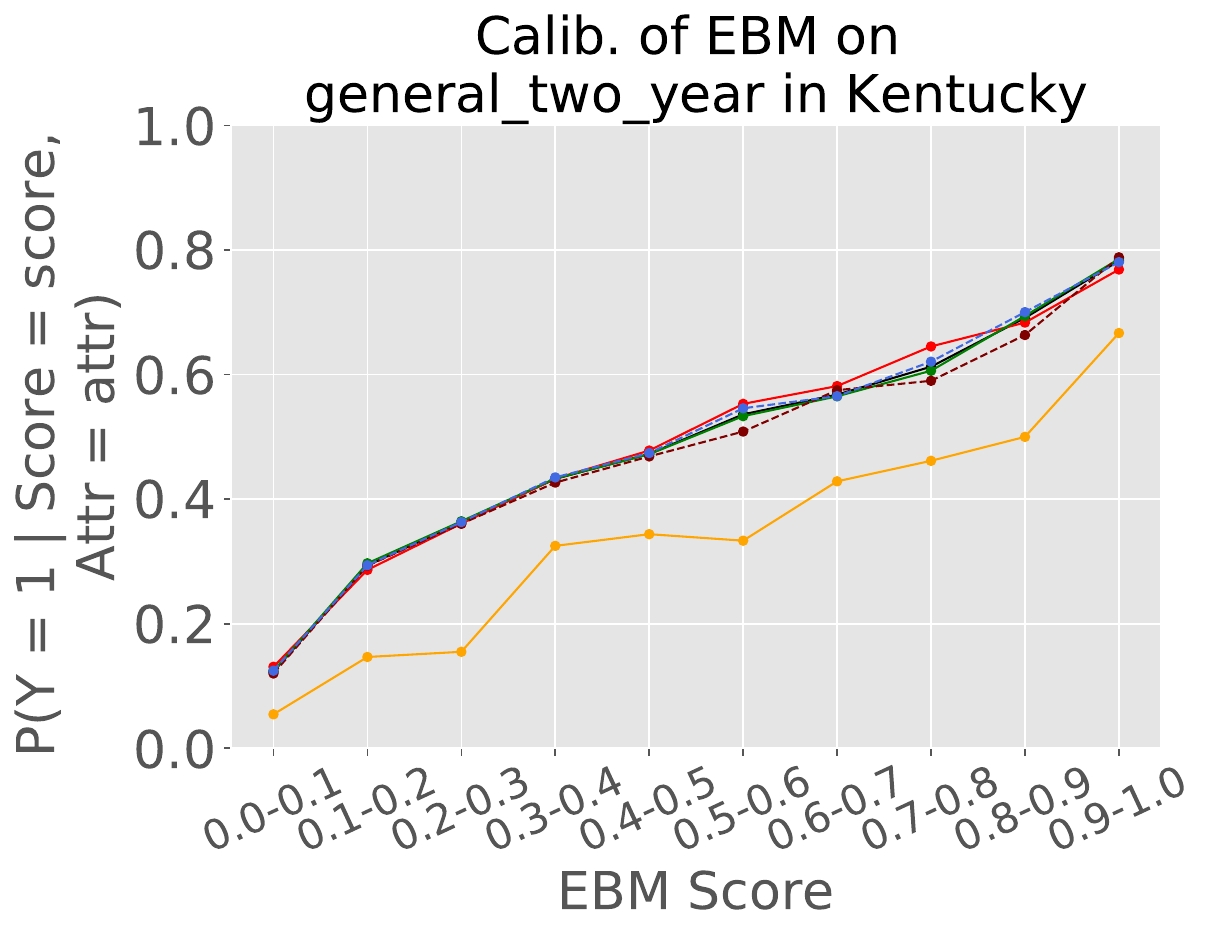}}
\subfigure{\includegraphics[width=.32\textwidth]{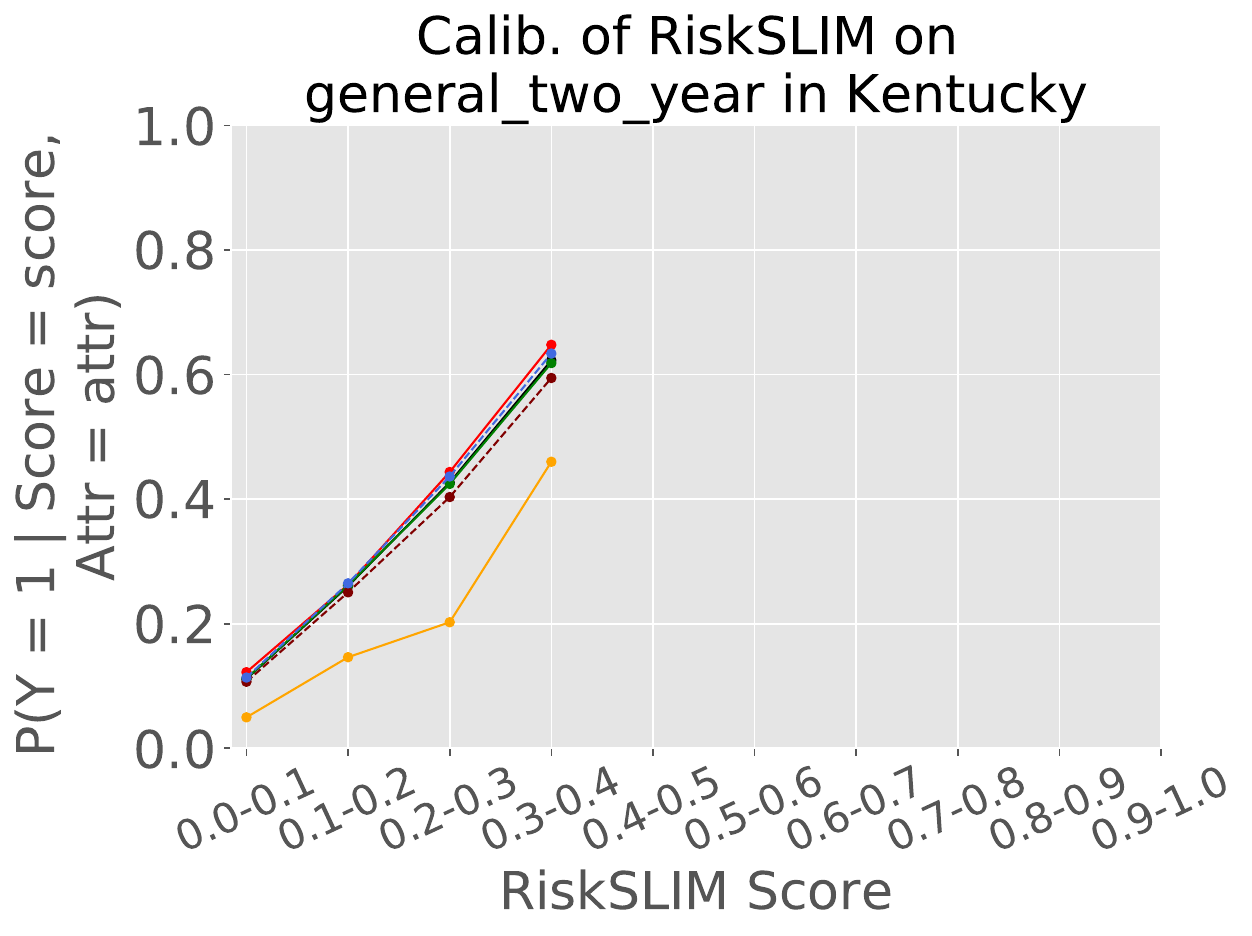}}
\label{fig:calib_arnold_nca}
\end{figure}
%%%%%%%%%%%%%
\subsubsection*{Calibration} \label{Calibration_results}
% Arnold calibration 
Figure \ref{fig:calib_arnold_nca} shows that the Arnold NCA raw score approximately satisfies monotonic and group calibration for race and gender on lower values of the risk score (i.e., scores less than 10) and except for the ``Other'' group. There are fewer individuals with high Arnold NCA scores (e.g., 12 or 13) in the dataset, leading to higher variance in predictions, which may be why higher Arnold NCA raw scores fail the calibration definitions. Interestingly, we found that the scaled version of Arnold NCA fully satisfied monotonic and group calibration, but had slightly worse predictive performance.
% EBM and RiskSLIM calibration
EBM and RiskSLIM both satisfy monotonic calibration and group calibration for all gender and race groups (excluding the ``Other'' group).

\subsubsection*{Balanced Group AUC (BG-AUC)} \label{Balance_AUC}

In Kentucky, AUC values are stable across sensitive attributes for all models. The discrepancies in AUC between African-Americans and Caucasians range from 0.3\% (RiskSLIM) to 2.1\% (Arnold NCA raw). The range gets smaller for gender groups, lying between 0.5\% (Arnold NCA) to 1.3\% (RiskSLIM). 
Hence, we found that RiskSLIM has the least AUC difference between race groups (excluding the ``Other'' group), whereas Arnold NCA has the least AUC difference between gender groups. 
We note that the differences in AUCs between models are small, with the largest differences manifesting with the ``Other'' group.

\begin{table*}[ht]
\begin{center}
\scriptsize
\tabcolsep=0.1cm
\renewcommand{\arraystretch}{1.4}
\caption{AUCs of the Arnold NCA Raw, EBM and RiskSLIM on Kentucky, conditioned on sensitive attributes. AUC ranges are given for each sensitive attribute.}
\begin{tabular}{|c|c|c|c|c|c|c|c|c|}\hline
    %% KENTUCKY
    \multicolumn{9}{c}{\textbf{Kentucky}} \\ \hline
    & & \multicolumn{3}{c}{\textbf{Race}} & & \multicolumn{3}{c}{\textbf{Sex}} \\ \hline
    \verb|    Model     | & \verb|      Label       | & \verb|Afr-Am.| & \verb|Cauc.| & \verb|Other Race| & \verb|race_range| & \verb|Female| & \verb|Male | & \verb|sex_range| \\ \hline
    Arnold NCA Raw & general\_two\_year &   0.692 & 0.713 &      0.653 &      0.059 &  0.714 & 0.709 &     0.005 \\ \hline
    EBM            & general\_two\_year &   0.742 & 0.751 &      0.696 &      0.055 &  0.745 & 0.753 &     0.008 \\ \hline
    RiskSLIM       & general\_two\_year &   0.705 & 0.708 &      0.620 &      0.088 &  0.699 & 0.712 &     0.013 \\ \hline
\end{tabular}
\end{center}
\vspace{-5mm}
\end{table*}

\textbf{Summary of Fairness Results:} For the two-year general recidivism problem on the Kentucky data set, we found no egregious violations of calibration and BG-AUC for the models we assessed (Arnold NCA, EBM and RiskSLIM). However, we did find small violations. For example, we found the Arnold NCA raw score violated calibration for higher scores.

A caveat is that we limited the discussion of the race groups to Caucasians and African-Americans, else the ``Other'' group would have caused all models to fail all definitions of fairness: calibration curves for the ``Other'' group are significantly beneath curves for other groups and prediction AUC is significantly lower for the ``Other'' group. This may be because we have the least data for the ``Other'' race group, which is only 2.49\% of the total sample. To ensure fairness, it is important that comparable amounts of data are gathered for each sensitive group when possible. However, in non-diverse states such as Kentucky, there may not be enough individuals in minority groups to create a large enough statistical sample. 

%%%%%%%%%%%%%%%%%%%%%
\subsection{A Discussion on the Interaction between Fairness and Interpretability} \label{F_I}

There are significant hurdles to using current fairness techniques with interpretable models. Moreover, the vast majority of the work on fairness has focused on the binary classification case. Thus, few definitions of fairness (let alone algorithms) work for problems where predictions are nonbinary.
        
We did not attempt to use fairness enforcement techniques because many fairness techniques require a non-interpretable transformation. Once these transformations are made, there is no way to correct them to produce an interpretable model afterwards. There are generally three approaches to fairness algorithms: preprocessing of features \citep{zemel2013fairrepr}, altering the training loss function \citep{berk2017convexfairness, agarwal2019fairregression}, and post-processing of predictions \citep{hardt2016eqodds, agarwal2018fairbinaryclass, pleiss2017calib}. The pre-processing steps are generally complicated transformations of the input features, which shreds the data's natural meaning. Similarly, post-processing approaches either transform the predictions in some way, performing ``fairness corrections'' \citep{pleiss2017calib} (which are non-interpretable), or require threshold selection, which is contrary to our goals of providing nonbinary risk assessments \citep{hardt2016eqodds}. The approaches to modify training loss functions are the most promising, but model optimization for both fairness and interpretability constraints would require new algorithms and is beyond the scope of this work.

In problems where fairness is a significant concern, machine learning outputs are likely to be used as decision tools rather than decision-makers, so it is surprising that so little work has thoroughly examined fairness for regression or probability estimation.

\section{Discussion and Future Work} \label{Discussion}

From this analysis, we conclude that the interpretable models can indeed perform approximately as well as the black-box models in various recidivism prediction problems. On the Broward data set, we found that RiskSLIM, EBM, and Additive Stumps perform as well or better than the best black-box models. On the Kentucky data set, we observed that EBM and Additive Stumps have extremely close performance to the best black-box models---Random Forest and XGBoost--- with average AUC differences around 1\%, which is less than the uncertainty gap. For the purposes of criminal recidivism prediction, our work indicates that there is no practical loss in accuracy by using interpretable models and much to be gained in interpretability. 

We observed that machine learning models for six-month outcomes generally outperform those for two-year outcomes, conditioning on the recidivism type. This may be because treatment/rehabilitation programs have a greater chance of taking effect over a two-year time span, as compared to the six-month time span, altering the probability of recidivism. Future work could investigate this hypothesis, or pose other hypotheses to explain this observation. 

We observed significant differences in the age distributions for Kentucky and Broward County, and hypothesized that this difference may be why machine learning models do not generalize well between regions. One might easily imagine regional feature distributions shifting over \textit{time} as well, which is supported by several studies \citep{totalarrestrate, violentrate, homociderate, agecrime}. Even though these studies focused on disparate crime types, they consistently observed a drop in the rate of offending among younger people since the 1990s. Studies have explicitly shown that the distributions of age versus arrest rate has changed over time as well. For instance, \citet{totalarrestrate} has reported that in the state of New York, the mean age of the total arrested population increased by two years between 1990 and 2010. They hypothesized that a decrease in arrests in younger people and an increase in arrests in older people together contributed to the increase in mean age. There are many reasons why data would change over time and over jurisdictions. Changing policies (e.g., the NYC stop and frisk program) could potentially alter who would be arrested and for what types of crime. New cultural phenomena, for instance originating from media, could also influence people's behavior at a large scale. 

The above observations lead us to conclude that different recidivism prediction models could be constructed for different locations and should be periodically updated. Machine learning models are well-suited for efficient creation and updating of these kinds of models. A possible future line of work is to separate the Kentucky data at the jurisdiction level, and perform a causal analysis of the effects of different judicial and policing practices on the recidivism distribution. While local jurisdictions, e.g., at the county or district level,  might not have sufficient resources to fit their own recidivism prediction models, our analysis on the Kentucky dataset shows that even considering \textit{state-level} recidivism models produces gains over models trained across nationwide data, such as the Arnold PSA. It may be more feasible for the state-level agencies to collect enough data and hire analysts to fit the models. Future work could also investigate using small quantities of location-specific data to fine-tune more general models.

Simple, linear models have been used for criminal justice applications for almost a century \citep{burgess1928factors, hart1924parole}. They have the advantage that one can easily quantify the contributions of each feature to the predicted score. Judicial actors without much statistics background can understand these scores, and use them to help solve societal issues. Interpretable models are extremely valuable for current decision-making processes in criminal justice: they allow error-checking, help ensure due process, and allow judges to incorporate information outside the database into their decision-making process in a calibrated manner. 

However, our work on interpretable risk prediction is only one step closer to what we view as the ultimate goal---placing recidivism prediction into the framework of formal decision analysis. Decision-making in the context of decision analysis involves the minimization of costs rather than risks. Towards this end, \citet{lakkarajurudin2017} considered several costs related to pretrial release decisions; these include the societal cost of releasing an individual who might commit a crime before their trial, the cost of assigning an officer to an individual, and the cost to taxpayers of keeping an individual incarcerated. The importance of risk predictions vary between decision-making problems (release, parole, sentencing, etc.). In some cases, they play a minor role, yet in others, predictions may comprise the sole deciding factor. Because of this, it would be useful to have a cost-benefit analysis \textit{per decision} that would help determine exactly when and where risk scores should participate. 

Hence, an important and necessary direction for the future work would be to incorporate the framework of classical decision analysis into decision-making in the criminal justice system. Decision analysis tools would ideally allow practitioners to strike a balance between relevant considerations: for instance, future risk to society, costs of treatment programs, costs to families involved in the criminal justice system, costs to the individual, as well as more traditional modelling objectives such as fairness, interpretability, transparency, and predictive performance. While the full data measuring costs and risks to all stakeholders in the criminal justice process may never be available, it is important to move in this direction, as this would bring us closer to more consistent and informed decision making. 

%% bibliography
\bibliography{recidivism}

\begin{thebibliography}{}

\bibitem[Agarwal et~al., 2018]{agarwal2018fairbinaryclass}
Agarwal, A., Beygelzimer, A., Dud{\'\i}k, M., Langford, J., and Wallach, H.
  (2018).
\newblock A reductions approach to fair classification.
\newblock In {\em Proceedings of the 35th International Conference on Machine
  Learning}.
\newblock \url{https://proceedings.mlr.press/v80/agarwal18a.html}.

\bibitem[Agarwal et~al., 2019]{agarwal2019fairregression}
Agarwal, A., Dud{\'\i}k, M., and Wu, Z.~S. (2019).
\newblock Fair regression: Quantitative definitions and reduction-based
  algorithms.
\newblock In {\em Proceedings of the 36th International Conference on Machine
  Learning}.
\newblock \url{https://proceedings.mlr.press/v97/agarwal19d.html}.

\bibitem[Alfred, 2006]{homociderate}
Alfred, B. (2006).
\newblock The crime drop in america: An explanation of some recent crime
  trends.
\newblock {\em Journal of Scandinavian Studies in Criminology and Crime
  Prevention}, 7:17--35.

\bibitem[{American Law Institute}, 2017]{modelpenalcode2017}
{American Law Institute} (2017).
\newblock Model penal code.
\newblock \url{https://www.ali.org/projects/show/sentencing/}.

\bibitem[Angelino et~al., 2018]{angelino2018}
Angelino, E., Larus-Stone, N., Alabi, D., Seltzer, M., and Rudin, C. (2018).
\newblock Certifiably optimal rule lists for categorical data.
\newblock {\em Journal of Machine Learning Research}, 19:1--79.

\bibitem[Angwin et~al., 2016]{AngwinLaMaKi16}
Angwin, J., Larson, J., Mattu, S., and Kirchner, L. (2016).
\newblock Machine bias.
\newblock Technical report, ProPublica.

\bibitem[Barabas et~al., 2019]{nytoped}
Barabas, C., Dinakar, K., and Doyle, C. (2019).
\newblock The problems with risk assessment tools.
\newblock {\em The New York Times}.
\newblock \url{https://www.nytimes.com/2019/07/17/opinion/pretrial-ai.html}.

\bibitem[Barocas and Selbst, 2016]{barocasdispimpact2016}
Barocas, S. and Selbst, A.~D. (2016).
\newblock Big data's disparate impact.
\newblock {\em California Law Review}, 104:671--732.

\bibitem[Berk, 2017]{berk2017impact}
Berk, R. (2017).
\newblock An impact assessment of machine learning risk forecasts on parole
  board decisions and recidivism.
\newblock {\em Experimental Criminology}, 13:193--216.

\bibitem[Berk et~al., 2017a]{berk2017convexfairness}
Berk, R., Heidari, H., Jabbari, S., Joseph, M., Kearns, M., Morgenstern, J.,
  Neel, S., and Roth, A. (2017a).
\newblock A convex framework for fair regression.
\newblock {\em arXiv preprint arXiv:1706.02409}.

\bibitem[Berk et~al., 2017b]{berk2017}
Berk, R., Heidari, H., Jabbari, S., Kearns, M., and Roth, A. (2017b).
\newblock Fairness in criminal justice risk assessments: The state of the art.
\newblock {\em Sociological Methods \& Research}.

\bibitem[Berk et~al., 2005]{berk2005developing}
Berk, R.~A., He, Y., and Sorenson, S.~B. (2005).
\newblock Developing a practical forecasting screener for domestic violence
  incidents.
\newblock {\em Evaluation Review}, 29(4):358--383.

\bibitem[Bindler and Hjalmarsson, 2018]{bindler2018agedistlondon}
Bindler, A. and Hjalmarsson, R. (2018).
\newblock How punishment severity affects jury verdicts: Evidence from two
  natural experiments.
\newblock {\em American Economic Journal: Economic Policy}, 10.

\bibitem[Binns, 2018]{binnsfair}
Binns, R. (2018).
\newblock Fairness in machine learning: Lessons from political philosophy.
\newblock {\em Journal of Machine Learning Research}, 81:1--11.

\bibitem[Breiman et~al., 1984]{breiman1984classification}
Breiman, L., Friedman, J., Stone, C.~J., and Olshen, R.~A. (1984).
\newblock {\em Classification and regression trees}.
\newblock CRC press.

\bibitem[Brennan et~al., 2009]{northpointe}
Brennan, T., Dieterich, W., and Ehret, B. (2009).
\newblock Evaluating the predictive validity of the {COMPAS} risk and needs
  assessment system.
\newblock {\em Criminal Justice and Behavior}, 36(1):21--40.

\bibitem[{Bureau of Justice Assistance}, 2020]{history}
{Bureau of Justice Assistance} (2020).
\newblock History of risk assessment.
\newblock {\em Bureau of Justice Assistance}.
\newblock \url{https://psrac.bja.ojp.gov/basics/history}.

\bibitem[Burgess, 1928]{burgess1928factors}
Burgess, E.~W. (1928).
\newblock Factors determining success or failure on parole.

\bibitem[Bushway and Piehl, 2007]{bushway2007inextricable}
Bushway, S.~D. and Piehl, A.~M. (2007).
\newblock The inextricable link between age and criminal history in sentencing.
\newblock {\em Crime \& Delinquency}, 53(1):156--183.

\bibitem[Cadigan and Lowenkamp, 2011]{ptra}
Cadigan, T.~P. and Lowenkamp, C.~T. (2011).
\newblock Implementing risk assessment in the federal pretrial services system.
\newblock {\em Federal Probation}, 75(2).

\bibitem[Carollo et~al., 2007]{conn}
Carollo, J., Hines, M., and Hedlund, J. (2007).
\newblock Expanded validation of a decision aid for pretrial conditional
  release.
\newblock Technical report, Central Connecticut State University.

\bibitem[Chen and Guestrin, 2016]{xgboost}
Chen, T. and Guestrin, C. (2016).
\newblock Xgboost: A scalable tree boosting system.
\newblock In {\em Proceedings of the 22nd acm sigkdd international conference
  on knowledge discovery and data mining}, pages 785--794.

\bibitem[Chouldechova, 2017]{chouldechova2016dispimpact}
Chouldechova, A. (2017).
\newblock Fair prediction with disparate impact: A study of bias in recidivism
  prediction instruments.
\newblock {\em Big data}, 5(2):153--163.

\bibitem[Cook and Laub, 2002]{violentrate}
Cook, P. and Laub, J. (2002).
\newblock After the epidemic recent trends in youth violence in the united
  states.
\newblock {\em Crime and Justice}, 29:1--37.

\bibitem[Corbett-Davies and Goel, 2018]{corbett-daviesmeasure}
Corbett-Davies, S. and Goel, S. (2018).
\newblock The measure and mismeasure of fairness: A critical review of fair
  machine learning.
\newblock {\em arXiv:1808.00023v2}.

\bibitem[Corbett-Davies et~al., 2017]{corbettdavies2017faircost}
Corbett-Davies, S., Pierson, E., Feller, A., Goel, S., and Huq, A. (2017).
\newblock Algorithmic decision making and the cost of fairness.
\newblock In {\em In Proceedings of the 23rd ACM SIGKDD International
  Conference on Knowledge Discovery and Data Mining}, pages 797--806.

\bibitem[Dawes et~al., 1989]{dawes1989clinical}
Dawes, R.~M., Faust, D., and Meehl, P.~E. (1989).
\newblock Clinical versus actuarial judgment.
\newblock {\em Science}, 243(4899):1668--1674.

\bibitem[Defronzo, 1984]{climate3}
Defronzo, J. (1984).
\newblock Climate and crime: Tests of an fbi assumption.
\newblock {\em Environment and Behavior}, 16.

\bibitem[Desmarais et~al., 2019]{desmarais2019rebuttal}
Desmarais, S., Garrett, B., and Rudin, C. (2019).
\newblock Risk assessment tools are not a failed 'minority report'.
\newblock {\em Law360}.
\newblock
  \url{https://www.law360.com/access-to-justice/articles/1180373/risk-assessment-tools-are-not-a-failed-minority-report-}.

\bibitem[Dieterich et~al., 2016]{northpointeresponse}
Dieterich, W., Mendoza, C., and Brennan, T. (2016).
\newblock {COMPAS} risk scales: Demonstrating accuracy equity and predictive
  parity: Performance of the {COMPAS} risk scales in broward county.
\newblock Technical report, Northpointe, Inc.

\bibitem[Dwork et~al., 2012]{dwork2012fairaware}
Dwork, C., Hardt, M., Pitassi, T., Reingold, O., and Zemel, R. (2012).
\newblock Fairness through awareness.
\newblock In {\em Proceedings of the 3rd Innovations in Theoretical Computer
  Science Conference}, ITCS '12, pages 214--226, New York, NY, USA. ACM.

\bibitem[{Electronic Privacy Information Center}, 2016]{epic}
{Electronic Privacy Information Center} (2016).
\newblock Algorithms in the criminal justice system.
\newblock {\em Electronic Privacy Information Center}.
\newblock \url{https://epic.org/algorithmic-transparency/crim-justice/}.

\bibitem[Fan et~al., 2008]{liblinear}
Fan, R.-E., Chang, K.-W., Hsieh, C.-J., Wang, X.-R., and Lin, C.-J. (2008).
\newblock Liblinear: A library for large linear classification.
\newblock {\em J. Mach. Learn. Res.}, 9:1871–1874.

\bibitem[Flores et~al., 2016]{flores16}
Flores, A.~W., Lowenkamp, C.~T., and Bechtel, K. (2016).
\newblock False positives, false negatives, and false analyses: A rejoinder to
  ``{M}achine bias: There's software used across the country to predict future
  criminals".
\newblock {\em Federal probation}, 80(2).

\bibitem[Frase et~al., 2015]{criminalsourcebook}
Frase, R.~S., Roberts, J., Hester, R., and Mitchell, K.~L. (2015).
\newblock Robina institute of criminal law and criminal justice, criminal
  history enhancements sourcebook.
\newblock
  \url{https://robinainstitute.umn.edu/publications/criminal-history-enhancements-sourcebook}.

\bibitem[Freeman, 2016]{freemanloomis}
Freeman, K. (2016).
\newblock Algorithmic injustice: How the wisconsin supreme court failed to
  protect due process rights in state v. loomis.
\newblock {\em North Carolina Journal of Law \& Technology}, 18.
\newblock \url{http://ncjolt.org/wp-content/uploads/2016/12/Freeman_Final.pdf}.

\bibitem[Freund and Schapire, 1997]{freund1997decision}
Freund, Y. and Schapire, R.~E. (1997).
\newblock A decision-theoretic generalization of on-line learning and an
  application to boosting.
\newblock {\em Journal of computer and system sciences}, 55(1):119--139.

\bibitem[Friedman, 2002]{friedman2002stochastic}
Friedman, J.~H. (2002).
\newblock Stochastic gradient boosting.
\newblock {\em Computational Statistics \&amp; Data Analysis}, 38(4):367--378.

\bibitem[Garrett and Stevenson, 2020]{pattern}
Garrett, B. and Stevenson, M. (2020).
\newblock Open risk assessments.
\newblock {\em Behavioral Science \& Law}.
\newblock
  \url{https://sites.law.duke.edu/justsciencelab/2019/09/15/comment-on-pattern-by-brandon-l-garrett-megan-t-stevenson/}.

\bibitem[Gelb et~al., 2018]{changingdist}
Gelb, A., Velazquez, T., Trust, P.~C., and of~America, U.~S. (2018).
\newblock The changing state of recidivism: Fewer people going back to prison.
\newblock {\em The Pew Charitable Trusts}.

\bibitem[Goel et~al., 2016]{goelfrisk}
Goel, S., Rao, J.~M., and Shroff, R. (2016).
\newblock Precinct or prejudice? understanding racial disparities in new york
  city's stop-and-frisk policy.
\newblock {\em Institute of Mathematical Statistics}, 10(1):365–394.

\bibitem[Grove and Meehl, 1996]{grove1996comparative}
Grove, W.~M. and Meehl, P.~E. (1996).
\newblock Comparative efficiency of informal (subjective, impressionistic) and
  formal (mechanical, algorithmic) prediction procedures: The
  clinical--statistical controversy.
\newblock {\em Psychology, Public Policy, and Law}, 2(2):293.

\bibitem[Hanson and Thornton, 2003]{hanson2003notes}
Hanson, R. and Thornton, D. (2003).
\newblock Notes on the development of static-2002.
\newblock {\em Ottawa, Ontario: Department of the Solicitor General of Canada}.

\bibitem[Hardt et~al., 2016]{hardt2016eqodds}
Hardt, M., Price, E., and Srebro, N. (2016).
\newblock Equality of opportunity in supervised learning.
\newblock In {\em Advances in neural information processing systems}, pages
  3315--3323.

\bibitem[Harris and Rice, 2008]{vrag}
Harris, G.~T. and Rice, M.~E. (2008).
\newblock {\em Encyclopedia of Psychology and Law}, chapter Violence Risk
  Appraisal Guide (VRAG), page 848.
\newblock SAGE Publications, Inc.

\bibitem[Hart, 1924]{hart1924parole}
Hart, H. (1924).
\newblock Predicting parole success.
\newblock {\em Journal of Criminal Law and Criminology}, 14.

\bibitem[Hoffman and Adelberg, 1980]{hoffman1980salient}
Hoffman, P.~B. and Adelberg, S. (1980).
\newblock The salient factor score: A nontechnical overview.
\newblock {\em Fed. Probation}, 44:44.

\bibitem[Howard et~al., 2009]{howard2009ogrs}
Howard, P., Francis, B., Soothill, K., and Humphreys, L. (2009).
\newblock {OGRS} 3: The revised offender group reconviction scale.
\newblock Technical report, Ministry of Justice.

\bibitem[James, 2018]{james2018congressional}
James, N. (2018).
\newblock Risk and needs assessment in the federal prison system.
\newblock Technical report, Congressional Research Service.

\bibitem[Kehl et~al., 2017]{kehl2017}
Kehl, D., Guo, P., and Kessler, S. (2017).
\newblock Algorithms in the criminal justice system: Assessing the use of risk
  assessments in sentencing.
\newblock {\em {}}.
\newblock \url{https://cyber.harvard.edu/publications/2017/07/Algorithms}.

\bibitem[Kim et~al., 2016]{totalarrestrate}
Kim, J., Bushway, S., and Tsao, H. (2016).
\newblock Identifying classes of explanation for crime drop: Period and cohort
  effects for new york state.
\newblock {\em Journal of Quantitative Criminology}, 32:357--375.

\bibitem[Kleiman et~al., 2007]{kleiman2007agedist}
Kleiman, M., Ostrom, B.~J., and Cheesman, F.~L. (2007).
\newblock Using risk assessment to inform sentencing decisions for nonviolent
  offenders in virginia.
\newblock {\em Crime \& Delinquency}, 53(1):106--132.

\bibitem[Kleinberg et~al., 2017]{kleinberginherent}
Kleinberg, J., Mullainathan, S., and Raghavan, M. (2017).
\newblock Inherent trade-offs in the fair determination of risk scores.
\newblock In {\em Proceedings of the 8th Conference on Innovations in
  Theoretical Computer Science}.

\bibitem[Lakkaraju and Rudin, 2017]{lakkarajurudin2017}
Lakkaraju, H. and Rudin, C. (2017).
\newblock {Learning Cost-Effective and Interpretable Treatment Regimes}.
\newblock In Singh, A. and Zhu, J., editors, {\em Proceedings of the 20th
  International Conference on Artificial Intelligence and Statistics},
  volume~54 of {\em Proceedings of Machine Learning Research}, pages 166--175,
  Fort Lauderdale, FL, USA. PMLR.
\newblock \url{http://proceedings.mlr.press/v54/lakkaraju17a.html}.

\bibitem[Larson et~al., 2016]{LarsonMaKiAn16}
Larson, J., Mattu, S., Kirchner, L., and Angwin, J. (2016).
\newblock How we analyzed the {COMPAS} recidivism algorithm.
\newblock Technical report, ProPublica.
\newblock
  \url{https://www.propublica.org/article/how-we-analyzed-the-compas-recidivism-algorithm}.

\bibitem[Latessa et~al., 2009]{ohiorisk}
Latessa, E., Smith, P., Lemke, R., Makarios, M., and Lowenkamp, C. (2009).
\newblock Creation and validation of the ohio risk assessment system.
\newblock Technical report, University of Cincinnati School of Criminal Justice
  Center for Criminal Justice Research.

\bibitem[Lazarsfeld, 1974]{vera}
Lazarsfeld, P.~F. (1974).
\newblock An evaluation of the pretrial services agency of the vera institute
  of justice.
\newblock {\em New York: Vera Institute}.

\bibitem[Lou et~al., 2013]{loucaruana2013}
Lou, Y., Caruana, R., Gehrke, J., and Hooker, G. (2013).
\newblock Accurate intelligible models with pairwise interactions.
\newblock In {\em 19th ACM SIGKDD International Conference on Knowledge
  Discovery and Data Mining {(KDD)}}, pages 623--631.
\newblock DOI: 10.1145/2487575.2487579.

\bibitem[Ludwig and Mullainathan, 2021]{ludwig21fragile}
Ludwig, J. and Mullainathan, S. (2021).
\newblock Fragile algorithms and fallible decision-makers: Lessons from the
  justice system.
\newblock {\em Journal of Economic Perspectives}, 35(4):71--96.
\newblock \url{https://www.aeaweb.org/articles?id=10.1257/jep.35.4.71}.

\bibitem[Matthews and Minton, 2017]{agecrime}
Matthews, B. and Minton, J. (2017).
\newblock Rethinking one of the criminology's 'brute facts': The age-crime
  curve and the crime drop in scotland.
\newblock {\em European Journal of Criminology}, 15(3):296--320.

\bibitem[{MHS Assessments}, 2017]{LSI2017brochure}
{MHS Assessments} (2017).
\newblock Level of service/case management inventory: An offender management
  system.
\newblock {\em MHS Public Safety}.
\newblock
  \url{https://issuu.com/mhs-assessments/docs/ls-cmi.lsi-r.brochure\_insequence}.

\bibitem[Milgram, 2014]{milgram2014ted}
Milgram, A. (2014).
\newblock Why smart statistics are the key to fighting crime.

\bibitem[Mishra, 2014]{climate1}
Mishra, A. (2014).
\newblock Climate and crime.
\newblock {\em Global Journal of Science Frontier Research: H, Environment \&
  Earth Science}, 14.

\bibitem[Nafekh and Motiuk, 2002]{nafekh2002statistical}
Nafekh, M. and Motiuk, L.~L. (2002).
\newblock {\em The Statistical Information on Recidivism, Revised 1 (SIR-R1)
  Scale: A Psychometric Examination}.
\newblock Correctional Service of Canada. Research Branch.

\bibitem[Neuilly et~al., 2011]{neuilly2011predicting}
Neuilly, M.-A., Zgoba, K.~M., Tita, G.~E., and Lee, S.~S. (2011).
\newblock Predicting recidivism in homicide offenders using classification tree
  analysis.
\newblock {\em Homicide studies}, 15(2):154--176.

\bibitem[Northpointe, 2013]{compas}
Northpointe (2013).
\newblock {\em Practitioner's Guide to COMPAS Core}.
\newblock
  \url{http://www.northpointeinc.com/downloads/compas/Practitioners-Guide-COMPAS-Core-_031915.pdf}.

\bibitem[{Northpointe Inc.}, 2009]{compasquestionnaire}
{Northpointe Inc.} (2009).
\newblock Measurement \& treatment implications of {COMPAS} core scales.
\newblock Technical report, Northpointe Inc.

\bibitem[of~Pretrial~Services, 2015]{cpat}
of~Pretrial~Services, C.~A. (2015).
\newblock The colorado pretrial assessment tool (cpat): Administration,
  scoring, and reporting manual.
\newblock
  \url{https://university.pretrial.org/HigherLogic/System/DownloadDocumentFile.ashx?DocumentFileKey=47e978bb-3945-9591-7a4f-77755959c5f5}.

\bibitem[O'Neil, 2016]{oneil2016weapons}
O'Neil, C. (2016).
\newblock {\em Weapons of Math Destruction}.
\newblock Crown Books.

\bibitem[Orbis, 2014]{spin}
Orbis (2014).
\newblock Service planning instrument: An innovative assessment and case
  planning tool.
\newblock
  \url{https://orbispartners.com/wp-content/uploads/2014/07/SPIn-Brochure.pdf}.

\bibitem[Pleiss et~al., 2017]{pleiss2017calib}
Pleiss, G., Raghavan, M., Wu, F., Kleinberg, J., and Weinberger, K. (2017).
\newblock On fairness and calibration.
\newblock In {\em Advances in Neural Information Processing Systems}, pages
  5680--5689.

\bibitem[{Pretrial Justice Institute}, 2020]{pji2020}
{Pretrial Justice Institute} (2020).
\newblock Updated position on pretrial risk assessment tools.
\newblock {\em Pretrial Justice Institute}.
\newblock
  \url{https://university.pretrial.org/viewdocument/updated-statement-on-pretrial-risk}.

\bibitem[{Public Safety Assessment}, 2019]{psaabout}
{Public Safety Assessment} (2019).
\newblock Risk factors and formulas.
\newblock {\em Laura and John Arnold Foundation}, {}.
\newblock \url{https://www.psapretrial.org/about/}.

\bibitem[Ranson, 2014]{climate2}
Ranson, M. (2014).
\newblock Crime, weather, and climate change.
\newblock {\em Journal of Environmental Economics and Management}, 67.

\bibitem[Richard, 2019]{berk2019}
Richard, B. (2019).
\newblock Accuracy and fairness for juvenile justice risk assessments.
\newblock {\em Journal of Empirical Legal Studies}, 16:174--194.

\bibitem[Roberts and von Hirsch, 2010]{sentencingbook}
Roberts, J. and von Hirsch, A. (2010).
\newblock {\em Previous Convictions at Sentening - Theoretical and Applied
  Perspective}.
\newblock Bloomsbury Publishing.

\bibitem[Rudin, 2019]{Rudin18}
Rudin, C. (2019).
\newblock Stop explaining black box machine learning models for high stakes
  decisions and use interpretable models instead.
\newblock {\em Nature Machine Intelligence}, 1:206--215.

\bibitem[Rudin et~al., 2020]{Rudin19AgeofUnfairness}
Rudin, C., Wang, C., and Coker, B. (2020).
\newblock The age of secrecy and unfairness in recidivism prediction.
\newblock {\em Harvard Data Science Review}, 2(1).
\newblock \url{https://hdsr.mitpress.mit.edu/pub/7z10o269}.

\bibitem[Sherman, 2007]{sherman2007power}
Sherman, L.~W. (2007).
\newblock The power few: experimental criminology and the reduction of harm.
\newblock {\em Journal of Experimental Criminology}, 3(4):299--321.

\bibitem[Singh and Mohapatra, 2021]{singh21riskassess}
Singh, A. and Mohapatra, S. (2021).
\newblock Development of risk assessment framework for first time offenders
  using ensemble learning.
\newblock {\em IEEE Access}, 9:135024--135033.

\bibitem[Skeem et~al., 2020]{skeem}
Skeem, J., Lin, Z., Jung, J., and Goel, S. (2020).
\newblock The limits of human predictions of recidivism.
\newblock {\em Science Advances}, 6.

\bibitem[Smith, 2016]{auditDNN}
Smith, B. (2016).
\newblock {\em Auditing Deep Neural Networks to Understand Recidivism
  Predictions}.
\newblock PhD thesis, Haverford College.

\bibitem[Soares and Angelov, 2019]{fairbydesign}
Soares, E. and Angelov, P.~P. (2019).
\newblock Fair-by-design explainable models for prediction of recidivism.
\newblock {\em ArXiv}, abs/1910.02043.

\bibitem[Starr, 2015]{starr2015riskassessment}
Starr, S.~B. (2015).
\newblock The risk assessment era: An overdue debate.
\newblock {\em Federal Sentencing Reporter}, 27:205--206.

\bibitem[Stevenson, 2018]{mstevenson}
Stevenson, M. (2018).
\newblock Assessing risk assessment in action.
\newblock {\em Minnesota Law Review}.
\newblock
  \url{http://www.minnesotalawreview.org/wp-content/uploads/2019/01/13Stevenson_MLR.pdf}.

\bibitem[Stevenson and Slobogin, 2018]{StevensonSl18}
Stevenson, M.~T. and Slobogin, C. (2018).
\newblock Algorithmic risk assessments and the double-edged sword of youth.
\newblock {\em Washington University Law Review}, 96(18-36).

\bibitem[{The Leadership Conference on Civil and Human Rights},
  2018]{civilrightsstatement}
{The Leadership Conference on Civil and Human Rights} (2018).
\newblock The use of pretrial "risk assessment" instrument: A shared statement
  of civil rights concerns.
\newblock {\em {}}.
\newblock
  \url{http://civilrightsdocs.info/pdf/criminal-justice/Pretrial-Risk-Assessment-Full.pdf}.

\bibitem[Tollenaar and {van der Heijden}, 2013]{tollenaar2013method}
Tollenaar, N. and {van der Heijden}, P. (2013).
\newblock Which method predicts recidivism best?: a comparison of statistical,
  machine learning and data mining predictive models.
\newblock {\em Journal of the Royal Statistical Society: Series A (Statistics
  in Society)}, 176(2):565--584.

\bibitem[Turner et~al., 2009]{turner2009development}
Turner, S., Hess, J., and Jannetta, J. (2009).
\newblock Development of the {California Static Risk Assessment Instrument}
  ({CSRA}).
\newblock {\em CEBC Working Papers}.

\bibitem[{United States Census Bureau}, 2015]{broward}
{United States Census Bureau} (2015).
\newblock Hispanic or latino origin by race 2011-2015 american community survey
  5-year estimates.
\newblock {\em {}}.
\newblock
  \url{https://factfinder.census.gov/faces/tableservices/jsf/pages/productview.xhtml?pid=ACS_15_5YR_B03002&prodType=table}.

\bibitem[{United States Census Bureau}, 2019]{kentucky}
{United States Census Bureau} (2019).
\newblock Quickfacts kentucy united states.
\newblock {\em {}}.
\newblock \url{https://www.census.gov/quickfacts/fact/table/KY,US/PST04521}.

\bibitem[Ustun and Rudin, 2015]{ustun2015slim}
Ustun, B. and Rudin, C. (2015).
\newblock Supersparse linear integer models for optimized medical scoring
  systems.
\newblock {\em Machine Learning}, pages 1--43.

\bibitem[Ustun and Rudin, 2017]{UstunRu2017KDD}
Ustun, B. and Rudin, C. (2017).
\newblock Optimized risk scores.
\newblock In {\em Proceedings of the 23rd {ACM} {SIGKDD} International
  Conference on Knowledge Discovery and Data Mining}.

\bibitem[Ustun and Rudin, 2019]{UstunRuRiskSlimJMLR19}
Ustun, B. and Rudin, C. (2019).
\newblock Learning optimized risk scores.
\newblock {\em Journal of Machine Learning Research}, 20(150):1--75.
\newblock \url{http://jmlr.org/papers/v20/18-615.html}.

\bibitem[Vapnik and Chervonenkis, 1964]{vapnik1964svm}
Vapnik, V. and Chervonenkis, A. (1964).
\newblock A note on one class of perceptrons.
\newblock {\em Automation and Remote Control}, 25.

\bibitem[Verma and Rubin, 2018]{verma2018fairnessreview}
Verma, S. and Rubin, J. (2018).
\newblock Fairness definitions explained.
\newblock In {\em ACM/IEEE International Workshop on Software Fairness}, pages
  1--7. ACM.

\bibitem[{Virginia Department of Criminal Justice Services}, 2018]{vprai}
{Virginia Department of Criminal Justice Services} (2018).
\newblock Virginia pretrial risk assessment instrument - (vprai).
\newblock \url
  {https://www.dcjs.virginia.gov/sites/dcjs.virginia.gov/files/publications/corrections/virginia-pretrial-risk-assessment-instrument-vprai_0.pdf}.

\bibitem[Wexler, 2017]{nyt-computers-crim-justice}
Wexler, R. (2017).
\newblock When a computer program keeps you in jail: How computers are harming
  criminal justice.
\newblock {\em New York Times}, page~27.
\newblock Section A.

\bibitem[Wolfgang, 1987]{wolfgang1987delinquency}
Wolfgang, M.~E. (1987).
\newblock {\em Delinquency in a birth cohort}.
\newblock University of Chicago Press.

\bibitem[W.Palocsay et~al., 2000]{paloscayneural}
W.Palocsay, S., PingWang, and Brookshire, R.~G. (2000).
\newblock Predicting criminal recidivism using neural networks.
\newblock {\em Socio-Economic Planning Sciences}, 34:271--284.

\bibitem[Zemel et~al., 2013]{zemel2013fairrepr}
Zemel, R., Wu, Y., Swersky, K., Pitassi, T., and Dwork, C. (2013).
\newblock Learning fair representations.
\newblock In {\em International Conference on Machine Learning}, pages
  325--333.

\bibitem[Zeng et~al., 2017]{ZengUsRu2017}
Zeng, J., Ustun, B., and Rudin, C. (2017).
\newblock Interpretable classification models for recidivism prediction.
\newblock {\em Journal of the Royal Statistical Society: Series {A} (Statistics
  in Society)}, 180(3):689--722.

\bibitem[Zweig, 2010]{zweig2010bail}
Zweig, J. (2010).
\newblock Extraordinary conditions of release under the bail reform act.
\newblock {\em Harvard Journal On Legislation}, 47:555--585.

\end{thebibliography}

%%%%%%%%%%%%%%%%%%%%%%
\newpage
\section{Appendix} \label{App}

\subsection{Nested Cross Validation Procedure}\label{nestedcv}

We applied five-fold nested cross validation to tune parameters. We split the entire data set into five equally-sized folds for the outer cross validation step. One fold was used as the holdout test set and the other four folds were used as the training set (call it ``outer training set''). The inner loop deals only with the outer training set ($\frac{4}{5}$ths of the data). On this outer training set, we conducted five-fold cross validation and grid-searched hyperparameter values. After this point, each hyperparameter value had five validation results. We selected the parameter values with the highest average validation results and then trained the model with this best set of parameters on the entire outer training set and tested it on the holdout test set. 

We repeated the process above until each one of the original five folds was used as the holdout test set. Ultimately, we had five holdout test results, with which we were able to calculate the average and standard deviation of the performance. 

We applied a variant of the nested cross validation procedure described above to perform the analysis discussed in Section \ref{Generalization}---where we  trained models on one region and tested on the other region. For instance, when we trained models on Broward and tested them on Kentucky, the Kentucky data was treated as the holdout test set. We split the Broward data into five folds and used four folds to do cross validation and constructed the final model using the best parameters. We then tested the final model on the entire Kentucky data set, as well as the holdout test set from Broward. We rotated the four folds and repeated the above process five times. 

%%%%%%%%%%%%%%%%%%%%%%%%%%%%%%%%%%%%%%%%%%%

\subsection{Broward Data Processing}\label{broward-processing}

The Broward County data set consists of publicly available criminal history, court data and COMPAS scores from Broward County, Florida. The criminal history and demographic information were computed from raw data released by ProPublica \citep{AngwinLaMaKi16}. The probational history was computed from public criminal records released by the Broward Clerk’s Office. 

The screening date is the date on which the COMPAS score was calculated. The features and labels were computed for an individual with respect to a particular screening date. For individuals who have multiple screening dates, we compute the features for each screening date, such that the set of events for calculating features for earlier screening dates is included in the set of events for later screening dates. On occasion, an individual will have multiple COMPAS scores calculated on the same date. There  appears  to  be  no  information  distinguishing  these  scores  other  than  their identification number, so we take the scores with the larger identification number. The recidivism labels were computed for the timescales of six months and two years. Some individuals were sentenced to prison as a result of their offense(s). We used only observations for which we have six months/two years of data subsequent to the individual's release date. 

Below, we describe details of the feature and label generation process. The constructed features are presented in Table \ref{table:fl_features} at the end of this section.
\begin{itemize}
    \item Degree “(0)” charges seem to be very minor offenses, so we exclude these charges. We infer whether a charge is a felony, misdemeanor, or traffic charge based off the charge degree.
    
    \item Some of our features rely on classifying the type of each offense (e.g., whether or not it is a violent offense). We infer this from the statute number, most of which correspond to statute numbers from the Florida state crime code.
    
    \item The raw Propublica data includes arrest data as well as charge data. Because the arrest data does not include the statute, which is necessary for us to determine offense type, we use the charge data to compute features that require the offense type. We use both charge and arrest data to predict recidivism.
    
    \item For each person on each COMPAS screening date, we identify the offense---which we call the current offense---that most likely triggered the COMPAS screening. The current offense date is the date of the most recent charge that occurred on or before the COMPAS screening date. Any charge that occurred on the current offense date is part of the current offense. In some cases, there is no prior charge that occurred near the COMPAS screening date, suggesting charges may be missing from the data set.  For this reason we consider charges  that  occurred  within  30  days  of  the  screening  date  for  computing  the  current offense.  If  there  are  no  charges  in  this  range,  we  say  the  current  offense  is  missing. We exclude observations with missing current offenses.  We used some of the COMPAS subscale items as features for our machine learning models. All such components of the COMPAS subscales that we compute are based on data that occurred prior to (not including) the current offense date.

    \item The events/documents data includes a number of events (e.g., ``File Affidavit Of Defense'' or ``File Order Dismissing Appeal'') related to each case,  and thus to each person. To determine how many prior offenses occurred while on probation, or if the current offense occurred  while  on  probation,  we  define  a  list  of  event  descriptions  indicating  that  an individual was taken on or off probation. Unfortunately, there appear to be missing events, as individuals often have consecutive ``On'' or consecutive ``Off'' events (e.g., two ``On'' events in a row, without an ``Off'' in between). In these cases, or if the first event is an ``Off'' event or the last event is an ``On'' event, we define two thresholds, $t_{on}$ and $t_{off}$. If an offense occurred within $t_{on}$ days after an ``On'' event or $t_{off}$ days before an ``Off'' event, we count the offense as occurring while on probation. We set $t_{on}$ to 365 and $t_{off}$ to 30. On the other hand, the ``number of times on probation'' feature is just the count of ``On'' events and the ``number of times the probation was revoked'' feature is just the count of “File order of Revocation of Probation” event descriptions (i.e., we do not infer missing probation events for these two features).
    
    \item Current age is defined as the age in years, rounded down to the nearest integer, on the COMPAS screening date.

    \item A  juvenile  charge  is  defined  as  an  offense  that  occurred  prior  to  the  defendant’s  18th birthday.
    
    \item Labels and features were computed using charge data.
    
    \item The final data set contains 1,954 records and 41 features.
\end{itemize}

% features
\begin{table}[ht!]
\begin{center}
\scriptsize
\tabcolsep=0.1cm
\renewcommand{\arraystretch}{1.1}
\caption{Features from Broward data set. Recall that charges can be convicted or non-convicted.}
\label{table:fl_features}
    \begin{tabular}{|c|p{6.5cm}|} \hline
    \textbf{Features} & \textbf{Explanation} \\ \hline
    person\_id & unique personal identifier \\ \hline
    sex & biological sex of the person \\ \hline
    race & race of the person \\ \hline
    screening\_date & date that triggered the COMPAS screening \\ \hline
    age\_at\_current\_charge & age at the person's current charge \\ \hline
    age\_at\_first\_charge & age at the person's first charge \\ \hline
    p\_arrest & count of prior arrests \\ \hline
    p\_charges & count of prior charges \\ \hline
    p\_violence & count of prior violent charges  \\ \hline
    p\_felony & count of prior felony-level charges \\ \hline
    p\_misdemeanor & count of prior misdemeanor-level charges \\ \hline
    p\_juv\_fel\_count & count of prior felony-level and juvenile charges \\ \hline
    p\_property & count of prior property-related charges \\ \hline
    p\_murder & count of prior murder charges \\ \hline
    p\_famviol & count of prior family violence charges \\ \hline
    p\_sex\_offenses & count of prior sex offense charges \\ \hline
    p\_weapon & count of prior weapon-related charges \\ \hline
    p\_felprop\_viol & count of prior felony-level, property-related, and violent charges \\ \hline
    p\_felassault & count of prior felony-level assault charges \\ \hline
    p\_misdeassault & count of prior misdemeanor-level assault charges \\ \hline
    p\_traffic & count of prior traffic-related charges \\ \hline
    p\_drug & count of prior drug-related charges \\ \hline
    p\_dui & count of prior DUI charges \\ \hline
    p\_stalking & count of prior stalking charges \\ \hline
    p\_voyeurism & count of prior voyeurism charges \\ \hline
    p\_fraud & count of prior fraud charges \\ \hline
    p\_stealing & count of prior stealing/theft charges \\ \hline
    p\_domestic & count of prior domestic violence charges \\ \hline
    p\_trespass & count of prior trespass charges \\ \hline
    p\_fta\_two\_year & count of prior failures to appear in court within \par last two years ($\leq 2$ years)\\ \hline
    p\_fta\_two\_year\_plus & count of prior failures to appear in court beyond \par last two years ($> 2$ years)\\ \hline
    p\_pending\_charge & count of times charged with a new offense when there was \par a pending case \\ \hline
    p\_probation & count of times charged with a new offense when the person was \par on probation \\ \hline
    p\_incarceration & whether or not the person was formerly sentenced to incarceration \\ \hline
    six\_month & whether or not the person had charges within \par last six months ($\leq 6$ months) \\ \hline
    one\_year & whether or not the person had charges within \par last year ($\leq 1$ year)\\ \hline
    three\_year & whether or not the person had charges within \par last three years ($\leq 3$ years)\\ \hline
    five\_year & whether or not the person had charges within \par last five years ($\leq 5$ years)\\ \hline
    current\_violence & whether or not the current charge is violent \\ \hline
    current\_violence20 & whether or not the current charge is violent and the \par person is $\leq 20$ years old \\ \hline 
    total\_convictions & total count of convictions  \\ \hline
    \end{tabular}
\end{center}
\end{table}

%% kentucky data processing
\subsection{Kentucky Data Processing}\label{kentucky-processing}

The Kentucky pretrial and criminal court data was provided by the Department of Shared Services, Research and Statistics in Kentucky. The Pretrial Services Information Management System (PRIM) data contains records regarding defendants, interviews, PRIM cases, bonds etc., that are connected with the pretrial services' interviews conducted between July 1, 2009 and June 30, 2018. The cases were restricted to have misdemeanor, felony, and other level charges. The data from another system, CourtNet, provided further information about cases, charges, sentences, dispositions etc. for CourtNet cases matched in the PRIM system. The Kentucky data can be accessed through a special data request to the Kentucky Department of Shared Services, Research and Statistics. Please refer to Table \ref{table:ky-raw-data} for all the raw datasets we processed, together with their sizes and general information provided.

CourtNet and PRIM data were processed separately and then combined together. We describe the details below. The constructed features are presented in Table \ref{table:ky_features} at the end of this section.

\begin{itemize}
    \item For the CourtNet data, we filtered out cases with filing date prior to Jan. 1st, 1996, which were claimed to be less reliable records by the Kentucky Department of Shared Services, Research and Statistics (which provided the data). To investigate what types of crimes the individuals were involved in for each charge, such as drug, property, traffic-related crime, we used the Kentucky Uniform Crime Reporting Code (UOR Code), as well as detecting keywords in the UOR description. 
    
    \item From the PRIM system data, we extracted the probation, failure to appear, case pending, and violent charge information at the PRIM case level, as well as the Arnold PSA risk scores computed at the time of each pretrial services' interview. Since Kentucky did not use Arnold PSA until July 1st, 2013, we filtered out records before the this date. We omitted records without risk scores since we want to compare the performance of the PSA with other models. Only 33 records are missing PSA scores; therefore we do not worry about missing records impacting the results. Additionally, some cases in the PRIM system have ``indictment'' for the arrest type, along with an  ``original'' arrest case ID, indicating that those cases were not new arrests. We matched these cases with the records that correspond to the original arrests to avoid overcounting the number of prior arrests. Then we inner-joined the data from the two systems using person-id and prim-case-id.
    
    \item For each individual, we used the date that is two years before the latest charge date in the Kentucky data, as a cutoff date. The data before the cutoff are used as criminal history information to compute features. The data after the cutoff are used to compute labels and check recidivism. In the data before the cutoff, the latest charge is treated as the current charge (i.e., the charge that would trigger a risk-assessment) for each individual. We compute features and construct labels using only convicted charges. However, the current charge can be either convicted or non-convicted. This ensures that our analysis includes all individuals that would receive a risk assessment, even if they were later found innocent of the current charge that triggered the risk assessment. It also ensures that criminal history features use only convicted charges, so that our risk assessments are not influenced by charges for crimes that the person may not have committed.
    
    \item In order to compute the labels, we must ensure that there are at least two years of data following an individual's current charge date. For individuals who are sentenced to prison due to their current charge, we consider their \textit{release date} instead of the current charge date. We omitted individuals for whom there were less than two years of data between their current charge date or release date, and the last date recorded in the data set. 
    
    \item To get the age at current charge information, we first calculated the date of birth (DOB) for each individual, using CourtNet case filing date and age \textit{at the CourtNet case filing date}. Then we calculated ``age at current charge'' using the DOB and charge date (the charge date sometimes differs from the case filing date). Notice that there are many errors in age records in the data. For instance, some people have age recorded over 150, which is certainly wrong but there is no way to correct it. To ensure the quality of our data, we limited the final current age feature to be inclusively between 18 and 70. This is also consistent with the range from Broward analysis. If the person was not sentenced to prison, we define current age as the age at current charge date. If the person was sentenced to prison, we compute current age by adding the sentence time to the age at the current charge date. Note that this differs from the way risk scores are computed in practice---usually risk scores are computed prior to the sentencing decision. This helps to handle distributional shift between the individuals with no prison sentence (for whom a 2-year evaluation can be handled directly) and the full population (some of whom may have been sentenced to prison and cannot commit a crime during their sentence). 
    
    \item We computed features using the data before the current charge date. The CourtNet data is organized by CourtNet cases, and each CourtNet case has charge level data. The PRIM data is organized by PRIM cases. Each CourtNet case can connect to multiple PRIM cases. This occurs because a new PRIM case is logged when an update occurs in the defendant's CourtNet case --- for example, if the defendant fails to appear in court.  
    Therefore, to compute the criminal history information, we first grouped on PRIM case level to summarize the charge information. Next, we grouped on CourtNet case level to summarize PRIM case level information. Last, we grouped on the individual level to summarize the criminal histories. 
    
    \item On computing the ADE feature: The ADE feature means number of times the individual was assigned to alcohol and drug education classes. Note that by Kentucky state law, any individual convicted for a DUI is assigned to ADE classes. This does not indicate whether the individual successfully completed ADE classes.
    
    \item We compute labels using the two years of data after the current charge date/release date. We constructed the \texttt{general} recidivism labels by checking whether a ``convicted charge'' occurred within two years or six months from the current charge (or release date). Then, using the charge types of the convicted charge, other recidivism prediction labels were generated, such as drug or property-related recidivism. The final data set contains 250,778 records and 40 features.
    
    \textit{Note: there are degrees of experimenter freedom in some of these data processing choices; exploring all the possible choices here is left for future studies.}    
\end{itemize}

The Arnold PSA features that were included in the Kentucky data set (e.g., prior convictions, prior felony convictions etc.) were computed by pretrial officers who had access to criminal history data from both inside and outside of Kentucky. However, the Kentucky data set we received contained criminal history information from within Kentucky only. Thus, the Arnold PSA features for Kentucky (which are included in our models as well) use both in-state and out-of-state information, but the remaining features (which we compute directly from the Kentucky criminal history data) are limited to in-state criminal history. 

Additionally, we were informed by Kentucky Pretrial Services team that the data set 's sentencing information may not be reliable due to unmeasured confounding, including shock probation and early releases that would allow a prisoner to be released much earlier than the end date of the sentence. Because the sentence could be anywhere from zero days to the full length, we conducted a sensitivity analysis by excluding the sentence information in the data processing, which is equivalent to the assumption that no prison sentence was served. For that analysis, the current age of each individual was calculated to be the age at the current charge, and the prediction labels were generated from new charges within six months (or two years) from the current charge. The sensitivity analysis yielded predictive results that were almost exactly the same as the results in the main text, when the sentence information was used to determine age and prediction interval.

\begin{table*}[h]
\begin{center}
\scriptsize
\tabcolsep=0.1cm
\renewcommand{\arraystretch}{1.2}
    \caption{The table lists raw datasets obtained from the Kentucky Department of Shared Services, Research and Statistics, the number of records within each data frame, and general descriptions of the data.}
    \label{table:ky-raw-data}
        \begin{tabular}{|l|c|p{9cm}|} \hline 
        \textbf{Data} & \textbf{Num. of Records} & \textbf{Information} \\ \hline
        KY\_Recidivism\_Defendants & 1,286,599 & contains unique identifiers for each defendant \\ \hline
        KY\_Recidivism\_Interviews & 1,490,545 & contains arrest information, risk level assessment, etc. \\ \hline
        KY\_Recidivism\_CNet\_Cases & 3,179,421 & contains distinct cases in CourtNet database, including case filing, disposition, etc. \\ \hline
        KY\_Recidivism\_PRIM\_Cases & 1,987,783 & contains distinct cases in PRIM database \\ \hline
        KY\_Recidivism\_CNet\_Charges & 8,683,273 & contains distinct charges within cases in CourtNet database \\ \hline
        KY\_Recidivism\_Sentences & 2,926,446 & contains sentence information for charges disposed with a conviction \\ \hline
        KY\_Recidivism\_Bonds & 2,947,148 & contains pretrial records of bonds and release conditions set by court. \\ \hline
        KY\_Recidivism\_Events & 2,256,330 & contains information about court-mandated events, including the type of event \par and whether or not a defendant was compliant, etc. \\ \hline
        KY\_Recidivism\_FTA & 231,737 & contains specific information about court appearances for which pretrial services \par indicated that the defendant failed to appear (FTA). \\ \hline
        KY\_Recidivism\_Supervision & 42,352 & contains information specific to defendants on monitored pretrial release. \\
        \hline
        \end{tabular}
    \end{center}
\end{table*}

\begin{table}[ht!]
\begin{center}
\scriptsize
\tabcolsep=0.1cm
\renewcommand{\arraystretch}{1.2}
\caption{Features from Kentucky data set. The charges are convicted. ADE means assignment to alcohol and drug education classes.}
\label{table:ky_features}
    \begin{tabular}{|c|p{7cm}|} \hline
    \textbf{Features} & \textbf{Explanation} \\ \hline
    person\_id & unique personal identifier \\ \hline
    sex & biological sex of the person \\ \hline
    race & race of the person \\ \hline
    current\_date & current charge date or the release date if there was a sentence \par on the current charge. \\ \hline
    age\_at\_current\_charge & age at the person's current charge, or the age at current charge \par plus the sentence time if there was a sentence on the current charge\\ \hline
    p\_arrest & count of prior arrests with convicted charges \\ \hline
    p\_charges & count of prior convicted charges \\ \hline
    p\_violence & count of prior violent charges \\ \hline
    p\_felony & count of prior felony-level charges \\ \hline
    p\_misdemeanor & count of prior misdemeanor-level charges \\ \hline
    p\_property & count of prior property-related charges \\ \hline
    p\_murder & count of prior murder charges \\ \hline
    p\_assault & count of prior assault charges \\ \hline
    p\_sex\_offenses & count of prior sex offense charges \\ \hline
    p\_weapon & count of prior weapon-related charges \\ \hline
    p\_felprop\_viol & count of prior felony-level, property-related, and violent charges \\ \hline
    p\_felassault & count of prior felony-level assault charges \\ \hline
    p\_misdeassault & count of prior misdemeanor-level assault charges \\ \hline
    p\_traffic & count of prior traffic-related charges \\ \hline
    p\_drug & count of prior drug-related charges \\ \hline
    p\_dui & count of prior DUI charges \\ \hline
    p\_stalking & count of prior stalking charges \\ \hline
    p\_voyeurism & count of prior voyeurism charges \\ \hline
    p\_fraud & count of prior fraud charges \\ \hline
    p\_stealing & count of prior stealing/theft charges \\ \hline
    p\_trespass & count of prior trespass charges \\ \hline
    ADE & count of times the person was assigned to alcohol/drug education classes \\ \hline
    treatment & count of times the person received treatment along with the sentence \\ \hline
    p\_fta\_two\_year & count of prior failures to appear in court within last two years ($\leq 2$ years)\\ \hline
    p\_fta\_two\_year\_plus & count of prior failures to appear in court beyond last two years ($> 2$ years)\\ \hline
    p\_pending\_charge & count of times charged with a new offense when there was a pending case \\ \hline
    p\_probation & count of times charged with a new offense when the person was on probation \\ \hline
    p\_incarceration & whether or not the person was formerly sentenced to incarceration  \\ \hline
    six\_month & whether or not the person had charges within last six months ($\leq 6$ months) \\ \hline
    one\_year & whether or not the person had charges within last year ($\leq 1$ year)\\ \hline
    three\_year & whether or not the person had charges within last three years ($\leq 3$ years)\\ \hline
    five\_year & whether or not the person had charges within last five years ($\leq 5$ years)\\ \hline
    current\_violence & whether or not the current charge was violent \\ \hline
    current\_violence20 & whether or not the current charge was violent and the person was $\leq 20$ years old \\ \hline 
    current\_pending\_charge & whether or not the person had a pending case during the current charge \\ \hline
    \end{tabular}
\end{center}
\end{table}

\subsection{Why We Compare Only Against COMPAS and the PSA}\label{other-tools} 

The variables included in risk assessments are often categorized into \textit{static} and \textit{dynamic} factors. Static factors are defined as factors that cannot be reduced over time (e.g. criminal history, gender, and age-at-first-arrest). Dynamic factors are defined as variables that can change over time to decrease the risk of recidivism; they allow insight into whether a high-risk individual can lower their risk through rehabilitation, and sometimes improve prediction accuracy. Examples of dynamic factors include current age, treatment for substance abuse, and mental health status \citep{kehl2017}.
Dynamic factors are often included in \textit{risk-and-needs-assessments} (RNAs), which in addition to identifying risk of recidivism, recommend interventions to practitioners  (e.g., treatment programs, social services, diversion of individuals from jail). 

With the exception of current age, our features all fall under the ``static'' classification. This renders us unable to compare against the risk assessment tools that use dynamic factors, whose formulas \textit{are} public. The risk assessments that we examined are listed in Table \ref{table:risk_assessments_comparison}. Since we have only criminal history and age variables, the only model we could compute from our data was the Arnold PSA. 

However, as we demonstrated in the main body of the paper, the fact that we do not possess dynamic factors is not necessarily harmful to the predictive performance of our models. The goal behind including dynamic factors in models is to improve prediction accuracy as well as be able to recommend interventions that reduce the probability of recidivism. While an admirable goal, the inclusion of dynamic factors does not come at zero cost and may not actually produce performance gains for recidivism prediction. In Sections \ref{Baselines} and \ref{Interpretable}, we show that standard machine learning techniques (using only the static factors) and interpretable machine learning models (using only static factors) are able to outperform a criminal justice model that utilizes both static and dynamic factors (COMPAS). Furthermore, the inclusion of additional, unnecessary factors increases the risk of data entry errors, or exposes models to additional feature bias \citep{corbett-daviesmeasure}. As  \citet{Rudin19AgeofUnfairness} reveals, data entry errors appear to be common in COMPAS score  calculations and could lead to scores that are either too high or too low.

Although the COMPAS suite is a proprietary (and thus black-box) risk-and-needs assessment, we were still able to compare against its risk assessments thanks to the Florida's strong open-records laws. Created by Northpointe (a subsidiary company of Equivant), COMPAS is a recidivism prediction suite which is used in criminal justice systems throughout the United States. It is comprised of three scores: Risk of General Recidivism, Risk of Violent Recidivism, and Risk of Failure to Appear. In this work, we examine the two risk scores relating to violent recidivism and general recidivism. Each risk score is an integer from one to ten \citep{northpointe}. 

As COMPAS scores are proprietary instruments, the precise forms of its models are not publicly available. However, it is known that the COMPAS scores are computed from a subset of 137 input variables that include vocational/educational status, substance abuse, and probational history, in addition to the standard criminal history variables \citep{northpointe}. As such, we cannot directly compute these risk scores, and instead utilize the COMPAS scores released by ProPublica in the Broward County recidivism data set. We do not compare against COMPAS on the Kentucky data set, as our data set  does not include COMPAS scores. 

The PSA was created by Arnold Ventures, and is a publicly available risk assessment tool. Similar to the COMPAS suite, it is comprised of three risk scores: Failure to Appear, New Criminal Activity, and New Violent Criminal Activity. Again, we compare against latter two scores. Both are additive integer models which take nine factors as input, relating to age, current charge, and criminal history. The New Criminal Activity model outputs a score from 1 to 6, while the New Violent Criminal Activity model outputs a binary score \citep{psaabout}. The PSA is an interpretable model.

\definecolor{Gray}{gray}{0.9}
\newcolumntype{g}{>{\columncolor{Gray}}c}
\newcolumntype{w}{>{\columncolor{white}}c}

\begin{table*}[t]
\begin{center}
\scriptsize
\tabcolsep=0.03cm
\renewcommand{\arraystretch}{1.2}
\caption{Variable comparison for currently-utilized actuarial risk assessments. We only have criminal history and age variables, but most models include many other variables. Abbreviations are: Correctional Offender Management Profiling for Alternative Sanctions (COMPAS); Connecticut Risk Assessment for Pretrial Decision Making (Connecticut); Colorado Pretrial Risk Assessment Tool (CPAT); California Static Risk Assessment (CSRA); Ohio Risk Assessment System (ORAS); Level-of-Service Case Management Inventory (LSI-CMR); Public Service Assessment(PSA); (Federal) Pretrial Risk Assessment (PTRA); Statistical Information on Recidivism Score (SIRS); Service Planning Instruments (SPIn); Vera Point Scale (VERA); Violence Risk Appraisal Guide (VRAG); Virginia Pretrial Risk Assessment Instrument (VPRAI).}

\label{table:risk_assessments_comparison}
\begin{tabular}{|w|g|g|w|w|w|w|w|w|w|w|} \hline 

    Models & Criminal History & Age & Finance & Residential Info & Edu/Emp & Peer/Family & Mental Health & Alc/Subs Abuse & Other \\ \hline\hline
    
    COMPAS \cite{compasquestionnaire} & X & X &  &  & X & X & X & X & X \\ \hline
    Connecticut \cite{conn}& X &  &  &  & X & X & X & X &  \\ \hline
    CPAT \cite{cpat}& X & X &  & X &  &  & X & X & X\\ \hline 
    CSRA \cite{turner2009development}& X & X &  &  &  &  &  & & X\\ \hline
    ORAS \cite{ohiorisk}& X & X &  & X & X &  &  & X & \\ \hline  
    LSI-CMI \cite{LSI2017brochure} & X &  & X &  & X & X &  & X & X \\ \hline
    PSA \cite{psaabout}& X & X &  &  &  &  &  & & \\ \hline
    PTRA \cite{ptra}& X & X & X & X & X & X &  & X & X \\ \hline
    Salient Factor \cite{hoffman1980salient} & X & X &  &  &  &  &  & X & \\ \hline
    SIRS \cite{nafekh2002statistical} & X & X &  & & X &  &  & & X\\ \hline
    SPIn \cite{spin}& X & X &  &  & X & X & X & X & X\\ \hline
    VERA \cite{vera}& X & X &  &  & X & X & X & X & X\\ \hline 
    VRAG \cite{vrag}& X & X &  &  &  & X & X & X & X\\ \hline
    VPRAI \cite{vprai} & X &  &  & X & X &  &  & X & X  \\ \hline

\end{tabular}
\end{center}
\end{table*}

%%%%%%%%%%%%%%%%%%%%%%%%%
\newpage
\subsection{Hyperparameters}\label{hyperparameters}

\subsubsection*{Baseline Models, CART, EBM}

We applied nested cross validation to tune the hyperparameters. Please refer to Table \ref{baselineparameters} for parameter details.

\begin{table}[ht!]
\begin{center}
\tabcolsep=0.5cm
\renewcommand{\arraystretch}{1.4}
\scriptsize
\caption{Hyperparameters for $\ell_1$ and $\ell_2$ Penalized Logistic Regression, Linear SVM, CART, Random Forest, XGBoost, and EBM. RiskSLIM and Additive Stumps are discussed separately.}

\label{baselineparameters}
\begin{tabular}{|p{1.2cm}|p{3.2cm}|p{3.7cm}|} \hline

    \textbf{Models} & \textbf{Kentucky} & \textbf{Broward} \\ \hline\hline
    
    $\ell_2$ Logistic \par Regression &  
    class\_weight: balanced \par solver: liblinear\cite{liblinear} \par penalty: $\ell_2$ \par C \(\in\) [1e-4, 1e-3, 1e-2, 1e-1, 1] &  
    class\_weight: balanced \par solver: liblinear \par penalty: $\ell_2$ \par C \(\in\) 100 values in [1e-5, 1e-2] \\ \hline

    $\ell_1$ Logistic \par Regression & 
    class\_weight: balanced \par solver: liblinear \par penalty: $\ell_1$ \par C \(\in\) [1e-4, 1e-3, 1e-2, 1e-1, 1] & 
    class\_weight: balanced \par solver: liblinear \par penalty: $\ell_1$ \par C \(\in\) 100 values in [1e-5, 1e-2] \\ \hline
    
    LinearSVM & 
    C \(\in\) [1e-4, 1e-3, 1e-2, 1e-1, 1]  & 
    C \(\in\) 100 values in [1e-5, 1e-2] \\ \hline
    
    CART & 
    max\_depth \(\in\) [5,6,7,8,9,10] & 
    max\_depth \(\in\) [1,2,3,4,5] \par  min\_impurity\_decrease \par \(\in\) [1e-3, 2e-3, \dots 5e-3] \\ \hline
    
    Random Forest & 
    n\_estimator \(\in\) [100,150,200] \par max\_depth \(\in\) [7,8,9] & 
    n\_estimator \(\in\) [50,100,200,400,600] \par max\_depth \(\in\) [1,2,3] \par min\_impurity\_decrease \par \(\in\) [1e-3, 2e-3, \dots, 1e-2] \\ \hline
    
    XGBoost & 
    learning\_rate \(\in\) [0.1] \par n\_estimator \(\in\) [100,150] \par max\_depth \(\in\) [4,5,6] & 
    learning\_rate \(\in\) [0.05] \par n\_estimator \(\in\) [50,100,200,400,600] \par max\_depth \(\in\) [1,2,3] \par gamma \(\in\) [6,8,10,12] \par min\_child\_weight \(\in\) [6,8,10,12] \par subsample \(\in\) [0.5]\\ \hline
    
    EBM \tablefootnote{The training procedure is slow for EBM, due to the size of Kentucky data, the nested cross validation we applied, and the cross-validation within the algorithm to choose number of pairwise interactions. Therefore, we tested only one set of parameters, which gave reliable results.} & n\_estimator \(\in\) [60] \par max\_tree\_splits \(\in\) [2] \par learning\_rate \(\in\) [0.1] & 
    n\_estimator \(\in\) [40,60,80,100] \par max\_tree\_splits \(\in\) [1,2,3] \par learning\_rate \(\in\) [0.01] \par holdout\_split \(\in\) [0.7, 0.9] \\ \hline
\end{tabular}
\end{center}
\end{table}

%%%%%%%%%%%%%%%%%%%%%%%%%%%%%%%
\subsubsection*{Additive Stumps}

Stumps were created for each feature as detailed in Section \ref{Stumps}. An additive model was created from the stumps using $\ell_1$-penalized logistic regression, and no more than 15 original features were involved in the additive models. But multiple stumps corresponding to each feature could be used in the models. We chose to limit the size of the model to 15 original features because then at most 15 plots would be generated to visualize the full model, which is a reasonable number of visualizations for users to digest.

We started with the smallest regularization parameter on $\ell_1$ penalty that provides at most 15 original features from the model. This will be our lower bound for nested cross validation. From there, we perform nested cross validation over a grid of regularization parameters, all of which are greater than or equal to the minimum value of the regularization parameter found above. Please refer to Table \ref{stumpsparameters} for more details.

\begin{table}[ht]
\begin{center}
\tabcolsep=0.3cm
\renewcommand{\arraystretch}{1.3}
\scriptsize
\caption{Hyperparameters for Additive Stumps}
\label{stumpsparameters}
\begin{tabular}{|c|p{3cm}|p{3cm}|} \hline

    \multicolumn{3}{c}{\textbf{Kentucky}} \\ \hline
    \textbf{Models} & \textbf{Two Year} & \textbf{Six Month} \\ \hline\hline
    
    %%%
    General &  C \(\in\) [1e-3, 2e-3] &  C \(\in\) [1e-3, 1.5e-3] \\ \hline
    
    %%%
    Violent & C \(\in\) [6e-4, 8e-4, 1e-3] & C \(\in\) [5e-4, 7e-4] \\ \hline
    
    %%%
    Drug & C \(\in\) [1e-3, 2e-3, 2.5e-3] & C \(\in\) [1e-3, 2e-3] \\ \hline

    %%%
    Property & C \(\in\) [1e-3, 1.5e-3] & C \(\in\) [1e-3, 1.5e-3] \\ \hline

    %%%
    Felony & C \(\in\) [1e-3, 1.5e-3] & C \(\in\) [5e-4, 8e-4] \\ \hline

    %%%
    Misdemeanor & C \(\in\) [1e-3, 1.5e-3] & C \(\in\) [5e-4, 1e-3] \\ \hline
    
    \multicolumn{3}{c}{\textbf{Broward}} \\ \hline
    
    %%%
    General & [1e-2, 2e-2\dots 1e-1] &  C \(\in\) [1e-2, 2e-2\dots 1e-1] \\ \hline
    
    %%%
    Violent & C \(\in\) [1e-2, 2e-2 \dots 7e-2] & C \(\in\) [1e-2, 2e-2 \dots 7e-2] \\ \hline
    
    %%%
    Drug & C \(\in\) [1e-2, 2e-2 \dots 9e-2] & C \(\in\) [1e-2, 2e-2 \dots 6e-2] \\ \hline

    %%%
    Property & C \(\in\) [1e-2, 2e-2 \dots 8e-2] & C \(\in\) [1e-2, 2e-2 \dots 6e-2] \\ \hline

    %%%
    Felony & C \(\in\) [1e-2, 2e-2 \dots 8e-2] & C \(\in\) [1e-2, 2e-2 \dots 8e-2] \\ \hline

    %%%
    Misdemeanor & C \(\in\) [1e-2, 2e-2 \dots 7e-2] & C \(\in\) [1e-2, 2e-2 \dots 7e-2] \\ \hline
    
    \multicolumn{3}{l}{\small{All models use "balanced" for the class\_weight, "liblinear" }} \\
    \multicolumn{3}{l}{\small{for the solver, and $\ell_1$ for the penalty.}}
    
\end{tabular}
\end{center}
\end{table}

\subsubsection*{RiskSLIM}

RiskSLIM is challenging to train, because it uses the CPLEX optimization software, which can be difficult to install and requires a license. Moreover, since RiskSLIM solves a very difficult mixed-integer nonlinear optimization problem, it can be slow to prove optimality, which makes it difficult to perform nested cross validation as nested cross validation requires many solutions of the optimization problem. A previous study \citep{auditDNN} also noted similar problems with algorithms that use CPLEX (this study trained on SLIM \citep{ustun2015slim}, which is similar to the training process of RiskSLIM in that they both require CPLEX). Here we provide details of how we trained RiskSLIM to help others use the algorithm more efficiently.  
\begin{itemize}
    \item We ran $\ell_1$-penalized logistic regression on the stumps training data with a relatively large regularization parameter to obtain a small subset of features (that is, we used $\ell_1$-penalized logistic regression for feature selection). Then we trained RiskSLIM using nested cross validation with this small subset of features. The maximum run-time, maximum offset, and penalty value were set to 1,000 seconds, 100, and 1e-6 respectively. The coefficient range was set to [-5, 5], which would give us small coefficients that are easy to add/subtract.
    
    \item If the model converged to optimality (optimality gap less than 5\%) within 1,000 seconds, we then ran $\ell_1$-penalized logistic regression again with a smaller regularization parameter to obtain a slightly larger subset of features to work with. We then trained RiskSLIM with nested cross validation again on this larger subset of features. If RiskSLIM also generated an optimality gap less than 5\% within 1,000 seconds and had better validation performance, we repeated this procedure.  
    
    \item Once either RiskSLIM could not converge to a 5\% optimality gap within 1,000 seconds, or the validation performance did not improve by adding more stumps, we stopped there, using the previously obtained RiskSLIM model as the final model. 
    
    \item This procedure generally stopped with between 12 and 20 stumps from $\ell_1$-penalized logistic regression. Beyond this number of stumps, we did not observe improvements in performance in validation. 
\end{itemize}

%%%%%%%%%%%%%%%%%%%%%%%%%%%%%
\newpage
\subsection{Figures}\label{app-figures}

% prob of recidivism -- violent two year, KY
\begin{figure}[h]
  \centering
  \caption{Probabilities of two-year and six-month \texttt{violent} recidivism, given the age at current charge.}
  \label{fig:age_distribution_violent}
  \includegraphics[width=.6\textwidth]{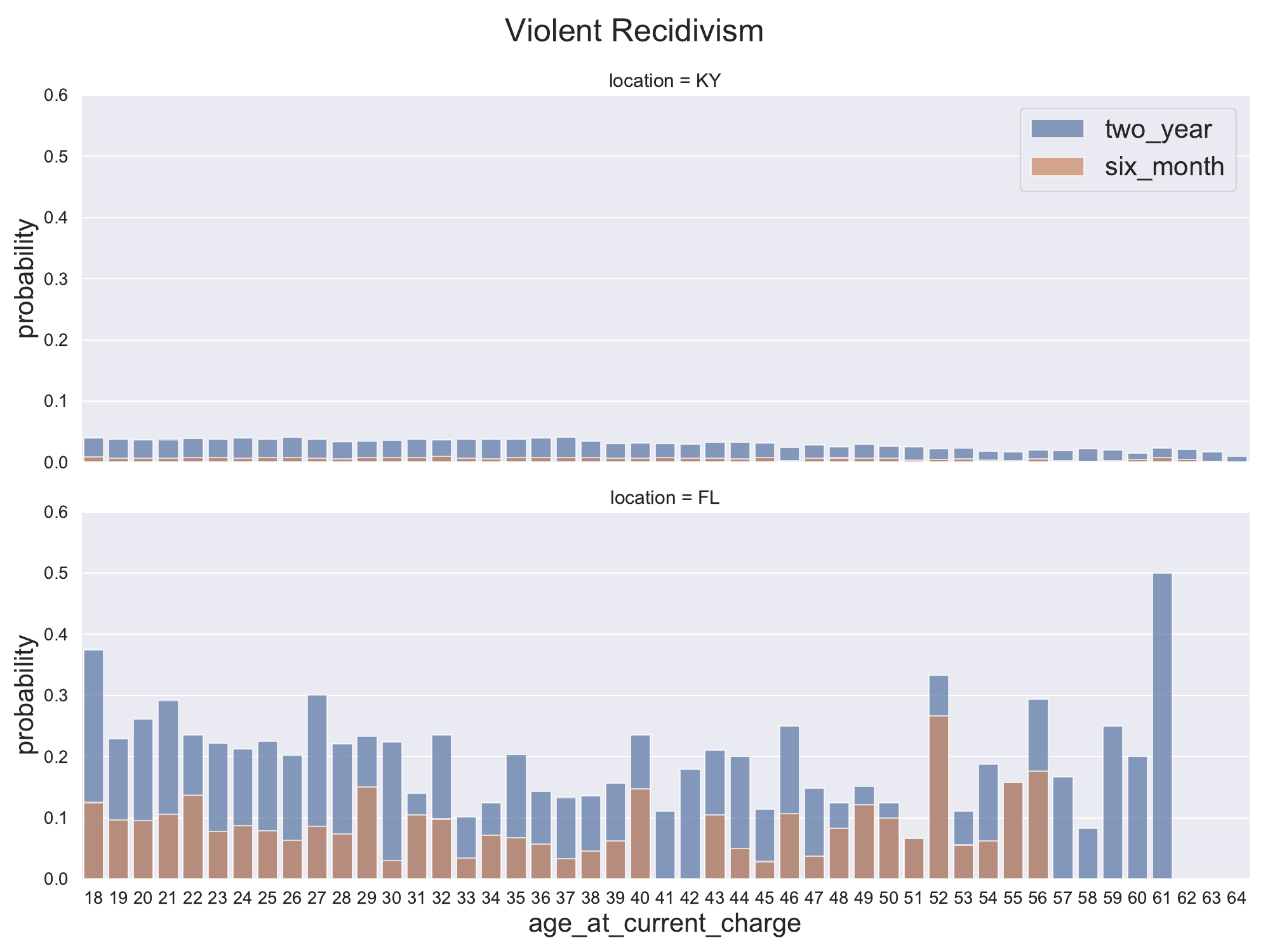}
\end{figure}
% conditional recidivism probability
\begin{figure}[h]
\centering
\caption{Base rates of all twelve types of recidivism on Kentucky data, conditioned (separately) on race and gender.}
\label{fig:data_recid_distributions}
\subfigure{\includegraphics[width=.5\textwidth]{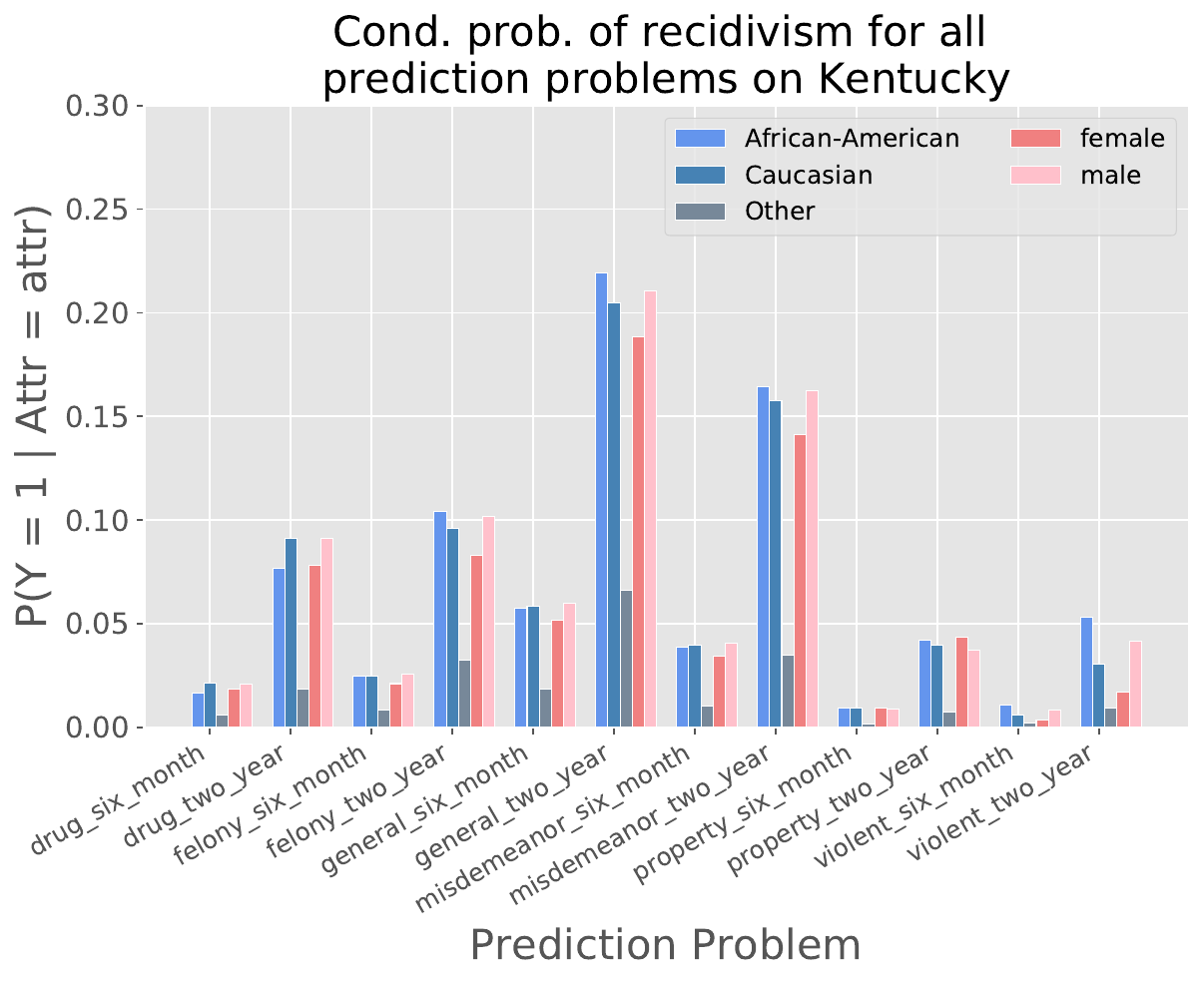}}
\end{figure}

\begin{figure}[ht!]
\centering
\caption{Calibration of the Arnold NVCA Raw, EBM and RiskSLIM for two-year \texttt{violent} recidivism on Kentucky. } \label{fig:calib_kentucky_two_year_violent}
\subfigure[For the Arnold NVCA raw score, the curves satisfy monotonic calibration until the score value of 7, where the probabilities drop to 0. This may be because there are few individuals with an Arnold NVCA raw score equal to 7 in the data. The curves for African-Americans/Caucasians and males/females are close enough to satisfy group calibration (but we note that the African-American (respectively, male) curve is consistently higher than the Caucasian (respectively, female) curve), especially for larger raw NVCA scores.]{
\includegraphics[width=.4\textwidth]{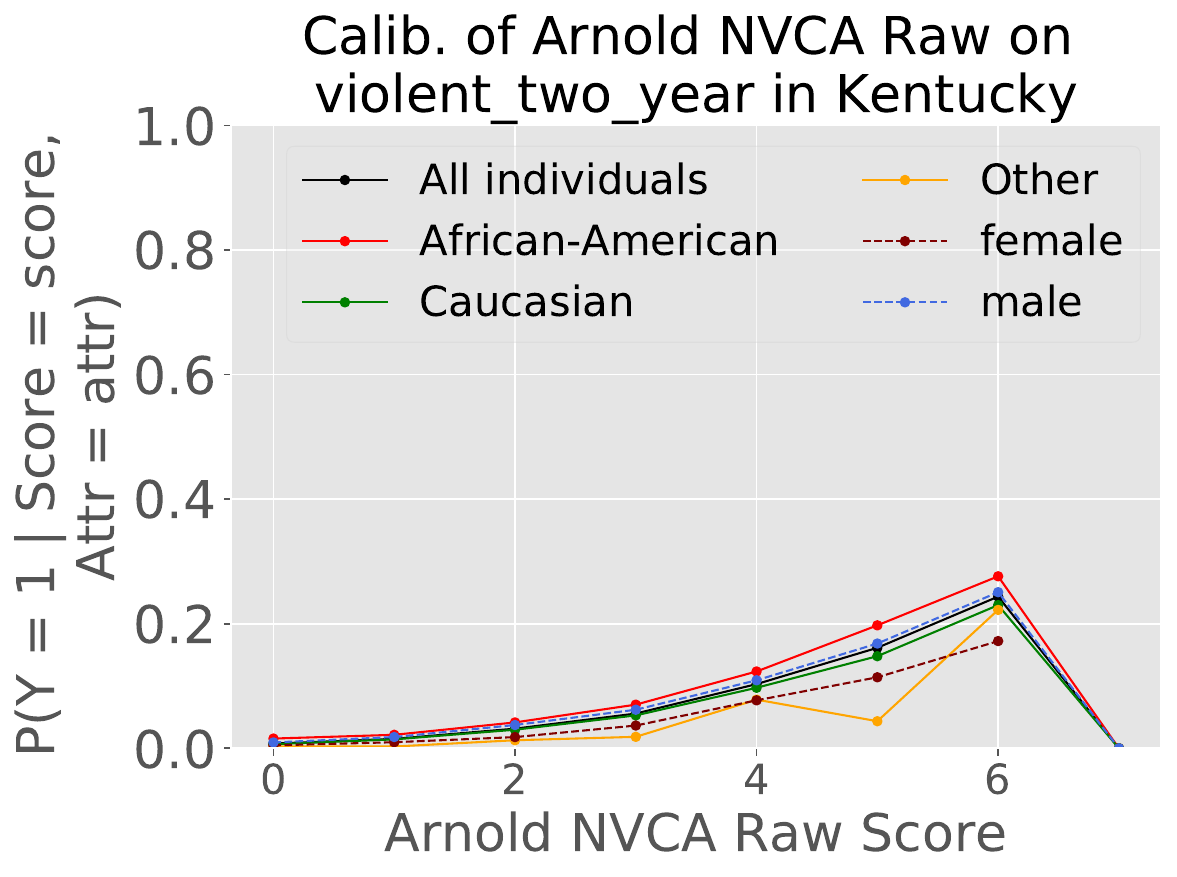}}\;\;\;\;
\subfigure[For EBM, the calibration curves for both gender and race groups are irregular, demonstrating that EBM satisfied neither group calibration nor monotonic calibration, on race and gender groups.]{
\includegraphics[width=.4\textwidth]{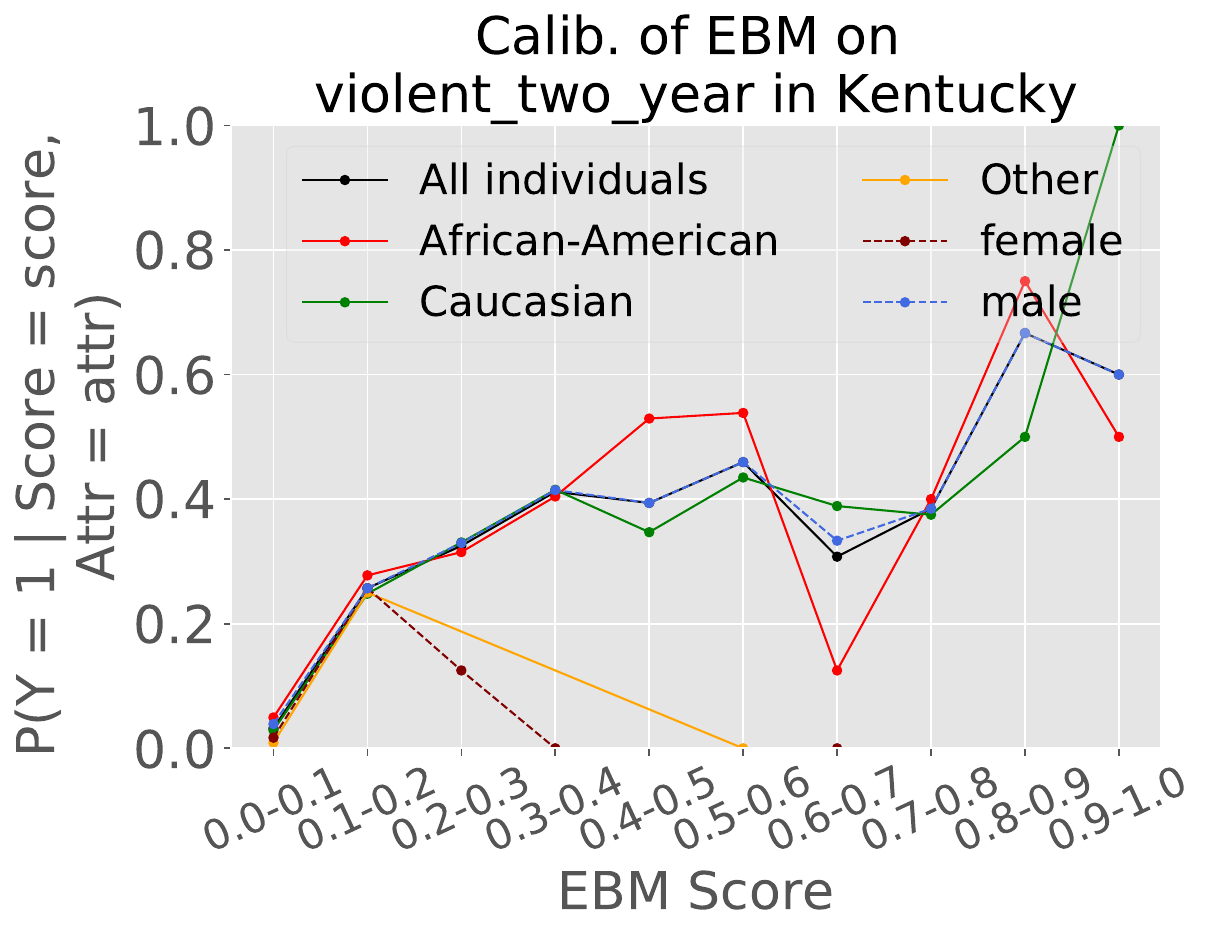}}
\subfigure[For RiskSLIM, the curves are monotonically increasing and roughly overlap with each other. The calibration curve for African-Americans is slightly higher than for the Caucasian and the ``Other'' race groups. For the two gender groups, the curves are close to each other. We conclude that both race and gender approximately satisfy group calibration.]{
    \includegraphics[width=.4\textwidth]{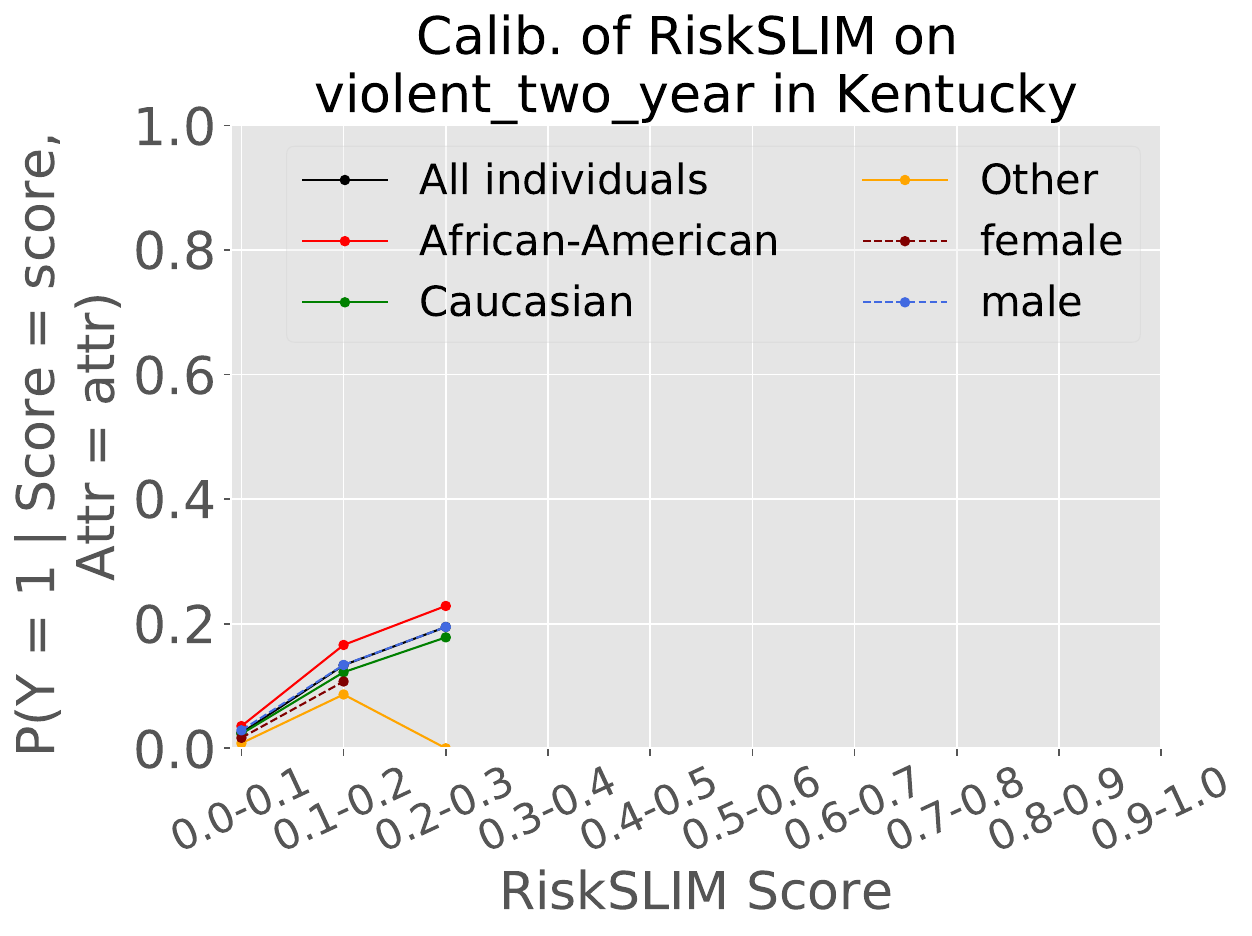}
}
\end{figure}

%%%%%%%%%%%%%%%%%%%%%%%%%%%%%%%%%%%%
\newpage
\subsection{Tables}\label{app-tables}
% KY - two year general, stumps table
\begin{table}[H]
\begin{center}
\scriptsize
\tabcolsep=0.1cm
\renewcommand{\arraystretch}{1.2}
\caption{Additive Stumps on two-year \texttt{general} recidivism. The model consists of twenty-eight stumps with an intercept. These binary features represent fifteen original  features; coefficients were rounded for display purposes only. }
\label{table:ky_stumps_general}
    \begin{tabular}{|l|r|r|} \hline
        1. age at current charge $\le$ 20 & 0.0082 & +... \\ \hline
        2. age at current charge $\le$ 21 & 0.0053 & +... \\ \hline
        3. age at current charge $\le$ 24 & 0.0322 & +... \\ \hline
        4. age at current charge $\le$ 27 & 0.0270 & +... \\ \hline
        5. age at current charge $\le$ 35 & 0.0108 & +... \\ \hline
        6. age at current charge $\le$ 39 & 0.1223 & +... \\ \hline
        7. age at current charge $\le$ 43 & 0.0311 & +... \\ \hline
        8. age at current charge $\le$ 47 & 0.0686 & +... \\ \hline
        9. prior arrest $\ge$ 2 & 0.6762 & +... \\ \hline
        10. prior arrest $\ge$ 3 & 0.3489 & +... \\ \hline
        11. prior arrest $\ge$ 4 & 0.2339 & +... \\ \hline
        12. prior arrest $\ge$ 5 & 0.1226 & +... \\ \hline
        13. prior charges $\ge$ 2 & 0.0124 & +... \\ \hline
        14. prior charges $\ge$ 2 3 & 0.0065 & +... \\ \hline
        15. prior violence $\ge$ 1 & 0.0474 & +... \\ \hline
        16. prior felony $\ge$ 1 & 0.1721 & +... \\ \hline
        17. prior misdemeanor $\ge$ 2 & 0.0162 & +... \\ \hline
        18. prior misdemeanor $\ge$ 3 & 0.0764 & +... \\ \hline
        19. prior misdemeanor $\ge$ 4 & 0.0733 & +... \\ \hline
        20. prior traffic $\ge$ 1 & 0.0394 & +... \\ \hline
        21. ADE $\ge$ 1 & 0.1583 & -... \\ \hline
        22. prior fta two year $\ge$ 1 & 0.3398 & +... \\ \hline
        23. prior fta two year $\ge$ 2 & 0.0617 & +... \\ \hline
        24. prior pending charge $\ge$ 1 & 0.3874 & +... \\ \hline
        25. prior probation $\ge$ 1 & 0.2265 & +... \\ \hline
        26. prior incarceration $\ge$ 1 & 0.3577 & +... \\ \hline
        27. six month $\ge$ 1 & 0.0148 & -... \\ \hline
        28. three year $\ge$ 1 & 0.0005 & +... \\ \hline
        29. Intercept & -1.1500 & +... \\ \hline
        \textbf{ADD POINTS FROM ROWS 1 TO 29}  &      \textbf{SCORE} & = ..... \\ \hline
        \multicolumn{3}{l}{Probability: Pr(Y = 1) = exp(score) / (1 + exp(score))} \\ \hline
    \end{tabular}
\end{center}
\end{table}

% sensitive attributes
\begin{table}[H]
\begin{center}
\scriptsize
\tabcolsep=0.1cm
\renewcommand{\arraystretch}{1.2}
\caption{Race and gender distributions for Kentucky. Due to the low percentage of the Asians and Indians in Kentucky, we included them in the "Other" category in the fairness analysis.}
\label{table:FL_KY_sensitive_attrs}

\begin{tabular}{|c|c|c|c|}\hline 
    \multicolumn{4}{c}{\textbf{Kentucky}} \\ \hline
    Attribute & Attribute Value & num\_inds &  \% total \\ \hline
    race & African-American &  42197 & 16.83 \\ \hline
    race & Asian            &    843 &  0.34 \\ \hline
    race & Caucasian        &  202341& 80.69 \\ \hline
    race & Indian           &    195 &  0.08 \\ \hline
    race & Other            &   5202 &  2.07 \\ \hline
    sex  & female           &  79207 & 31.58 \\ \hline
    sex  & male             & 171571 & 68.42 \\ \hline
\end{tabular}
\end{center}
\end{table} 

% PSA NCA table
\definecolor{Gray}{gray}{0.9}
\begin{table}[H]
\begin{center}
\scriptsize
\tabcolsep=0.1cm
\caption{Arnold Public Safety Assessment (PSA): New Criminal Activity (NCA)}

\label{table:arnold_psa_nca}
\begin{tabular}{|c|c|c|} \hline
    \multicolumn{3}{g}{New Criminal Activity (NCA)} \\
     
     Risk Factor & Value & Points\\ \hline\hline
    
     Age at Current Arrest & 23 or older & 0 \\
                            & 22 or younger & 2 \\
    Pending Charge at Time of Offense & No & 0 \\
                                    & Yes & 3 \\ 
                                    
    Prior Misdemeanor Conviction & No & 0 \\
                                    & Yes & 1 \\                                            
    Prior Felony Conviction & No & 0 \\
                            & Yes & 1 \\

    Prior Violent Conviction  & 0 & 0 \\
                            & 1 & 1 \\
                            & 2 & 1 \\
                            & 3 or more & 2                   \\
    Prior FTA in Past 2 Years & 0 & 0 \\
                            & 1 & 1\\
                            & 2 or more & 2 \\
    Prior Sentence to Incarceration & No & 0 \\
                                    & Yes & 2 \\ \hline
\end{tabular}

\vspace*{2em}
\begin{tabular}{|c|c|} \hline
    \multicolumn{2}{g}{Point Scaling} \\
     
     Total NCA Points & NCA Scaled Score \\ \hline\hline
    
     0 & 1 \\ 
     1 & 2  \\ 
     2 & 2   \\ 
     3 & 3 \\ 
     4 & 3 \\ 
     5 & 4 \\ 
     6 & 4 \\ 
     7 &  5\\ 
     8 & 5 \\ 
     9 & 6 \\ 
     10 & 6 \\
     11 & 6 \\
     12 & 6 \\
     13 & 6 \\ \hline

\end{tabular}
\end{center}
\end{table}

% PSA NVCA table
\definecolor{Gray}{gray}{0.9}

\begin{table}[H]
\begin{center}
\scriptsize
\tabcolsep=0.1cm
\caption{Arnold Public Safety Assessment (PSA): New Violent Criminal Activity (NVCA)}

\label{table:arnold_psa_nvca}
\begin{tabular}{|c|c|c|} \hline 
    \multicolumn{3}{g}{New Violent Criminal Activity (NVCA)} \\
     
     Risk Factor & Value &Points\\ \hline\hline
    
    Current Violent Offense & No & 0 \\
                            & Yes & 2 \\ 
                            
    Current Violent Offense and 20 Years or Younger & No & 0 \\
    & Yes & 1 \\ 
    Pending Charge at Time of Offense & No & 0 \\
                                    & Yes & 1 \\ 
                                    
    Prior Conviction (Misdemeanor or Felony) & No & 0 \\
                                    & Yes & 1 \\ \                                            

    Prior Violent Conviction  & 0 & 0 \\
                            & 1 & 1 \\
                            & 2 & 1 \\
                            & 3 or more & 2                   \\\hline
\end{tabular}

\vspace*{2em}
\begin{tabular}{|c|c|} \hline 
    \multicolumn{2}{g}{Point Scaling} \\
     
     Total NVCA Points & NVCA Scaled Score \\ \hline\hline
    
     0 & No \\ 
     1 & No  \\ 
     2 & No   \\ 
     3 & No \\ 
     4 & Yes \\ 
     5 & Yes \\ 
     6 & Yes \\ 
     7 & Yes \\ \hline

\end{tabular}
\end{center}
\end{table}

% Broward baseline prediction results
\begin{table*}[ht!]
\begin{center}
\scriptsize
\tabcolsep=0.25cm
\renewcommand{\arraystretch}{1.2}
\caption{Broward baseline models. Results are the average value of test AUCs from five-fold nested cross validation, with standard deviation listed in parentheses.}
\label{table:broward_baseline_pred_res}
\begin{tabular}{|c|c|c|c|c|c|c|} \hline
    \multicolumn{7}{c}{\textbf{Baseline Models}} \\ \hline
    Labels& Logistic ($\ell_2$) & Logistic($\ell_1$) & Linear SVM & RF & XGBoost &  \textbf{Performance Range} \\ \hline
    \multicolumn{7}{c}{\textbf{Two Year}} \\ \hline
    General & 0.670 (0.021) & 0.650 (0.021) & 0.670 (0.020) & 0.658 (0.027) & 0.655 (0.022) & 0.020 \\ \hline
    Violent & 0.675 (0.037) & 0.663 (0.039) & 0.659 (0.032) & 0.671 (0.036) & 0.676 (0.048) & 0.017 \\ \hline
    Drug & 0.711 (0.048) & 0.733 (0.035) & 0.695 (0.037) & 0.703 (0.040) & 0.722 (0.039) & 0.038 \\ \hline
    Property & 0.717 (0.052) & 0.730 (0.057) & 0.683 (0.048) & 0.712 (0.027) & 0.733 (0.034) & 0.051 \\ \hline
    Felony & 0.646 (0.041) & 0.648 (0.050) & 0.621 (0.036) & 0.647 (0.046) & 0.644 (0.037) & 0.027 \\ \hline
    Misdemeanor & 0.630 (0.019) & 0.597 (0.013) & 0.628 (0.018) & 0.629 (0.027) & 0.627 (0.024) & 0.033 \\ \hline
    
    \multicolumn{7}{c}{\textbf{Six Month}} \\ \hline
    General & 0.625 (0.022) & 0.608 (0.022) & 0.618 (0.028) & 0.615 (0.026) & 0.623 (0.014) & 0.017 \\ \hline
    Violent & 0.685 (0.024) & 0.651 (0.038) & 0.619 (0.036) & 0.668 (0.045) & 0.685 (0.033) & 0.066 \\ \hline
    Drug & 0.673 (0.084) & 0.696 (0.022) & 0.640 (0.081) & 0.675 (0.055) & 0.698 (0.038) & 0.058 \\ \hline
    Property & 0.727 (0.047) & 0.725 (0.053) & 0.659 (0.069) & 0.687 (0.047) & 0.725 (0.048) & 0.068 \\ \hline
    Felony & 0.611 (0.050) & 0.613 (0.054) & 0.580 (0.086) & 0.591 (0.061) & 0.585 (0.066) & 0.034 \\ \hline
    Misdemeanor & 0.612 (0.038) & 0.586 (0.040) & 0.586 (0.016) & 0.593 (0.039) & 0.608 (0.031) & 0.027 \\ \hline
\end{tabular}
\end{center}
\end{table*} 

% Kentucky baseline prediction results
\begin{table*}[ht!]
\begin{center}
\scriptsize
\tabcolsep=0.25cm
\renewcommand{\arraystretch}{1.2}
\caption{Kentucky baseline models. Results are the average value of test AUCs from five-fold nested cross validation, with standard deviation listed in parentheses.}
\label{table:kentucky_baseline_pred_res}
\begin{tabular}{|c|c|c|c|c|c|c|} \hline
    \multicolumn{7}{c}{\textbf{Baseline Models}} \\ \hline
    Labels & Logistic ($\ell_2$) & Logistic($\ell_1$) & Linear SVM  & RF & XGBoost & \textbf{Performance Range} \\ \hline
    
    \multicolumn{7}{c}{\textbf{Two Year}} \\ \hline
    General     & 0.745 (0.004) & 0.745 (0.004) & 0.746 (0.004) & 0.753 (0.003) & 0.759 (0.003) & 0.014 \\ \hline
    Violent     & 0.768 (0.002) & 0.769 (0.003) & 0.769 (0.003) & 0.777 (0.005) & 0.784 (0.004) & 0.016 \\ \hline
    Drug        & 0.730 (0.003) & 0.730 (0.003) & 0.733 (0.003) & 0.743 (0.002) & 0.749 (0.002) & 0.019 \\ \hline
    Property    & 0.785 (0.005) & 0.785 (0.005) & 0.787 (0.005) & 0.801 (0.004) & 0.806 (0.004) & 0.021 \\ \hline
    Felony      & 0.765 (0.001) & 0.765 (0.001) & 0.768 (0.002) & 0.779 (0.002) & 0.784 (0.001) & 0.019 \\ \hline
    Misdemeanor & 0.729 (0.005) & 0.729 (0.005) & 0.730 (0.006) & 0.738 (0.005) & 0.744 (0.005) & 0.016 \\ \hline
    
    \multicolumn{7}{c}{\textbf{Six Month}} \\ \hline
    General     & 0.761 (0.004) & 0.761 (0.004) & 0.764 (0.005) & 0.779 (0.003) & 0.785 (0.004) & 0.024 \\ \hline
    Violent     & 0.833 (0.007) & 0.834 (0.006) & 0.833 (0.007) & 0.843 (0.006) & 0.847 (0.005) & 0.014 \\ \hline
    Drug        & 0.782 (0.003) & 0.782 (0.003) & 0.785 (0.003) & 0.803 (0.003) & 0.811 (0.002) & 0.029 \\ \hline
    Property    & 0.834 (0.012) & 0.834 (0.013) & 0.831 (0.014) & 0.857 (0.011) & 0.860 (0.011) & 0.029 \\ \hline
    Felony      & 0.799 (0.002) & 0.800 (0.002) & 0.804 (0.003) & 0.824 (0.003) & 0.831 (0.002) & 0.032 \\ \hline
    Misdemeanor & 0.746 (0.007) & 0.746 (0.007) & 0.748 (0.007) & 0.765 (0.006) & 0.774 (0.006) & 0.028 \\ \hline
\end{tabular}
\end{center}
\end{table*}

% Broward interpretable prediction results
\begin{table*}[ht!]
\begin{center}
\scriptsize
\tabcolsep=0.25cm
\renewcommand{\arraystretch}{1.2}
\caption{AUCs of intepretable models on Broward data. For the violence problem, we use  the Arnold New Violent Criminal Activity score. For the general problem, we use the Arnold New Criminal Activity score. }
\label{table:broward_interpretable_pred_res}
\begin{tabular}{|c|c|c|c|c|c|c|c|} \hline
    & \multicolumn{4}{c}{\textbf{Interpretable Models}} & &  \multicolumn{2}{c}{\textbf{Existing Risk Models}} \\ \hline
    Labels & CART & EBM & Additive Stumps & RiskSLIM &  \textbf{Performance Range} & Arnold PSA & COMPAS \\ \hline
    
    \multicolumn{8}{c}{\textbf{Two Year}} \\ \hline
    General & 0.613 (0.025) & 0.664 (0.027) & 0.651 (0.020) & 0.624 (0.022) & 0.051 & 0.605 (0.022) & 0.631 (0.019) \\ \hline
    Violent & 0.613 (0.045) & 0.673 (0.045) & 0.665 (0.034) & 0.655 (0.055) & 0.059 & 0.649 (0.028) & - \\ \hline
    Drug & 0.666 (0.026) & 0.685 (0.043) & 0.716 (0.037) & 0.697 (0.027) & 0.049 & - & - \\ \hline
    Property    & 0.686 (0.059) & 0.736 (0.034) & 0.736 (0.033) & 0.717 (0.020) & 0.052 & - & - \\ \hline
    Felony & 0.596 (0.033) & 0.655 (0.050) & 0.631 (0.028) & 0.590 (0.036) & 0.065 & - & - \\ \hline
    Misdemeanor & 0.577 (0.036) & 0.636 (0.029) & 0.609 (0.020) & 0.579 (0.015) & 0.059 & - & - \\ \hline

    \multicolumn{8}{c}{\textbf{Six Month}} \\ \hline
    General & 0.549 (0.021) & 0.622 (0.022) & 0.620 (0.019) & 0.585 (0.021) & 0.074 & 0.577 (0.018) & 0.609 (0.019) \\ \hline
    Violent & 0.631 (0.050) & 0.680 (0.040) & 0.676 (0.029) & 0.671 (0.039) & 0.049 & 0.675 (0.038) & - \\ \hline
    Drug & 0.569 (0.074) & 0.672 (0.043) & 0.656 (0.068) & 0.650 (0.068) & 0.102 & - & - \\ \hline
    Property & 0.637 (0.052) & 0.725 (0.031) & 0.725 (0.036) & 0.703 (0.023) & 0.089 & - & - \\ \hline
    Felony & 0.513 (0.014) & 0.606 (0.049) & 0.574 (0.036) & 0.561 (0.045) & 0.093 & - & - \\ \hline
    Misdemeanor & 0.535 (0.021) & 0.608 (0.042) & 0.582 (0.036) & 0.576 (0.024) & 0.073 & - & - \\ \hline
\end{tabular}
\end{center}
\end{table*}

% Kentucky interpretable predcition results
\begin{table*}[ht!]
\begin{center}
\scriptsize
\tabcolsep=0.25cm
\renewcommand{\arraystretch}{1.2}
\caption{AUCs of interpretable models on Kentucky data. For the violence problem, we use the Arnold New Violent Criminal Activity score. For the general problem, we use the Arnold New Criminal Activity score. }
\label{table:kentucky_interpretable_pred_res}

\begin{tabular}{|c|c|c|c|c|c|c|} \hline
    & \multicolumn{4}{c}{\textbf{Interpretable Models}} & & \textbf{Existing Risk Models} \\ \hline
    Labels & CART & EBM & Additive Stumps & RiskSLIM  & \textbf{Performance Range} & Arnold PSA \\ \hline
    
    \multicolumn{7}{c}{\textbf{Two Year}} \\ \hline
    General     & 0.746 (0.003) & 0.751 (0.004) & 0.748 (0.004) & 0.708 (0.003) & 0.042 & 0.711 (0.004) \\ \hline
    Violent     & 0.763 (0.007) & 0.777 (0.004) & 0.770 (0.005) & 0.744 (0.008) & 0.032 & 0.743 (0.003) \\ \hline
    Drug        & 0.736 (0.002) & 0.740 (0.001) & 0.738 (0.002) & 0.708 (0.005) & 0.032 & - \\ \hline
    Property    & 0.790 (0.003) & 0.798 (0.006) & 0.796 (0.005) & 0.761 (0.003) & 0.037 & - \\ \hline
    Felony      & 0.771 (0.002) & 0.776 (0.001) & 0.773 (0.002) & 0.757 (0.007) & 0.019 & - \\ \hline
    Misdemeanor & 0.730 (0.005) & 0.735 (0.005) & 0.729 (0.006) & 0.701 (0.002) & 0.033 & - \\ \hline
    
    \multicolumn{7}{c}{\textbf{Six Month}} \\ \hline
    General     & 0.772 (0.005) & 0.773 (0.004) & 0.771 (0.004) & 0.737 (0.002) & 0.037 & 0.718 (0.004) \\ \hline
    Violent     & 0.822 (0.011) & 0.843 (0.006) & 0.836 (0.004) & 0.810 (0.009) & 0.033 & 0.794 (0.011) \\ \hline
    Drug        & 0.794 (0.003) & 0.793 (0.004) & 0.796 (0.004) & 0.763 (0.004) & 0.033 & - \\ \hline
    Property    & 0.839 (0.014) & 0.850 (0.012) & 0.851 (0.010) & 0.832 (0.010) & 0.019 & - \\ \hline
    Felony      & 0.811 (0.003) & 0.820 (0.003) & 0.813 (0.003) & 0.790 (0.006) & 0.030 & - \\ \hline
    Misdemeanor & 0.760 (0.006) & 0.757 (0.006) & 0.751 (0.006) & 0.705 (0.005) & 0.055 & - \\ \hline
\end{tabular}
\end{center}
\end{table*}

% KY Model -- trained on KY; tested on entire FL
\begin{table*}[ht!]
\begin{center}
\scriptsize
\tabcolsep=0.2cm
\renewcommand{\arraystretch}{1.2}
\caption{Training baseline models and interpretable models on the Kentucky data set using five-fold nested cross validation and testing the best-performing model on the Broward data set.}
\label{table:train_kentucky_test_broward}
\begin{tabular}{|c|c|c|c|c|c|c|c|c|c|} \hline
    & \multicolumn{4}{c}{\textbf{Baseline Models}} & & \multicolumn{3}{c}{\textbf{Interpretable Models}} &\\ \hline
    Labels & Logistic ($\ell_2$) & Logistic ($\ell_1$) & Linear SVM & Random Forest & XGBoost & CART & EBM & Additive Stumps & RiskSLIM\\ \hline
    
    \multicolumn{10}{c}{\textbf{Two Year}} \\ \hline
    General & 0.615(0.001) & 0.614(0.001) & 0.610(0.000) & 0.619(0.001) & 0.617(0.003) & 0.595(0.009) & 0.612(0.002) & 0.608(0.001) & 0.568(0.000) \\ \hline
    Violent & 0.655(0.001) & 0.653(0.002) & 0.630(0.000) & 0.652(0.000) & 0.652(0.004) & 0.622(0.030) & 0.640(0.010) & 0.652(0.002) & 0.629(0.018) \\ \hline
    Drug & 0.629(0.001) & 0.629(0.001) & 0.618(0.000) & 0.614(0.002) & 0.637(0.002) & 0.621(0.010) & 0.629(0.003) & 0.631(0.001) & 0.625(0.000) \\ \hline
    Property & 0.664(0.001) & 0.672(0.001) & 0.649(0.000) & 0.668(0.002) & 0.674(0.008) & 0.649(0.017) & 0.665(0.011) & 0.659(0.001) & 0.639(0.021) \\ \hline
    Felony      & 0.630(0.001) & 0.630(0.001) & 0.624(0.000) & 0.631(0.001) & 0.627(0.005) & 0.611(0.003) & 0.623(0.005) & 0.624(0.000) & 0.614(0.000) \\ \hline
    Misdemeanor & 0.558(0.000) & 0.558(0.000) & 0.551(0.000) & 0.561(0.001) & 0.576(0.002) & 0.555(0.004) & 0.571(0.003) & 0.557(0.000) & 0.539(0.002) \\ \hline
    
    \multicolumn{10}{c}{\textbf{Six Month}} \\ \hline
    General & 0.577(0.002) & 0.576(0.001) & 0.569(0.000) & 0.577(0.001) & 0.581(0.002) & 0.562(0.007) & 0.571(0.004) & 0.562(0.001) & 0.553(0.000) \\ \hline
    Violent & 0.641(0.002) & 0.644(0.001) & 0.614(0.000) & 0.643(0.001) & 0.626(0.004) & 0.611(0.013) & 0.622(0.009) & 0.650(0.001) & 0.637(0.002) \\ \hline
    Drug & 0.607(0.004) & 0.604(0.003) & 0.589(0.000) & 0.567(0.005) & 0.593(0.007) & 0.580(0.018) & 0.618(0.006) & 0.576(0.001) & 0.566(0.020) \\ \hline
    Property & 0.662(0.001) & 0.665(0.002) & 0.635(0.000) & 0.652(0.002) & 0.656(0.013) & 0.634(0.016) & 0.657(0.008) & 0.640(0.004) & 0.619(0.000) \\ \hline
    Felony  & 0.586(0.001) & 0.584(0.002) & 0.575(0.000) & 0.589(0.002) & 0.580(0.002) & 0.563(0.003) & 0.571(0.005) & 0.574(0.001) & 0.550(0.001) \\ \hline
    Misdemeanor & 0.558(0.002) & 0.558(0.000) & 0.550(0.000) & 0.552(0.002) & 0.563(0.004) & 0.554(0.012) & 0.559(0.002) & 0.542(0.001) & 0.526(0.003) \\ \hline
\end{tabular}
\end{center}
\end{table*}

% KY Model -- trained on FL; tested on FL holdout test
\begin{table*}[ht!]
\begin{center}
\scriptsize
\tabcolsep=0.2cm
\renewcommand{\arraystretch}{1.2}
\caption{Training baseline models and interpretable models on the Broward County data set using five-fold nested cross validation and testing the resulting best-performing model on a held out portion of the Broward data set.}
\label{table:train_broward_test_broward}
\begin{tabular}{|c|c|c|c|c|c|c|c|c|c|} \hline
    & \multicolumn{4}{c}{\textbf{Baseline Models}} & & \multicolumn{3}{c}{\textbf{Interpretable Models}} &\\ \hline
    Labels & Logistic ($\ell_2$) & Logistic ($\ell_1$) & Linear SVM & Random Forest & XGBoost & CART & EBM & Additive Stumps & RiskSLIM\\ \hline

    \multicolumn{10}{c}{\textbf{Two Year}} \\ \hline
    General & 0.669(0.020) & 0.649(0.021) & 0.670(0.020) & 0.657(0.034) & 0.659(0.019) & 0.629(0.028) & 0.663(0.031) & 0.644(0.027) & 0.622(0.021) \\ \hline
    Violent & 0.679(0.038) & 0.662(0.035) & 0.662(0.034) & 0.675(0.037) & 0.677(0.050) & 0.600(0.037) & 0.675(0.049) & 0.673(0.035) & 0.670(0.032) \\ \hline
    Drug & 0.716(0.047) & 0.734(0.034) & 0.702(0.043) & 0.688(0.044) & 0.720(0.034) & 0.672(0.041) & 0.690(0.054) & 0.709(0.044) & 0.706(0.027) \\ \hline
    Property & 0.721(0.057) & 0.731(0.057) & 0.687(0.052) & 0.725(0.039) & 0.729(0.040) & 0.685(0.058) & 0.738(0.031) & 0.733(0.039) & 0.703(0.036) \\ \hline
    Felony  & 0.651(0.040) & 0.652(0.053) & 0.622(0.036) & 0.649(0.045) & 0.647(0.039) & 0.598(0.034) & 0.656(0.050) & 0.640(0.031) & 0.603(0.042) \\ \hline
    Misdemeanor & 0.634(0.017) & 0.602(0.012) & 0.632(0.017) & 0.629(0.022) & 0.624(0.020) & 0.585(0.041) & 0.633(0.025) & 0.603(0.016) & 0.558(0.026) \\ \hline
    
    \multicolumn{10}{c}{\textbf{Six Month}} \\ \hline
    General & 0.624(0.024) & 0.607(0.019) & 0.619(0.026) & 0.620(0.025) & 0.621(0.019) & 0.553(0.014) & 0.620(0.027) & 0.617(0.035) & 0.600(0.021) \\ \hline
    Violent & 0.680(0.027) & 0.650(0.038) & 0.614(0.039) & 0.670(0.039) & 0.689(0.031) & 0.623(0.043) & 0.683(0.040) & 0.683(0.032) & 0.691(0.032) \\ \hline
    Drug & 0.672(0.082) & 0.696(0.025) & 0.649(0.080) & 0.687(0.065) & 0.686(0.044) & 0.569(0.074) & 0.655(0.035) & 0.704(0.054) & 0.719(0.039) \\ \hline
    Property & 0.726(0.049) & 0.725(0.053) & 0.648(0.058) & 0.698(0.046) & 0.720(0.052) & 0.637(0.052) & 0.723(0.030) & 0.699(0.038) & 0.663(0.048) \\ \hline
    Felony  & 0.620(0.058) & 0.613(0.054) & 0.587(0.086) & 0.611(0.076) & 0.601(0.047) & 0.524(0.015) & 0.605(0.052) & 0.584(0.034) & 0.557(0.043) \\ \hline
    Misdemeanor & 0.616(0.030) & 0.583(0.039) & 0.590(0.022) & 0.601(0.049) & 0.620(0.044) & 0.543(0.033) & 0.612(0.050) & 0.576(0.037) & 0.556(0.040) \\ \hline
\end{tabular}
\end{center}
\end{table*}

% FL Model -- trained on FL; tested on entire KY
\begin{table*}[ht!]
\begin{center}
\scriptsize
\tabcolsep=0.2cm
\renewcommand{\arraystretch}{1.4}
\caption{Training baseline and interpretable models on the Broward County data set using five-fold nested cross validation and testing the resulting best-performing model on the Kentucky data set.}
\label{table:train_broward_test_kentucky}
\begin{tabular}{|c|c|c|c|c|c|c|c|c|c|} \hline
    & \multicolumn{4}{c}{\textbf{Baseline Models}} & & \multicolumn{3}{c}{\textbf{Interpretable Models}} &\\ \hline
    Labels & Logistic ($\ell_2$) & Logistic ($\ell_1$) & Linear SVM & Random Forest & XGBoost & CART & EBM & Additive Stumps & RiskSLIM\\ \hline
    
    \multicolumn{10}{c}{\textbf{Two Year}} \\ \hline
    General     & 0.664(0.007) & 0.653(0.001) & 0.658(0.007) & 0.701(0.005) & 0.689(0.006) & 0.626(0.025) & 0.704(0.003) & 0.653(0.009) & 0.649(0.037) \\ \hline
    Violent     & 0.674(0.005) & 0.650(0.007) & 0.611(0.013) & 0.729(0.005) & 0.724(0.005) & 0.589(0.053) & 0.720(0.005) & 0.657(0.018) & 0.663(0.025) \\ \hline
    Drug        & 0.649(0.008) & 0.632(0.003) & 0.554(0.005) & 0.655(0.022) & 0.650(0.006) & 0.613(0.013) & 0.656(0.008) & 0.626(0.009) & 0.634(0.012) \\ \hline
    Property    & 0.628(0.022) & 0.663(0.014) & 0.556(0.017) & 0.695(0.018) & 0.669(0.023) & 0.548(0.018) & 0.687(0.011) & 0.590(0.014) & 0.593(0.052) \\ \hline
    Felony      & 0.671(0.006) & 0.661(0.002) & 0.592(0.014) & 0.724(0.003) & 0.706(0.014) & 0.592(0.042) & 0.725(0.006) & 0.676(0.023) & 0.631(0.059) \\ \hline
    Misdemeanor & 0.638(0.007) & 0.619(0.026) & 0.579(0.010)  & 0.665(0.011) & 0.645(0.014) & 0.574(0.053) & 0.669(0.007) & 0.621(0.017) & 0.631(0.025) \\ \hline
    
    \multicolumn{10}{c}{\textbf{Six Month}} \\ \hline
    General     & 0.676(0.006) & 0.665(0.004) & 0.601(0.011) & 0.698(0.009) & 0.685(0.010) & 0.613(0.018) & 0.709(0.005) & 0.663(0.012) & 0.602(0.046) \\ \hline
    Violent     & 0.653(0.015) & 0.662(0.021) & 0.533(0.011) & 0.762(0.047) & 0.773(0.007) & 0.625(0.059) & 0.757(0.004) & 0.728(0.026) & 0.723(0.004) \\ \hline
    Drug        & 0.663(0.031) & 0.678(0.008) & 0.521(0.006) & 0.682(0.009) & 0.658(0.027) & 0.600(0.082) & 0.609(0.037) & 0.619(0.025) & 0.635(0.017) \\ \hline
    Property    & 0.681(0.012) & 0.708(0.009) & 0.529(0.012) & 0.719(0.053) & 0.718(0.010) & 0.555(0.007) & 0.715(0.018) & 0.643(0.022) & 0.696(0.053) \\ \hline
    Felony      & 0.685(0.008) & 0.679(0.008) & 0.556(0.011) & 0.719(0.018) & 0.683(0.025) & 0.552(0.049) & 0.724(0.010) & 0.652(0.039) & 0.621(0.036) \\ \hline
    Misdemeanor & 0.664(0.003) & 0.658(0.008) & 0.558(0.016) & 0.670(0.004) & 0.662(0.006) & 0.604(0.019) & 0.676(0.006) & 0.615(0.019) & 0.583(0.070) \\ \hline
\end{tabular}
\end{center}
\vspace{8mm}
\end{table*}

% FL Model -- trained on FL; tested on KY holdout test
\begin{table*}[ht!]
\begin{center}
\scriptsize
\tabcolsep=0.2cm
\renewcommand{\arraystretch}{1.4}
\caption{Training baseline models and interpretable models on the Kentucky data set using five-fold nested cross validation and testing the resulting best-performing model on a held out portion of the Kentucky data set.}
\label{table:train_kentucky_test_kentucky}
\begin{tabular}{|c|c|c|c|c|c|c|c|c|c|} \hline
    & \multicolumn{4}{c}{\textbf{Baseline Models}} & & \multicolumn{3}{c}{\textbf{Interpretable Models}} &\\ \hline
    Labels & Logistic ($\ell_2$) & Logistic ($\ell_1$) & Linear SVM & Random Forest & XGBoost & CART & EBM & Additive Stumps & RiskSLIM\\ \hline
    
    \multicolumn{10}{c}{\textbf{Two Year}} \\ \hline
    General     & 0.739(0.003) & 0.739(0.003) & 0.740(0.004) & 0.752(0.004) & 0.757(0.003) & 0.746(0.003) & 0.750(0.004) & 0.747(0.004) & 0.704(0.004) \\ \hline
    Violent     & 0.765(0.001) & 0.766(0.002) & 0.767(0.002) & 0.776(0.004) & 0.783(0.004) & 0.763(0.007) & 0.776(0.004) & 0.771(0.005) & 0.741(0.010) \\ \hline
    Drug        & 0.723(0.002) & 0.723(0.002) & 0.727(0.002) & 0.739(0.002) & 0.745(0.002) & 0.733(0.002) & 0.737(0.002) & 0.734(0.003) & 0.708(0.002) \\ \hline
    Property    & 0.780(0.004) & 0.779(0.004) & 0.784(0.004) & 0.801(0.004) & 0.805(0.004) & 0.790(0.004) & 0.797(0.005) & 0.796(0.005) & 0.764(0.009) \\ \hline
    Felony  & 0.758(0.002) & 0.758(0.002) & 0.763(0.002) & 0.778(0.002) & 0.783(0.001) & 0.771(0.002) & 0.775(0.001) & 0.773(0.001) & 0.765(0.001) \\ \hline
    Misdemeanor & 0.722(0.005) & 0.722(0.005) & 0.724(0.006) & 0.736(0.006) & 0.742(0.005) & 0.729(0.005) & 0.733(0.006) & 0.729(0.006) & 0.693(0.010) \\ \hline
    
    \multicolumn{10}{c}{\textbf{Six Month}} \\ \hline
    General     & 0.752(0.004) & 0.752(0.004) & 0.757(0.004) & 0.775(0.003) & 0.780(0.003) & 0.769(0.005) & 0.770(0.004) & 0.768(0.004) & 0.736(0.004) \\ \hline
    Violent     & 0.828(0.006) & 0.830(0.005) & 0.834(0.005) & 0.843(0.005) & 0.846(0.005) & 0.821(0.011) & 0.842(0.005) & 0.837(0.004) & 0.809(0.005) \\ \hline
    Drug        & 0.770(0.003) & 0.771(0.003) & 0.777(0.004) & 0.794(0.004) & 0.799(0.002) & 0.783(0.005) & 0.785(0.004) & 0.786(0.004) & 0.752(0.006) \\ \hline
    Property    & 0.830(0.010) & 0.829(0.011) & 0.830(0.013) & 0.856(0.009) & 0.860(0.011) & 0.839(0.014) & 0.849(0.011) & 0.851(0.010) & 0.835(0.009) \\ \hline
    Felony      & 0.790(0.002) & 0.791(0.002) & 0.798(0.003) & 0.823(0.003) & 0.829(0.003) & 0.811(0.005) & 0.818(0.004) & 0.812(0.004) & 0.790(0.005) \\ \hline
    Misdemeanor & 0.735(0.006) & 0.735(0.006) & 0.740(0.007) & 0.760(0.005) & 0.766(0.005) & 0.754(0.005) & 0.753(0.006) & 0.750(0.006) & 0.705(0.005) \\ \hline
\end{tabular}
\end{center}
\vspace{8mm}
\end{table*}

% BG-AUC Violent
\begin{table*}[ht!]
\begin{center}
\scriptsize
\tabcolsep=0.1cm
\renewcommand{\arraystretch}{1.4}
\caption{AUCs of the Arnold NVCA Raw, EBM and RiskSLIM on Kentucky for two-year \texttt{violent} recidivism, conditioned on sensitive attributes. AUC ranges are also given for each sensitive attribute class}
\begin{tabular}{|c|c|c|c|c|c|c|c|c|}\hline
    %% KENTUCKY
    \multicolumn{9}{c}{\textbf{Kentucky}} \\ \hline
    & & \multicolumn{3}{c}{\textbf{Race}} & & \multicolumn{3}{c}{\textbf{Sex}} \\ \hline
    \verb|     Model     | & \verb|      Label       | & \verb|Afr-Am.| & \verb|Cauc.| & \verb|Other Race| & \verb|race_range| & \verb|Female| & \verb|Male | & \verb|sex_range| \\ \hline
    Arnold NVCA Raw & violent\_two\_year &   0.728 & 0.740 &      0.767 &      0.039 &  0.728 & 0.734 &     0.006 \\ \hline
    EBM             & violent\_two\_year &   0.775 & 0.770 &      0.766 &      0.009 &  0.744 & 0.766 &     0.022 \\ \hline
    RiskSLIM        & violent\_two\_year &   0.744 & 0.736 &      0.680 &      0.063 &  0.706 & 0.730 &     0.024 \\ \hline
\end{tabular}
\end{center}
\end{table*}

% features
%\input{tables/FL_features.tex}
%\input{tables/KY_features.tex}

%%%%%%%%%%%%%%%%%%%%%%%%%

% 2-year Kentucky
\begin{table}[ht!]
\begin{center}
\scriptsize
\tabcolsep=0.1cm
\renewcommand{\arraystretch}{1.2}
\caption{Two Year Prediction Problems---Kentucky. Here, counts of prior arrests indicate the counts of arrests with at least one convicted charge. All charges mentioned are convicted charges. ADE indicates assignment to alcohol and drug education classes.}
% general
\begin{tabular}{|p{4cm}|r|r|} \hline
\multicolumn{3}{c}{\textbf{Two Year General Recidivism}} \\ \hline
\multicolumn{3}{l}{Pr(Y = +1) = 1 / (1 + exp(-(-2 + score)))} \\ \hline
number of prior arrests$\geq$2 & 1 points & +... \\ \hline
number of prior arrests$\geq$3 & 1 points & +... \\ \hline
number of prior arrests$\geq$5 & 1 points & +... \\ \hline
\textbf{ADD POINTS FROM ROWS 1 TO 3}  &  	\textbf{SCORE} & = .....\\  \hline
\end{tabular}
% violent

\vspace*{2em}
\begin{tabular}{|p{4cm}|r|r|} \hline
\multicolumn{3}{c}{\textbf{Two Year Violent Recidivism}} \\ \hline
\multicolumn{3}{l}{Pr(Y = +1) = 1 / (1 + exp(-(-6 + score)))} \\ \hline
sex = Male & 1 points & +... \\ \hline
age at current charge $\leq$ 27 & 1 points & +... \\ \hline 
number of prior arrests$\geq$2 & 1 points & +... \\ \hline
number of prior violent charges$\geq$1 & 1 points & +... \\ \hline
sentenced to incarceration before = Yes & 1 points & +... \\ \hline
\textbf{ADD POINTS FROM ROWS 1 TO 5}  &  	\textbf{SCORE} & = .....\\  \hline
\end{tabular}

\vspace*{2em}

%drug
\begin{tabular}{|p{4cm}|r|r|} \hline
\multicolumn{3}{c}{\textbf{Two Year Drug Recidivism}} \\ \hline
\multicolumn{3}{l}{Pr(Y = +1) = 1 / (1 + exp(-(-4 + score)))} \\ \hline
number of prior arrests$\geq$2 & 1 points & +... \\ \hline
number of prior drug related \par charges$\geq$1 & 1 points & +... \\ \hline
number of times charged with \par a new offense when there is a pending case$\geq$1 & 1 points & +... \\ \hline
\textbf{ADD POINTS FROM ROWS 1 TO 3}  &  	\textbf{SCORE} & = .....\\  \hline
\end{tabular}
% property

\vspace*{2em}
\begin{tabular}{|p{4cm}|r|r|} \hline
\multicolumn{3}{c}{\textbf{Two Year Property Recidivism}} \\ \hline
\multicolumn{3}{l}{Pr(Y = +1) = 1 / (1 + exp(-(-4 + score)))} \\ \hline
number of prior property related \par charges$\geq$1 & 1 points & +... \\ \hline
number of prior arrests$\geq$3 & 1 points & +... \\ \hline
number of times charged with \par a new offense when there is a pending case$\geq$1 & 1 points & +... \\ \hline
number of prior ADE $\geq$1 & -1 points & +... \\
\textbf{ADD POINTS FROM ROWS 1 TO 4}  &  	\textbf{SCORE} & = .....\\  \hline
\end{tabular}

\vspace*{2em}

% felony
\begin{tabular}{|p{4cm}|r|r|} \hline
\multicolumn{3}{c}{\textbf{Two Year Felony Recidivism}} \\ \hline
\multicolumn{3}{l}{Pr(Y = +1) = 1 / (1 + exp(-(-5 + score)))} \\ \hline
age at current charge $\leq$ 43 & 1 points & +... \\ \hline
number of prior arrests$\geq$2 & 1 points & +... \\ \hline
number of prior felony level charges$\geq$1 & 1 points & +... \\ \hline
number of times charged with \par a new offense when there is a pending case$\geq$1 & 1 points & +... \\ \hline
sentenced to incarceration before = Yes & 1 points & +... \\ \hline
\textbf{ADD POINTS FROM ROWS 1 TO 5}  &  	\textbf{SCORE} & = .....\\  \hline
\end{tabular}
% misdemeanor

\vspace*{2em}
\begin{tabular}{|p{4cm}|r|r|} \hline
\multicolumn{3}{c}{\textbf{Two Year Misdemeanor Recidivism}} \\ \hline
\multicolumn{3}{l}{Pr(Y = +1) = 1 / (1 + exp(-(-3 + score)))} \\ \hline
number of prior arrests$\geq$2 & 1 points & +... \\ \hline
number of times charged with \par a new offense when there is a pending case$\geq$1 & 1 points & +... \\ \hline
sentenced to incarceration before = Yes & 1 points & +... \\ \hline
\textbf{ADD POINTS FROM ROWS 1 TO 3}  &  	\textbf{SCORE} & = .....\\  \hline
\end{tabular}
\end{center}
\end{table}
%%%%%%%%%%%%%%%%%%%%%%%%%%%%%%%%%%%%%%

%%%%%%%%%%%%%%%% Six Month Kentucky %%%%%%%%%%%%%%%
\begin{table}[h]
\begin{center}
\scriptsize
\tabcolsep=0.1cm
\renewcommand{\arraystretch}{1.3}
\caption{Six Month Prediction Problems---Kentucky. Here, counts of prior arrests indicate the counts of arrests with at least one convicted charge. All charges mentioned are convicted charges. ADE means assignment to alcohol and drug education classes. }
% general
\begin{tabular}{|p{4cm}|r|r|} \hline
\multicolumn{3}{c}{\textbf{Six Month General Recidivism}} \\ \hline
\multicolumn{3}{l}{Pr(Y = +1) = 1 / (1 + exp(-(-4 + score)))} \\ \hline
number of prior arrests$\geq$2 & 1 points & +... \\ \hline
number of prior arrests$\geq$4 & 1 points & +... \\ \hline
number of times charged with \par a new offense when there is a pending case$\geq$1 & 1 points & +... \\ \hline
\textbf{ADD POINTS FROM ROWS 1 TO 3}  &  	\textbf{SCORE} & = .....\\  \hline
\end{tabular}
% violent

\vspace*{2em}
\begin{tabular}{|p{4cm}|r|r|} \hline
\multicolumn{3}{c}{\textbf{Six Month Violent Recidivism}} \\ \hline
\multicolumn{3}{l}{Pr(Y = +1) = 1 / (1 + exp(-(-7 + score)))} \\ \hline
number of prior violent charges$\geq$1 & 1 points & +... \\ \hline
number of prior arrests$\geq$3 & 1 points & +... \\ \hline
number of prior felony level charges$\geq$1 & 1 points & +... \\ \hline
current violent charge = Yes & 1 points & +... \\ \hline
number of times charged with \par a new offense when there is a pending case$\geq$1 & 1 points & +... \\ \hline
\textbf{ADD POINTS FROM ROWS 1 TO 5}  &  	\textbf{SCORE} & = .....\\  \hline
\end{tabular}

\vspace*{2em}

% drug
\begin{tabular}{|p{4cm}|r|r|} \hline
\multicolumn{3}{c}{\textbf{Six Month Drug Recidivism}} \\ \hline
\multicolumn{3}{l}{Pr(Y = +1) = 1 / (1 + exp(-(-5 + score)))} \\ \hline
number of prior drug related charges$\geq$1 & 1 points & +... \\ \hline
number of prior drug related charges$\geq$3 & 1 points & +... \\ \hline
number of times charged with \par a new offense when there is a pending case$\geq$1 & 1 points & +... \\ \hline
number of prior ADE$\geq$1 & -1 points & +... \\ \hline
\textbf{ADD POINTS FROM ROWS 1 TO 4}  &  	\textbf{SCORE} & = .....\\  \hline
\end{tabular}
% property

\vspace*{2em}
\begin{tabular}{|p{4cm}|r|r|} \hline
\multicolumn{3}{c}{\textbf{Six Month Property Recidivism}} \\ \hline
\multicolumn{3}{l}{Pr(Y = +1) = 1 / (1 + exp(-(-7 + score)))} \\ \hline
number of prior property related charges$\geq$1 & 2 points & +...\\ \hline
number of prior felony level charges$\geq$1 & 1 points & +... \\ \hline
number of prior FTA within last two years $\geq$1 & 1 points & +... \\ \hline
number of times charged with \par a new offense when there is a pending case$\geq$1 & 1 points & +... \\ \hline
\textbf{ADD POINTS FROM ROWS 1 TO 4}  &  	\textbf{SCORE} & = .....\\  \hline
\end{tabular}

\vspace*{2em}

% felony
\begin{tabular}{|p{4cm}|r|r|} \hline
\multicolumn{3}{c}{\textbf{Six Month Felony Recidivism}} \\ \hline
\multicolumn{3}{l}{Pr(Y = +1) = 1 / (1 + exp(-(-5 + score)))} \\ \hline
number of prior arrests$\geq$3 & 1 points & +... \\ \hline
number of prior felony level charges$\geq$1 & 1 points & +... \\ \hline
number of times charged with \par a new offense when there is a pending case$\geq$1 & 1 points & +... \\ \hline
\textbf{ADD POINTS FROM ROWS 1 TO 3}  &  	\textbf{SCORE} & = .....\\  \hline
\end{tabular}
% misdemeanor

\vspace*{2em}
\begin{tabular}{|p{4cm}|r|r|} \hline
\multicolumn{3}{c}{\textbf{Six Month Misdemeanor Recidivism}} \\ \hline
\multicolumn{3}{l}{Pr(Y = +1) = 1 / (1 + exp(-(-4 + score)))} \\ \hline
number of prior arrests$\geq$2 & 1 points & +... \\ \hline
number of prior arrests$\geq$4 & 1 points & +... \\ \hline
\textbf{ADD POINTS FROM ROWS 1 TO 2}  &  	\textbf{SCORE} & = .....\\  \hline
\end{tabular}
\end{center}
\end{table}
\begin{table}[t]
\begin{center}
\scriptsize
\tabcolsep=0.1cm
\renewcommand{\arraystretch}{1.3}
\caption{Two Year Prediction Problems---Broward. Here, counts of prior arrests indicate the counts of arrests with at least one non-convicted or convicted charge. All charges mentioned are non-convicted charges.}
% general
\begin{tabular}{|p{4cm}|r|r|} \hline
\multicolumn{3}{c}{\textbf{Two Year General Recidivism}} \\ \hline
\multicolumn{3}{l}{Pr(Y = +1) = 1 / (1 + exp(-(-2 + score)))} \\ \hline
age at current charge $\leq$31 & 1 points & +... \\ \hline

number of prior misdemeanor level charges $\geq$4 & 1 points & +... \\ \hline
had charge(s) within last three years = Yes & 1 points & +... \\ \hline
\textbf{ADD POINTS FROM ROWS 1 TO 3}  &  	\textbf{SCORE} & = .....\\  \hline
\end{tabular}
% violent

\vspace*{2em}
\begin{tabular}{|p{4cm}|r|r|} \hline
\multicolumn{3}{c}{\textbf{Two Year Violent Recidivism}} \\ \hline
\multicolumn{3}{l}{Pr(Y = +1) = 1 / (1 + exp(-(-4 + score)))} \\ \hline
age at current charge$\leq$30 & 1 points & +... \\ \hline
number of prior violent charges$\geq$4 & 1 points & +... \\ \hline
number of prior arrests$\geq$7 & 1 points & +... \\ \hline
current violent charge=Yes & 1 points & +... \\ \hline
had charge(s) within last three year = Yes & 1 points & +... \\ \hline
\textbf{ADD POINTS FROM ROWS 1 TO 5}  &  	\textbf{SCORE} & = .....\\  \hline
\end{tabular}

\vspace*{2em}

% drug
\begin{tabular}{|p{4cm}|r|r|} \hline
\multicolumn{3}{c}{\textbf{Two Year Drug Recidivism}} \\ \hline
\multicolumn{3}{l}{Pr(Y = +1) = 1 / (1 + exp(-(-4 + score)))} \\ \hline
age at current charge$\leq$33 & 1 points & +... \\ \hline
number of prior drug related charges$\geq$1 & 1 points & +...\\ \hline
number of prior drug related charges$\geq$4 & 1 points & +...\\ \hline
\textbf{ADD POINTS FROM ROWS 1 TO 3}  &  	\textbf{SCORE} & = .....\\  \hline
\end{tabular}
% property

\vspace*{2em}
\begin{tabular}{|p{4cm}|r|r|} \hline
\multicolumn{3}{c}{\textbf{Two Year Property Recidivism}} \\ \hline
\multicolumn{3}{l}{Pr(Y = +1) = 1 / (1 + exp(-(-4 + score)))} \\ \hline
age at current charge $\leq$18 & 1 points & +... \\ \hline
age at current charge $\leq$23 & 1 points & +... \\ \hline
number of prior property related charges$\geq$1 & 1 points & +... \\ \hline
number of prior property related charges$\geq$5 & 1 points & +... \\ \hline
number of prior violent charges$\geq$4 & 1 points & +... \\ \hline
\textbf{ADD POINTS FROM ROWS 1 TO 5}  &  	\textbf{SCORE} & = .....\\  \hline
\end{tabular}

\vspace*{2em}

% felony
\begin{tabular}{|p{4cm}|r|r|} \hline
\multicolumn{3}{c}{\textbf{Two Year Felony Recidivism}} \\ \hline
\multicolumn{3}{l}{Pr(Y = +1) = 1 / (1 + exp(-(-3 + score)))} \\ \hline
age at current charge $\leq$33 & 1 points & +... \\ \hline
number of prior misdemeanor \par level charges$\geq$4 & 1 points & +... \\ \hline
number of prior property related charges$\geq$4 & 1 points & +... \\ \hline
\textbf{ADD POINTS FROM ROWS 1 TO 3}  &  	\textbf{SCORE} & = .....\\  \hline
\end{tabular}
% misdemeanor

\vspace*{2em}
\begin{tabular}{|p{4cm}|r|r|} \hline
\multicolumn{3}{c}{\textbf{Two Year Misdemeanor Recidivism}} \\ \hline
\multicolumn{3}{l}{Pr(Y = +1) = 1 / (1 + exp(-(-2 + score)))} \\ \hline
age at first charge$\leq$30 & 1 points & +... \\ \hline
number of FTA within last two years$\geq$1 & 1 points & +... \\ \hline
\textbf{ADD POINTS FROM ROWS 1 TO 2}  &  	\textbf{SCORE} & = .....\\  \hline
\end{tabular}
\end{center}
\end{table}
%%%%%%%%%%%%%%%%%%%%%%%%%%%%%%%%%%%%%%%%%%%%%%%%%

%%%%%%%%%%%%%% six month Broward %%%%%%%%%%%%%%%
\begin{table}[t]
\begin{center}
\scriptsize
\tabcolsep=0.1cm
\renewcommand{\arraystretch}{1.3}
\caption{Six Month Prediction Problems---Broward. Here, counts of prior arrests indicate the counts of arrests with at least one non-convicted or convicted charge. All charges mentioned are non-convicted charges.}
% general
\begin{tabular}{|p{4cm}|r|r|} \hline
\multicolumn{3}{c}{\textbf{Six Month General Recidivism}} \\ \hline
\multicolumn{3}{l}{Pr(Y = +1) = 1 / (1 + exp(-(-3 + score)))} \\ \hline
age at first charge$\leq$28 & 1 points & +... \\ \hline
had charge(s) within last three years = Yes & 1 points & +... \\ \hline
\textbf{ADD POINTS FROM ROWS 1 TO 2}  &  	\textbf{SCORE} & = .....\\  \hline
\end{tabular}
% violent

\vspace*{2em}
\begin{tabular}{|p{4cm}|r|r|} \hline
\multicolumn{3}{c}{\textbf{Six Month Violent Recidivism}} \\ \hline
\multicolumn{3}{l}{Pr(Y = +1) = 1 / (1 + exp(-(-4 + score)))} \\ \hline
current violent charge = Yes & 1 points & +... \\ \hline
number of prior violent charges $\geq$4 & 1 points & +... \\ \hline
had charge(s) within last three years = Yes & 1 points & +... \\ \hline
\textbf{ADD POINTS FROM ROWS 1 TO 3}  &  	\textbf{SCORE} & = .....\\  \hline
\end{tabular}

\vspace*{2em}

% drug
\begin{tabular}{|p{4cm}|r|r|} \hline
\multicolumn{3}{c}{\textbf{Six Month Drug Recidivism}} \\ \hline
\multicolumn{3}{l}{Pr(Y = +1) = 1 / (1 + exp(-(-5 + score)))} \\ \hline
age at first charge$\leq$21 & 1 points & +... \\ \hline
number of prior drug charges$\geq$2 & 1 points & +... \\ \hline
had charge(s) within last year = Yes & 1 points & +... \\ \hline
\textbf{ADD POINTS FROM ROWS 1 TO 3}  &  	\textbf{SCORE} & = .....\\  \hline
\end{tabular}
% property

\vspace*{2em}
\begin{tabular}{|p{4cm}|r|r|} \hline
\multicolumn{3}{c}{\textbf{Six Month Property Recidivism}} \\ \hline
\multicolumn{3}{l}{Pr(Y = +1) = 1 / (1 + exp(-(-5 + score)))} \\ \hline
age at current charge$\leq$29 & 1 points & + ... \\
number of prior misdemeanor level charges$\geq$5 & 1 points & +... \\ \hline
number of prior property related charges$\geq$1 & 1 points & +... \\ \hline
number of prior property related charges$\geq$4 & 1 points & +... \\ \hline
\textbf{ADD POINTS FROM ROWS 1 TO 4}  &  	\textbf{SCORE} & = .....\\  \hline
\end{tabular}

\vspace*{2em}

% felony
\begin{tabular}{|p{4cm}|r|r|} \hline
\multicolumn{3}{c}{\textbf{Six Month Felony Recidivism}} \\ \hline
\multicolumn{3}{l}{Pr(Y = +1) = 1 / (1 + exp(-(-3 + score)))} \\ \hline
age at current charge$\leq$29 & 1 points & +... \\ \hline
number of prior property related charges$\geq$4 & 1 points & +... \\ \hline
\textbf{ADD POINTS FROM ROWS 1 TO 2}  &  	\textbf{SCORE} & = .....\\  \hline
\end{tabular}
% misdemeanor

\vspace*{2em}
\begin{tabular}{|p{4cm}|r|r|} \hline
\multicolumn{3}{c}{\textbf{Six Month Misdemeanor Recidivism}} \\ \hline
\multicolumn{3}{l}{Pr(Y = +1) = 1 / (1 + exp(-(-3 + score)))} \\ \hline
age at current charge$\leq$19 & 1 points & +... \\ \hline
number of prior weapon related charges$\geq$1 & 1 points & +... \\ \hline
had charge(s) within last three years = Yes & 1 points & +... \\ \hline
\textbf{ADD POINTS FROM ROWS 1 TO 3}  &  	\textbf{SCORE} & = .....\\  \hline
\end{tabular}
\end{center}
\end{table}

\end{document}